\documentclass[final,1p,times,onecolumn]{elsarticle}

\usepackage{amssymb}
\usepackage{amsmath,amsfonts}
\usepackage{longtable}
\usepackage{geometry}
\usepackage{algorithmic}
\usepackage{algorithm}
\usepackage{array}
\usepackage{textcomp}
\usepackage{stfloats}
\usepackage{url}
\usepackage{verbatim}
\usepackage{graphicx}
\usepackage{subcaption}
\usepackage{changepage}
\usepackage[caption=false,font=normalsize,labelfont=sf,textfont=sf]{subfig}
\usepackage{booktabs}
\usepackage{tabularx}
\usepackage{arydshln}
\usepackage{xcolor}
\usepackage{tabularx}
\usepackage{pgfplots}
\usepackage{pgfplotstable}
\pgfplotsset{compat=1.17}
\newcommand\boldmin{\mathop{\mathbf{min}}}
\newcommand{\hl}[1]{\textcolor{black}{#1}}
\begin{document}
\begin{frontmatter}



\title{Self-supervised visual learning in the low-data regime: a comparative evaluation\tnoteref{FootDOI1}}
\tnotetext[FootDOI1]{Final author version. Accepted for publication in Neurocomputing, Elsevier. DOI: 10.1016/j.neucom.2024.129199}


\author[1]{Sotirios Konstantakos}
\ead{skwnstantakos@hua.gr}

\author[1]{Jorgen Cani}
\ead{cani@hua.gr}

\author[1]{Ioannis Mademlis}
\ead{imademlis@hua.gr}

\author[1]{Despina Ioanna Chalkiadaki}
\ead{it2022112@hua.gr}

\author[2]{Yuki M. Asano}
\ead{y.m.asano@uva.nl}

\author[2]{Efstratios Gavves}
\ead{egavves@uva.nl}

\author[1]{Georgios Th. Papadopoulos\corref{cor1}}
\ead{G.Th.Papadopoulos@hua.gr}
\cortext[cor1]{Corresponding author}

\affiliation[1]{organization={Department of Informatics and Telematics, Harokopio University of Athens},
            city={Athens},
            country={Greece}}
\affiliation[2]{organization={QUVA Lab, University of Amsterdam},
            city={Amsterdam},
            country={Netherlands}}

\begin{abstract}
Self-Supervised Learning (SSL) is a valuable and robust training methodology for contemporary Deep Neural Networks (DNNs), enabling unsupervised pretraining on a `pretext task' that does not require ground-truth labels/annotation. This allows efficient representation learning from massive amounts of unlabeled training data, which in turn leads to increased accuracy in a `downstream task' by exploiting supervised transfer learning. Despite the relatively straightforward conceptualization and applicability of SSL, it is not always feasible to collect and/or to utilize very large pretraining datasets, especially when it comes to real-world application settings. In particular, in cases of specialized and domain-specific application scenarios, it may not be achievable or practical to assemble a relevant image pretraining dataset in the order of millions of instances or it could be computationally infeasible to pretrain at this scale, e.g., due to unavailability of sufficient computational resources that SSL methods typically require to produce improved visual analysis results. This situation motivates an investigation on the effectiveness of common SSL pretext tasks, when the pretraining dataset is of relatively limited/constrained size. This work briefly introduces the main families of modern visual SSL methods and, subsequently, conducts a thorough comparative experimental evaluation in the low-data regime, targeting to identify: a) what is learnt via low-data SSL pretraining, and b) how do different SSL categories behave in such training scenarios. Interestingly, for domain-specific downstream tasks, in-domain low-data SSL pretraining outperforms the common approach of large-scale pretraining on general datasets. Grounded on the obtained results, valuable insights are highlighted regarding the performance of each category of SSL methods, which in turn suggest straightforward future research directions in the field.
\end{abstract}




\end{frontmatter}

\section{Introduction}
\label{sec::Introduction}
The introduction and wide use of large-scale Deep Neural Networks (DNNs) has resulted in revolutionary effects and tremendous advances in the broader field of Artificial Intelligence (AI) over the past decade, by establishing a well-defined methodology where optimal input representations are learnt, with respect to a given task, from massive quantities of training data \cite{lecun2015deep}. In this respect, the computer vision field has significantly benefited from the use of DNNs for tasks such as object detection \cite{girshick2014rich} \cite{ren2015faster}, semantic segmentation \cite{long2015fully} \cite{chen2017deeplab}, image captioning \cite{vinyals2015show}, etc. The typical approach is to employ supervised training, where a large number of images tightly coupled with their ground-truth labels are jointly utilized, so that the DNN directly learns the desired downstream task (i.e. a direct mapping from the input images to the desired high-level semantic entities). However, training from scratch, i.e., with randomly initialized DNN parameters, usually requires extremely large annotated datasets and a significant amount of computational power. Both of these resources are practically limited in the majority of real-world cases and for most computer vision tasks. Notably, ground-truth labels can be scarce or expensive to obtain for large datasets: manual annotation may be time-consuming, costly, requiring specialized expertise, and/or contaminating the data with human biases and subjective judgments.

The typical approach for developing a visual analysis module for a real-world application relies on the adoption of the so-called `transfer learning' paradigm, i.e. pretraining a DNN (from scratch) on a large-scale annotated dataset and then fine-tuning it in the targeted downstream task; both steps follow a supervised learning methodology, i.e. they require the availability of ground-truth annotated datasets. The most common approach by far is to pretrain for whole-image classification on ImageNet-1k \cite{imagenet_cvpr09}. ImageNet-1k is composed of approximately 1.3 million manually annotated images across 1,000 classes. The resulting DNN model encodes knowledge about image semantics and can be exploited as a good initialization for finetuning on any downstream task; this way, comparatively small and limited annotated downstream datasets can be efficiently utilized \cite{Wang2023}. ImageNet-pretrained models are publicly available for several well-known DNN architectures, such as variants of the ResNet Convolutional Neural Network (CNN) \cite{he2016deep}. However, the more the downstream and the pretraining datasets/tasks differ from each other, the less effective this strategy is \cite{risojevic2021we}. Thus, on the one hand, transfer learning does alleviate the burden of manually annotating a massive number of images for each desired downstream task. On the other hand, the bottleneck of manual labelling remains intact also for the large-scale pretraining dataset, in cases where the publicly available DNN models that have been pretrained on ImageNet-1k (or any other broad/massive dataset) prove to be unsuitable for a given downstream task/domain.

\begin{figure}[htbp]
    \centering
    \includegraphics[width=0.8\linewidth]{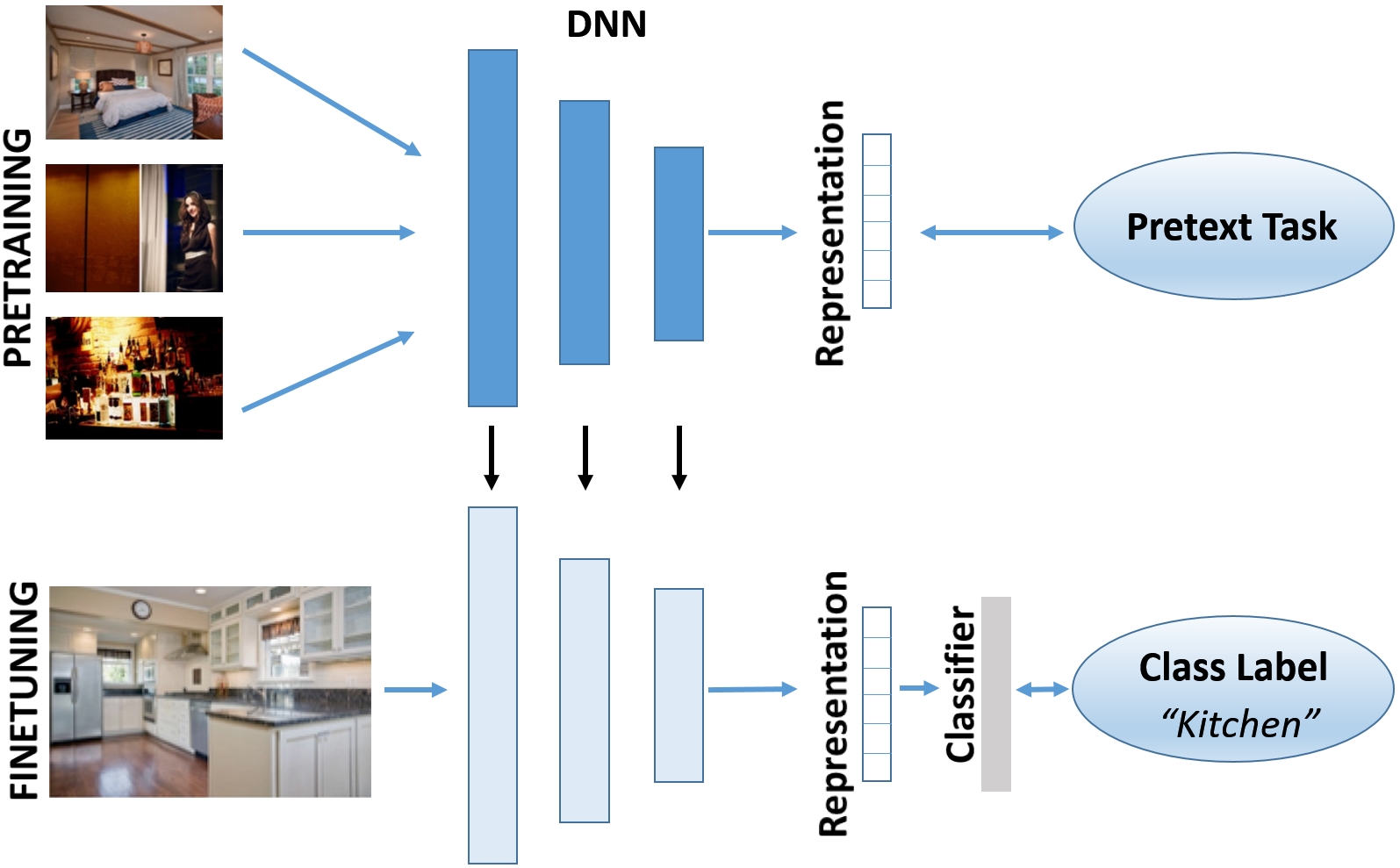}
    \caption{Conceptualization of the SSL paradigm}
    \label{fig::SSLFlow}
\end{figure}

Self-Supervised Learning (SSL) has emerged as an alternative transfer learning strategy, where the DNN is pretrained on a `pretext task' that does not require ground-truth labels/annotation to be available. Essentially, pseudo-labels are automatically created from the training images themselves \cite{chen2021ssl++}. Thus, massive domain-specific datasets can easily be exploited for pretraining, without requiring any human annotation effort; such unlabelled images may be abundant. In fact, SSL has proven beneficial even if ground-truth annotation/labels are available for the massive pretraining dataset, but simply ignored by the SSL pretext task. There are many examples in the literature where SSL pretraining leads to higher downstream accuracy (after task-specific finetuning) than regular supervised pretraining \cite{azizi2023robust}. Table \ref{tab::SSLTerms} defines the most common relevant terms, while Figure \ref{fig::SSLFlow} illustrates the conceptualization of the SSL paradigm. Overall, SSL leverages the visual structure and relationships present in the images themselves to provide supervision signals for representation learning, through pretext tasks, such as completing an image or reconstructing it after applying various transformations (e.g., color alternations, translations, etc.). The typical goal is for the DNN to learn semantically rich image representations without relying on ground-truth labels, although SSL has alternatively been applied to learning features that encode image/style aesthetics \cite{Zhang2023}. Thorough experimental evaluations have so far showcased that the very large amounts of data allow the DNNs to learn useful image representations during pretraining, even without exploiting ground-truth semantics \cite{konkle2022self}. To this end, the dominant trend in recent years is to scale SSL regarding both the pretraining dataset size and the DNN model complexity \cite{goyal2021self}, where even ImageNet-1k being considered rather small under contemporary standards.

Although the SSL methodology has proven beneficial in the case of abundance of relevant unlabelled data, it is not always feasible or practical to assemble and/or to utilize very large pretraining datasets in real-world scenarios. In particular, for several application cases (e.g., medical imaging domain) it can be a significant challenge to create such a relevant image pretraining dataset in the order of millions of instances, albeit no manual annotation is required for SSL. Additionally, even in the case that such a sizeable set of relevant images can be collected, it is not always feasible to pretrain DNNs at this scale, e.g., due to unavailability of sufficient computational resources that SSL methods typically require to produce improved visual analysis results.


\begin{table*}[!htbp]
\centering
\caption{Definition of main terms in self-supervised learning}
\begin{tabular}{lp{9cm}}
\toprule
\begin{tabular}{l}Term\end{tabular}
&\begin{tabular}{p{5cm}}Definition\end{tabular}\\
\midrule
\begin{tabular}{l}Pretext task\end{tabular}
&\begin{tabular}{p{9cm}}An auxiliary task for DNN pretraining, which exploits parts of the data themselves as pseudo-ground-truth. It allows the DNN to learn useful data representations without utilizing actual ground-truth annotation. Different pretext tasks can be alternatively employed for a given pretraining dataset.\end{tabular}\\
\begin{tabular}{l}Downstream task\end{tabular}
&\begin{tabular}{p{9cm}}The desired application (e.g., whole-image classification, object detection, etc.) to which the pretrained DNN will be adjusted. A single pretrained DNN can be adjusted for multiple different downstream tasks.\end{tabular}\\
\begin{tabular}{l}Downstream dataset\end{tabular}
&\begin{tabular}{p{9cm}}Each downstream task implies the existence of a relevant downstream dataset, typically accompanied by actual relevant ground-truth annotation. The downstream dataset is usually significantly smaller than the pretraining one.\end{tabular}\\
\begin{tabular}{l}Pseudo-labels\end{tabular}
&\begin{tabular}{p{9cm}}Pseudo-ground-truth labels automatically generated from the pretraining data themselves, based on the type of the employed pretext task.\end{tabular}\\
\begin{tabular}{l}Pretrained model\end{tabular}
&\begin{tabular}{p{8cm}}A DNN pretrained on a dataset without exploiting ground-truth labels, using a pretext task.\end{tabular}\\
\begin{tabular}{l}Finetuning\end{tabular}
&\begin{tabular}{p{9cm}}The process of adjusting the parameters of a pretrained DNN, by partially retraining it on a specific downstream task.\end{tabular}\\
\begin{tabular}{l}Transfer learning\end{tabular}
&\begin{tabular}{p{9cm}}The act of exploiting data representations learnt via pretraining a DNN, while finetuning it for a downstream task. The most common approach is to simply initialize the DNN architecture with the pretrained model's parameters, before finetuning it.\end{tabular}\\
\begin{tabular}{l}Linear probing\end{tabular}&
\begin{tabular}{p{9cm}}A common method for evaluating the quality of learnt data representations, by training a linear classifier on top of a frozen pretrained DNN. This is the simplest downstream benchmark, since the limited capabilities of the linear classification head highlight the quality of the learnt representations.\end{tabular}\\
\bottomrule
\end{tabular}
\label{tab::SSLTerms}
\end{table*}

The challenges and limitations faced when applying the SSL methodology in real-world scenarios provide a strong motivation for investigating the effectiveness of the most common SSL pretext tasks in situations where the pretraining dataset is of relatively limited size. So far, the relevant literature has markedly not explored the behavior of SSL methods when the assumption of abundance of relevant data is not present, since the ability to pretrain on (at least) ImageNet-scale datasets is almost always assumed to be a prerequisite. Yet, a study in the low-data regime would be important, but currently missing, for practitioners who necessarily work with specific image domains (e.g., X-rays), where it is difficult to obtain massive amounts of even unlabeled data.

This work attempts to remedy this gap in the literature and to conduct a thorough comparative evaluation of self-supervised visual learning methods in the low-data regime. For the purposes of this study, `low-data' has been selected to mean a pretraining dataset with a size in the range of ~50k - ~300k images, which is significantly smaller than the typical dataset sizes used for SSL pretraining (e.g., ImageNet-1k that contains ~1.2 million labeled images). The main goals of this study is to identify: a) what is learnt via SSL pretraining in domains where there are no (extremely) large-scale datasets available, and b) how do different SSL categories of methods behave in such training scenarios. To this end, the main contributions of this work are the following ones:
\begin{itemize}
   \item An in-depth comparative experimental evaluation is conducted under the assumption of performing pretraining in the low-data regime, using one representative pretext task from each of the 4 most common SSL categories. Downstream accuracy metrics are supplemented by visual/qualitative explanations and statistical analyses of the image representations learnt by the various SSL methods. Comparisons are made against traditional baselines, such as supervised pretraining approaches or training from scratch with random parameter initialization, depending on the specific setup.
   \item Grounded on the obtained results, valuable insights are highlighted regarding the performance of each main category of SSL methods (or individual SSL approaches), which in turn suggest straightforward future research directions in the field.
\end{itemize}

Although SSL pretext tasks can be designed and employed for many different types of data (e.g., timeseries \cite{Ying2024}, text \cite{kenton2019bert}, video \cite{schiappa2023self} \cite{Chen23}, audio \cite{liu2022audio}, point clouds \cite{Wu23}, or even multimodal data \cite{Tao23} \cite{Luo22}), this article focuses on image analysis for computer vision applications. Moreover, it focuses on generic SSL methods and not ones explicitly designed for specific tasks (e.g., for multi-view clustering \cite{Huang2023}, product attribute recognition \cite{Dai2023}, etc.). The remainder of this paper is organized as follows: Section \ref{sec::PretextTasks} briefly presents the most common categories of pretext tasks for visual SSL. In Section \ref{sec::Methodology}, the experimental evaluation methodology and all pretraining or downstream datasets/tasks utilized in this work are detailed. Section \ref{sec::Experiments} illustrates and discusses the obtained results, emphasizing on the extraction of valuable observations and insights. Conclusions about the potential of SSL in the low-data regime are drawn in Section \ref{sec::Conclusions}. Finally, the most important algorithms falling under each of the main SSL pretext task categories are detailed in the Appendix.

\section{Pretext Tasks}
\label{sec::PretextTasks}


\begin{figure*}[t]
    \centering
    \begin{subfigure}{0.45\textwidth}
        \centering        \includegraphics[width=\linewidth]{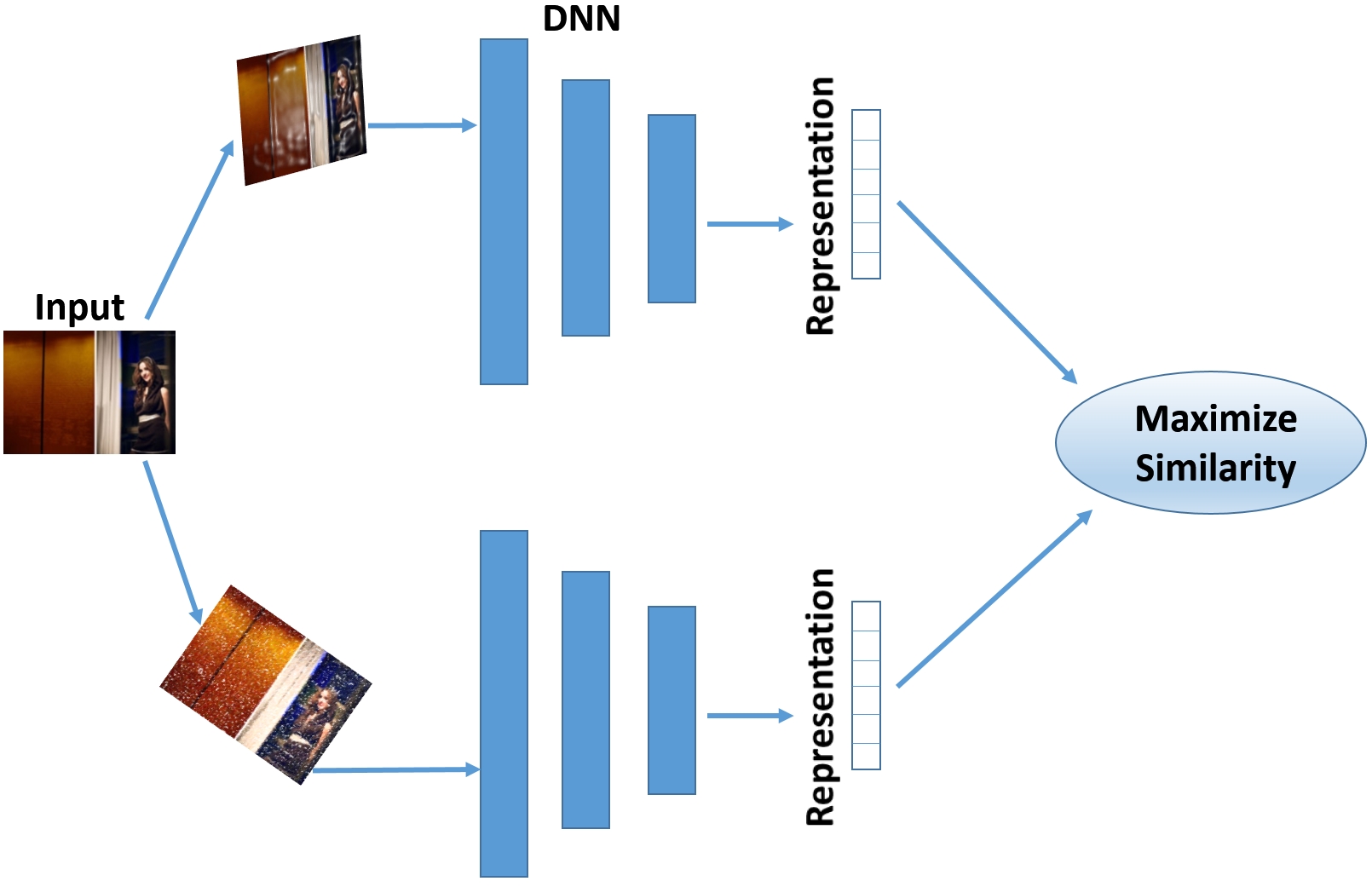}
        \caption{Contrastive}
        \label{fig:Contrastive}
    \end{subfigure}
    \hfill
    \begin{subfigure}{0.45\textwidth}
        \centering
        \includegraphics[width=\linewidth]{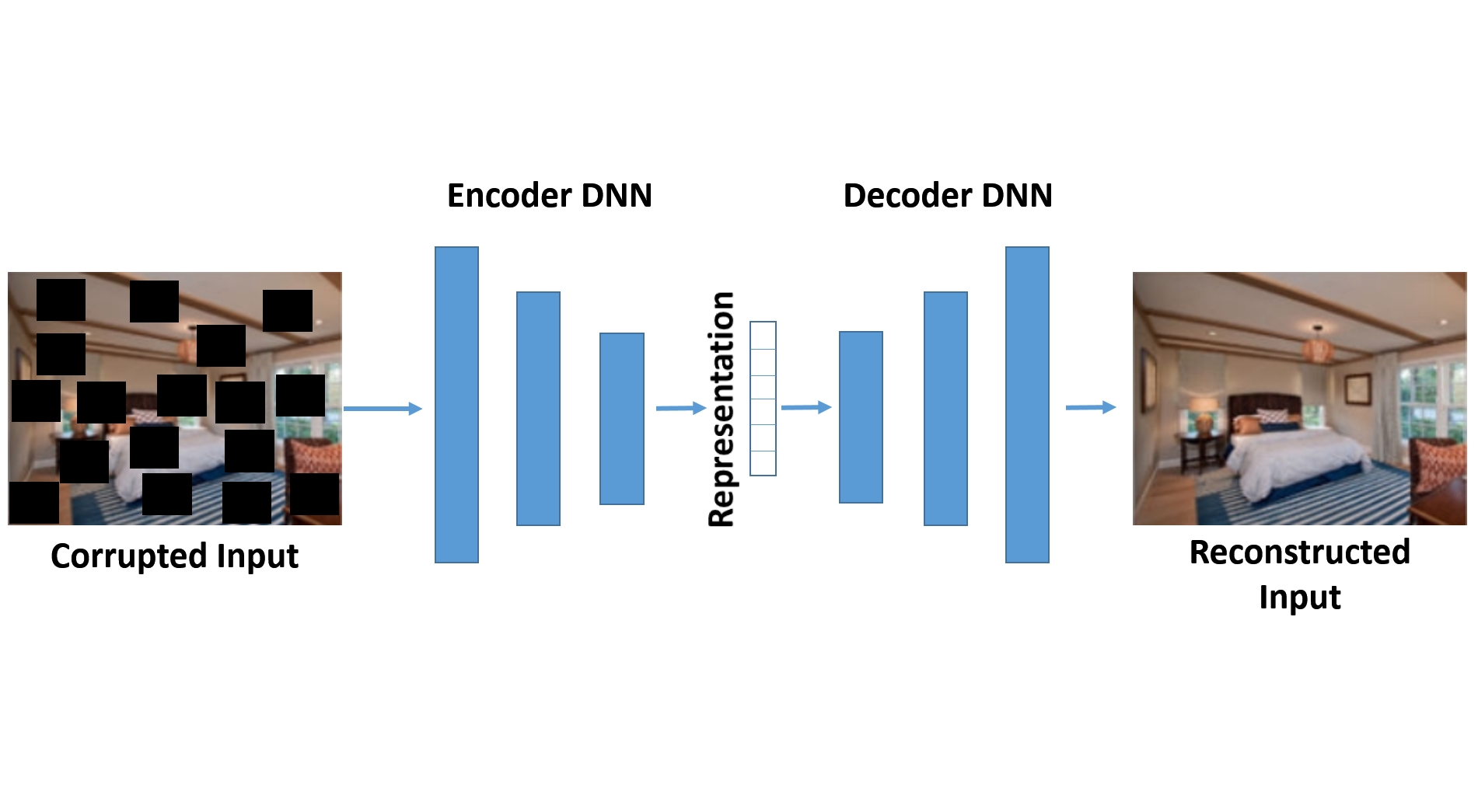}
        \caption{Generative}
        \label{fig:Generative}
    \end{subfigure}
    
    \medskip
    
    \begin{subfigure}{0.45\textwidth}
        \centering
        \includegraphics[width=\linewidth]{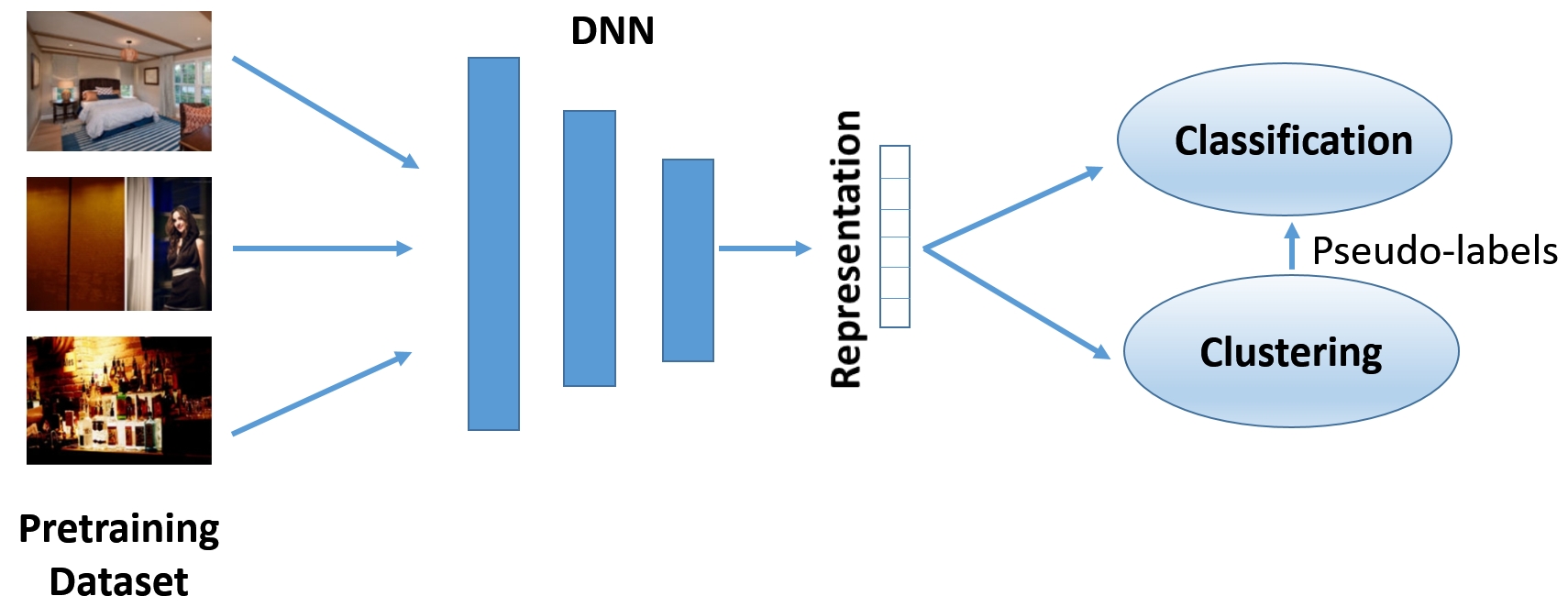}
        \caption{Clustering}
        \label{fig:Clustering}
    \end{subfigure}
    \hfill
    \begin{subfigure}{0.45\textwidth}
        \centering
        \includegraphics[width=\linewidth]{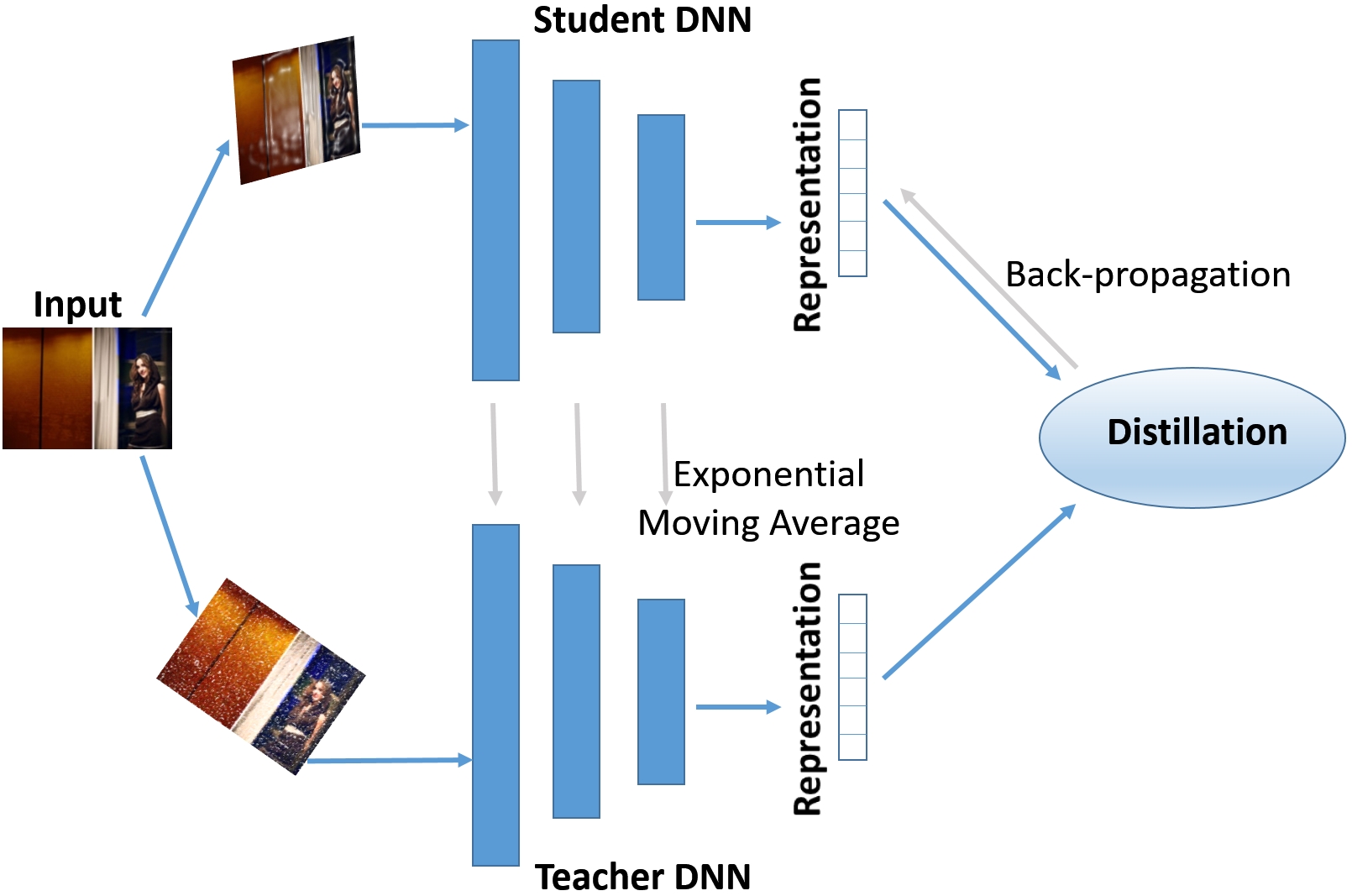}
        \caption{Self-distillation}
        \label{fig:SelfDistillation}
    \end{subfigure}
    \caption{Abstract indicative illustrations of the 4 main categories of SSL pretext tasks}
   \label{fig:SSL_categories}
\end{figure*}

Pretext tasks are auxiliary objectives/assignments that a DNN learns during SSL pretraining, without relying on explicit ground-truth annotation/labels. Each pretext task essentially defines a way to extract pseudo-labels from the training images themselves, so that a DNN being trained for this task gradually learns to capture the visual structure and regularities of the dataset. Simple, common tasks of this nature involve, for example, the prediction of the rotation angle of an image \cite{gidaris2018unsupervised} or the relative position of image blocks \cite{doersch2015unsupervised}, after applying to the input image a rotation transformation or splitting it into rectangular blocks, respectively. Pretraining the DNN for such a pretext task allows it to learn meaningful representations from unlabeled data without any explicit supervision. This knowledge can then be exploited via regular supervised finetuning of the SSL-pretrained DNN model on a separate annotated downstream dataset, designed for the actually desired task/function. The ability of SSL pretraining to increase accuracy in downstream tasks, compared to direct training from scratch with a randomly initialized DNN model or even to apply transfer learning subsequently to traditional supervised pretraining, depends heavily on the quality and meaningfulness of the considered pretext task.

\subsection{Main categories of SSL pretext tasks for images}
Taking into account the different types of pretext tasks employed during SSL pretraining, literature methods can be divided into four main categories, namely contrastive, generative, clustering and self-distillation. Figure \ref{fig:SSL_categories} abstractly summarizes how the aforementioned main categories of SSL pretext tasks operate, based on key representative approaches per case. It needs to be highlighted that in all cases the actual goal of SSL pretraining is for the DNN to learn semantically meaningful visual representations from the input images, instead of simply aiming at solving the pretext task itself. The main SSL pretext task categories are as follows:
\begin{itemize}
    \item Contrastive: The aim is for the DNN to learn to differentiate between similar (positive) and dissimilar (negative) images. Since no ground-truth semantics/class labels are exploited, the positive pairs are constructed by simply applying data augmentation transformations to an anchor image, which modify its appearance but not its semantic content. In contrast, a negative pair is constructed by simply selecting two different training images. The DNN is trained to generate internal representations that are close to/away from each other in the latent space for positive/negative image pairs, respectively, via contrastive loss functions such as InfoNCE \cite{oord2018representation}.
    \item Generative: The DNN is trained to output information reflecting the input image. Typically, the latter has been modified so that it is partially occluded, missing or transformed. Then, the target output is either the original, unmodified image, or a vector encoding the transformation that has been applied to the input. Architecturally, variants of Denoising Autoencoders \cite{vincent2008extracting} are the most common form of DNN to be utilized; these consist of an Encoder and a consecutive Decoder subnetwork, with the Encoder being typically the one which is retained for downstream finetuning. Other generative neural architectures have also been employed, such as Variational Autoencoders (VAEs) \cite{kingma2013auto} or Generative Adversarial Networks (GANs) \cite{goodfellow2014generative}. When the input is partially missing/occluded and the desired target output is the full original image, generative pretext tasks are essentially image reconstruction tasks.
    \item Clustering: Clustering pretext tasks typically employ an image clustering step, using any clustering algorithm such as K-Means \cite{Lloyd1982}, which groups the set of training images into a number of clusters. Then, during SSL pretext pretraining, these cluster assignments are utilized as pseudo-labels for supervised classification. Alternating phases of clustering and SSL, as well as on-line clustering approaches have also been attempted.
    \item Self-distillation: In neural distillation, a `student' DNN is trained by utilizing the predictions of another `teacher' DNN as target pseudo-ground-truth for a given input \cite{hinton2015distilling}. In SSL self-distillation pretext tasks, the teacher and the student are usually different variants of the single DNN model being pretrained; for instance, during different epochs or stages of training. The target of the self-distillation loss may be the latent image representation itself, while a pair derived from a single training image (e.g., the original and one transformed/augmented variant) may be utilized as respective input for the two DNNs.
\end{itemize}

The most commonly employed loss functions for SSL DNN pretraining in computer vision are summarized in Table \ref{tab::SSLLossFunctions}. In Cross-Entropy (CE), which is used for classification, $C$ is the total number of classes, $\mathbf{y},\mathbf{p} \in \mathbb{R}^{C}$ denote the target and predicted output vectors, respectively, while $y_c$/$p_c$ is the ground-truth/predicted probability corresponding to the $c$-th class. In the cosine similarity loss function, $\mathbf{v}_1$ and $\mathbf{v}_2$ are two feature vectors whose similarity is being assessed. In the InfoNCE loss, $sim(\cdot)$ typically corresponds to cosine similarity, $\mathbf{a} \in \mathbb{R}^{L}$ and $\mathbf{p} \in \mathbb{R}^{L}$ are the representation vectors of an anchor and a `positive' image belonging to the same class, respectively, while $\mathbf{R} \in \mathbb{R}^{L \times (B+1)}$ contains in its columns $\mathbf{r}_k$, $1 \leq k \leq B+1$, the vector $\mathbf{p}$ and the representation vectors of $B$ `negative' images. For both Mean Squared Error (MSE) and Mean Absolute Error (MAE), $\mathbf{v} \in \mathbb{R}^L$ and $\hat{\mathbf{v}} \in \mathbb{R}^L$ are the pseudo-ground-truth and the predicted/reconstructed vectors, respectively, while $L$ is typically the dimensionality of an image representation.

Algorithmic details about a selection of important SSL methods for each of the four categories are provided in the Appendix of this article. Further details and a comprehensive taxonomy can be found in the survey \cite{Gui2023survey}.

\begin{table*}
\centering
\caption{Summary of most commonly employed loss functions in visual SSL}
\label{tab::SSLLossFunctions}
\small
\begin{tabularx}{\textwidth}{|>{\centering\arraybackslash}p{4cm}|X|}
    \hline
    \begin{tabular}{c}Loss function\end{tabular}&\begin{tabular}{l}Details\end{tabular}\\
    \hline
    \begin{tabular}{c}Cross-Entropy (CE)\end{tabular}& 
    \begin{tabular}{l}
    \textit{Expression:} $\mathcal{L}_{\text{CE}}(\mathbf{y}, \mathbf{p}) = -\sum_{c=1}^{C} y_c \log(p_c)$\\
    \textit{Applications:} Clustering, self-distillation.\\
    \end{tabular}\\
    \hline
    \begin{tabular}{c}Cosine similarity\end{tabular}&
    \begin{tabular}{l}\textit{Expression:} $\mathcal{L}_{\text{sim}}(\mathbf{v}_1, \mathbf{v}_2) = \frac{\mathbf{v}_1^T \mathbf{v}_2}{\|\mathbf{v}_1\| \|\mathbf{v}_2\|}$\\
    \textit{Applications:} Contrastive, self-distillation.\\
    \end{tabular}\\
    \hline
    \begin{tabular}{c}InfoNCE (Noise\\Contrastive Estimation) \cite{gutmann2012noise}\end{tabular}&
    \begin{tabular}{l}\textit{Expression:} $\mathcal{L}_{\text{NCE}}(\mathbf{a}, \mathbf{p}, \mathbf{R}) = -\log \left( \frac{\exp(\text{sim}(\mathbf{a}, \mathbf{p}))}{\sum_{k=1}^{B+1} \exp(\text{sim}(\mathbf{a}, \mathbf{r}_k))} \right)$\\
    \textit{Applications:} Contrastive training in visual SSL.\\
    \end{tabular}\\
    \hline
    \begin{tabular}{c}Mean Squared Error (MSE)\end{tabular}&
    \begin{tabular}{l}\textit{Expression:} $\mathcal{L}_{\text{MSE}}(\mathbf{v}, \hat{\mathbf{v}}) = \frac{1}{L} \sum_{i=1}^{L} (v_i - \hat{v}_i)^2$\\
    \textit{Applications:} Self-distillation, image reconstruction generative SSL\\
    \end{tabular}\\
    \hline
    \begin{tabular}{c}Mean Absolute Error (MAE)\end{tabular}&
    \begin{tabular}{l}
    \textit{Expression:} $\mathcal{L}_{\text{MAE}}(\mathbf{v}, \hat{\mathbf{v}}) = \frac{1}{L} \sum_{i=1}^{L} |v_i - \hat{v}_i|$\\ \textit{Applications:} Image reconstruction generative SSL.\\
    \end{tabular}\\
    \hline
\end{tabularx}
\end{table*}

\subsection{Discussion on the behavior of the SSL pretraining categories}
The main categories of SSL methods for image analysis tasks, as presented in this section, are characterized by various trade-offs in different aspects. Two important points of comparison comprise the following ones: a) the reliance on handcrafted design choices (e.g., augmentation transformations, generative pseudo-ground-truth targets, etc.) as prior knowledge, b) the susceptibility of training to solution/representation collapse modes.


With respect to the first point, the majority of SSL methods relies on prior knowledge and exhibits various degrees of robustness to such choices. Essentially, the worst offenders are early generative SSL methods (e.g., rotation prediction, colorization, \hl{jigsaw puzzle reassembly}, etc.), which \hl{exploit known visual structure of common images \cite{Zhao2020SSL} and} can be nowadays considered outdated. Additionally, contrastive, self-distillation and clustering SSL approaches heavily depend on the data augmentation transformations applied during pretraining, which is a form of prior knowledge about the visual medium \hl{\cite{Zhai2024}. On top of this, proper selection of said image transformations relies on knowledge of the specific desired downstream tasks, due to different invariances. For instance, multi-cropping and rotation may be semantically irrelevant for whole-image classification, but not as much in the case of monocular depth estimation, while aggressive cropping aids occlusion invariance but potentially limits other desirable invariances \cite{Purushwalkam2020}}. As a result, pretraining with these common augmentation transformations in order to obtain high-level semantic image representations that are invariant to them may be inappropriate for low-level downstream vision tasks. Among the common SSL algorithms, the only ones that do not rely at all on similar forms of prior knowledge are the image reconstruction generative tasks \hl{\cite{Moutakanni2024}}.

The image reconstruction generative SSL approaches also inherently do not suffer from the danger of trivial solutions via representation collapse, which is theoretically an additional advantage. However, in practice, recent methods in categories that are the most susceptible to this scenario, i.e., self-distillation and SSL via clustering, have converged to clever methodological and/or implementation nuisances for essentially bypassing this issue and avoiding collapse altogether \hl{\cite{He2024preventing}}. Mainstream contrastive SSL methods, which inherently do not suffer from collapse issues due to negative sampling, typically come with the cost of requiring very large mini-batch sizes or auxiliary structures (e.g., a large memory bank) \hl{\cite{Jang2023Self}}. A well-known exception is Barlow Twins \cite{zbontar2021barlow}, since it does not use the contrastive loss and does not employ negative sampling, but in fact DNNs pretrained with it struggle to reach good downstream accuracy without the use of very high-dimensional latent representations \hl{\cite{zbontar2021barlow} \cite{Kalantidis2021}}.

It has to be noted that it is a self-distillation SSL approach, i.e., DINO \cite{caron2021emerging}, that has exhibited strong zero-shot learning capabilities when combined with a complex ViT, with DINOv2 \cite{Oquab2023DinoV2} pretrained on an extremely large-scale dataset reaching the zero-shot performance of competing weakly supervised methods. Essentially, very high model complexity and pretraining dataset size contribute significantly to the performance of DINOv2, but nevertheless self-distillation seems to currently dominate the SSL landscape in computer vision. Even so, research on image reconstruction generative approaches and on SSL via on-line clustering is still on-going.

Regarding robustness analysis, several common SSL-learnt representations have been empirically evaluated against downstream distribution shifts and corruptions in \cite{chhipa2023can}. DINO, a self-distillation method, and SwAV \cite{caron2020unsupervised}, an approach of SSL via on-line clustering, are found to be the most robust. However, no image reconstruction generative pretext task is considered in that comparative evaluation. On the other hand, the reconstructive method MAE \cite{he2022masked} is explicitly tested with regard to its pretraining data efficiency in \cite{Kong2023}: the study concludes that MAE can learn occlusion-invariant image representations from rather small datasets, but lacks a systematic comparative evaluation of the issue. Nevertheless, it is shown in \cite{Xie2023} that reconstructive generative pretext tasks lead to overfitting on small pretraining datasets, if the underlying DNN model is sufficiently complex.


\section{Comparative Evaluation Framework in the Low-Data Regime}
\label{sec::Methodology}
The goal of this study is to investigate in depth how the different SSL methodologies behave and what eventual downstream recognition performance is achieved, when pretraining occurs in the low-data regime. This stands in contrast to the currently dominant trend of ever-increasing scale, regarding both the pretraining dataset size and the DNN model complexity in SSL research. As argued in Section \ref{sec::Introduction}, low-data pretraining remains highly relevant in cases of domain-specific downstream tasks, where generic ImageNet-1k pretraining-derived or zero-shot representations cannot naturally generalize well. Therefore, a thorough empirical comparative evaluation is designed that incorporates all SSL algorithmic categories outlined in Section \ref{sec::PretextTasks}, in order to identify what can be learnt via SSL in low-data computer vision scenarios.

As mentioned in Section \ref{sec::Introduction}, `low-data' \hl{in this article} implies a pretraining dataset with a \hl{total} size in the range of ~50k - ~300k images, which is significantly smaller than the typical dataset sizes used for SSL pretraining in the literature (e.g., ImageNet-1k). \hl{Early exploratory experiments indicated that SSL pretraining is largely ineffective on datasets with a total size below this range, i.e., lower than ~50k images, since the lack of ground-truth labels makes it critical to have enough images for learning adequate representations. This observation is in line with published literature (e.g., \cite{Xie2023} \cite{El2021}), leading to a reasonable choice of ~50k images (in total) as the lower bound of a meaningful SSL low-data regime. Still, even in such datasets, their actual training subset is obviously in fact smaller.}

In the defined evaluation framework, 4 widely used SSL methods from different algorithmic categories are compared under different setups on image classification and, in one case, on object detection. The considered approaches are MAE \cite{he2022masked}, SimCLR \cite{chen2020simple}, DINO \cite{caron2021emerging} and DeepClusterV2 \cite{caron2020unsupervised} that belong to the generative, contrastive, self-distillation and clustering SSL category, respectively. It needs to be highlighted that the 4 aforementioned methods do not necessarily correspond to the 4 best-performing in the literature; however, they are relatively recent, representative of their respective algorithmic category and exhibit performance comparable to the state-of-the-art. The employed neural architecture for each considered SSL method has been selected based on the design choices described in the original paper of each approach; thus, ResNet-50 \cite{he2016deep} is used for the cases of SimCLR and DeepClusterV2, while ViT-L/16 \cite{dosovitskiy2020image} is utilized for MAE and DINO. This strategy was preferred because not all SSL methods seem to work equally well with Vision Transformers and each SSL approach has an informally declared `default' architecture. For the baselines pretrained in a supervised manner, both ResNet-50 and ViT-L/16 variants have been employed. As an additional baseline, an off-the-shelf pretrained ViT-L/14 DINOv2 model has also been evaluated. In all cases, except in the object detection experiment, linear probing has been employed for downstream finetuning, instead of end-to-end finetuning, due to its widespread usage in the SSL literature and its ability to demonstrate the quality of the representations learnt during pretraining. To this end, each pretrained model is kept frozen and its inferred features are utilized to train a simple downstream linear classifier. In the followings, this linear probing setup is implied wherever finetuning is mentioned for whole-image classification.

In order to better structure the wide set of performed experiments and to facilitate the derivation of key/detailed insights, the experiments in the defined experimental framework are divided into the following main classes: a) Main experiments, which are related to whole-image classification tasks; these are further broken down into i) object- and scene-centric (based on the recognition target of the downstream task) and ii) transfer-learning and single-dataset setup (taking into account whether different datasets are used for the pretraining and downstream tasks or not), b) Robustness experiments, which investigate robustness aspects of the various SSL pretraining tasks, and c) Domain-specific experiments, which investigate SSL performance in specialized visual domains that differ significantly from typical/conventional natural RBG images. Four different domains are evaluated using a whole-image classification downstream task, while one domain (UAV-captured infrared footage) is evaluated on an object detection downstream task.

\subsection{Main experiments}
\label{sec::MainExperiments}
The main experiments concern whole-image classification tasks and are divided into: a) object- and scene-centric, taking into account the semantic granularity of the recognition target of the downstream task, and b) transfer-learning and single-dataset, considering the dataset setup (i.e., when the pretraining dataset and the downstream dataset for supervised finetuning differ or not, respectively). ImageNet-100\footnote{This is a low-data variant of the ImageNet-1k dataset \cite{imagenet_cvpr09}, transformed according to \cite{zhao2020maintaining}} and Places-100\footnote{This is a low-data variant of the Places-365 dataset \cite{zhou2017places}, transformed to Places-100 according to \cite{he2021quantifying}} are selected as the pretraining dataset for the object- and scene-centric case, accordingly. Additionally, the considered downstream datasets for the transfer-learning setup are COCO \cite{lin2014microsoft} and STL-10 \cite{coates2011stl10} for the object-centric case, and ADE20K \cite{zhou2017scene} and SUN Database \cite{xiao2010sun} for the scene-centric one. For the single-dataset setup, where both the SSL pretraining and the subsequent supervised downstream finetuning tasks are conducted on a single dataset, ImageNet-100 and Places-100 are used for the object- and the scene-centric cases, respectively. Table \ref{tab::MainExperimentsConfiguration} summarizes the configurations of the considered main experiments, while notation is also included for clarity purposes.

Moreover, the employed SSL pretraining methods are quantitatively compared with respect to downstream classification accuracy against baseline benchmarks, so as to better demonstrate the recognition capabilities of the various SSL configurations. To this end, in the transfer-learning setup, two sets of DNN variants pretrained according to the regular supervised approach are considered, namely one pretrained on ImageNet-1k (object-centric) or Places-205 (scene-centric) and one on ImageNet-100 (object-centric) or Places-100 (scene-centric). Both of these exploit supervised pretraining, but the first one, i.e., pretraining on ImageNet-1k or Places-205, is not low-data. Additionally, an off-the-shelf DINOv2 model, which has been distilled from a DNN pretrained on a dataset of 142 million images (again, non-low-data), is also employed as a competing baseline.
On the other hand, the single-dataset experiments are also repeated under the semi-supervised downstream finetuning protocol described in \cite{lee2013pseudo}\cite{zhai2019s4l}. This includes two different variants of downstream finetuning, namely one using 5\% and one using 10\% of the ground-truth class labels.

\begin{table*}
\caption{\hl{Configurations of main experiments}}
\label{tab::MainExperimentsConfiguration}
\small
\setlength{\tabcolsep}{4pt}
\resizebox{\textwidth}{!}{
\begin{tabular}{c|c|c|c|c}
\hline
Granularity & Setup & Pretraining & Downstream & Notation \\
\hline\hline
Object-centric &
\begin{tabular}{c} Transfer-learning \\ Transfer-learning \\ Single-dataset \end{tabular} &
\begin{tabular}{c} ImageNet-100 \\ ImageNet-100 \\ ImageNet-100 \end{tabular} &
\begin{tabular}{c} COCO \\ STL-10 \\ ImageNet-100 \end{tabular} &
\begin{tabular}{c} $M^O_T$(IN-100$\downarrow$COCO) \\ $M^O_T$(IN-100$\downarrow$STL-10) \\ $M^O_S$(IN-100$\downarrow$IN-100) \end{tabular} \\
\hline
Scene-centric &
\begin{tabular}{c} Transfer-learning \\ Transfer-learning \\ Single-dataset \end{tabular} &
\begin{tabular}{c} Places-100 \\ Places-100 \\ Places-100 \end{tabular} &
\begin{tabular}{c} ADE20K \\ SUN Database \\ Places-100 \end{tabular} &
\begin{tabular}{c} $M^S_T$(P-100$\downarrow$ADE) \\ $M^S_T$(P-100$\downarrow$SUN) \\ $M^S_S$(P-100$\downarrow$P-100) \end{tabular} \\
\hline
\end{tabular}
}
\end{table*}

\subsection{Robustness experiments}
\label{sec::RobustnessExperiments}
A complementary set of experiments is conducted in order to evaluate robustness aspects of the considered SSL methods. For that purpose, three typical robustness experimental benchmarks/protocols of the literature are considered, namely a) Noisy-ImageNet-100, b) Imbalanced-ImageNet-100, and c) VTAB, as detailed below.

Regarding the Noisy-ImageNet-100 benchmark, it involves the evaluation of robustness, using three datasets that have been derived by applying various forms of noise, perturbations and corruptions to a subset of ImageNet-1k or by selecting naturally adversarial images for its classes; the considered datasets are ImageNet-A \cite{hendrycks2021natural}, ImageNet-P and ImageNet-C \cite{hendrycks2019benchmarking}. For each considered SSL method, the corresponding ImageNet-100-pretrained variant from the main experiments (i.e., after downstream finetuning/linear probing on regular ImageNet-100) is evaluated on each of the three Noisy-ImageNet-100 datasets, which have been derived by removing from ImageNet-A/P/C any classes not included in ImageNet-100.

The Imbalanced-ImageNet-100 dataset concerns the evaluation of resilience against uneven class distribution, using an imbalanced variant of ImageNet called ImageNet-Long-Tail \cite{openlongtailrecognition}. Thus, for each SSL method the corresponding ImageNet-100-pretrained variant from the main experiments undergoes supervised downstream finetuning on Imbalanced-ImageNet-100, which has been derived by adapting for ImageNet-100 the process applied in \cite{openlongtailrecognition} to ImageNet-1k. 

In order to evaluate the robustness of the learnt visual representations against cross-domain shifts the Visual Task Adaptation Benchmark (VTAB) \cite{zhai2019visual} has been adopted. In particular, for each considered SSL method, the corresponding ImageNet-100-pretrained variant from the main object-centric experiments (Section \ref{sec::MainExperiments}) has been finetuned and tested across the 19 VTAB downstream datasets, using the off-the-self VTAB protocol\footnote{\url{https://github.com/google-research/task_adaptation}}. VTAB evaluates downstream accuracy metrics for three high-level group tasks (each computed based on multiple datasets), namely the `natural', the `specialized' and the `structured' one.

Apart from evaluating the robustness of the considered SSL methods, additional quantitative comparisons include the two baselines used for the main experiments (as detailed in Section \ref{sec::MainExperiments}), namely supervised pretraining on ImageNet-1k and on ImageNet-100. As per relevant protocols, downstream finetuning is subsequently conducted for Imbalanced-ImageNet-100, before evaluation, but not for Noisy-ImageNet-100. Additionally, for the case of the off-the-shelf (non-low-data) pretrained DINOv2 baseline, a linear classification head finetuned on ImageNet-100/Imbalanced-Imagenet-100 is utilized for inference in the Noisy-ImageNet-100/Imbalanced-ImageNet-100 experimental benchmark, respectively.

\subsection{Domain-specific experiments}
\label{sec::DomainSpecificExperiments}

\begin{figure}[t]
\centering
\begin{tabular}{cc}
\includegraphics[width=0.30\linewidth]{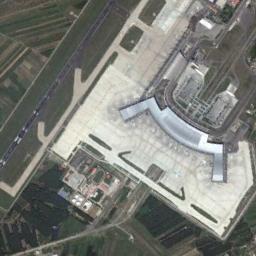}&
\includegraphics[width=0.30\linewidth]{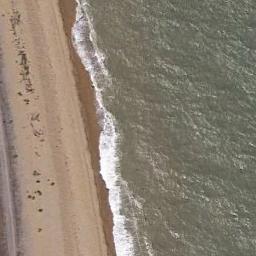}\\
\includegraphics[width=0.30\linewidth]{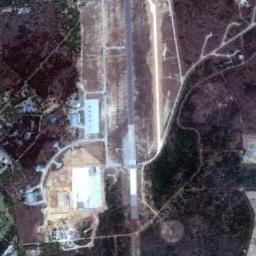}&
\includegraphics[width=0.30\linewidth]{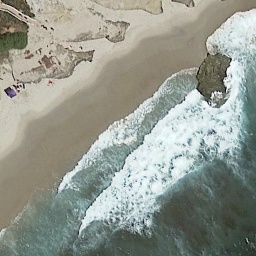}\\
(a)&(b)\\
\end{tabular}
\caption{Example images from the MLRSNet remote sensing dataset \cite{qi2020mlrsnet} belonging to classes a) `Airport' and b) `Beach'}
\label{fig::MLRSNetDataset}
\end{figure}

\begin{figure}[t]
\centering
\begin{tabular}{cc}
\begin{tabular}{c}\includegraphics[width=0.25\linewidth]{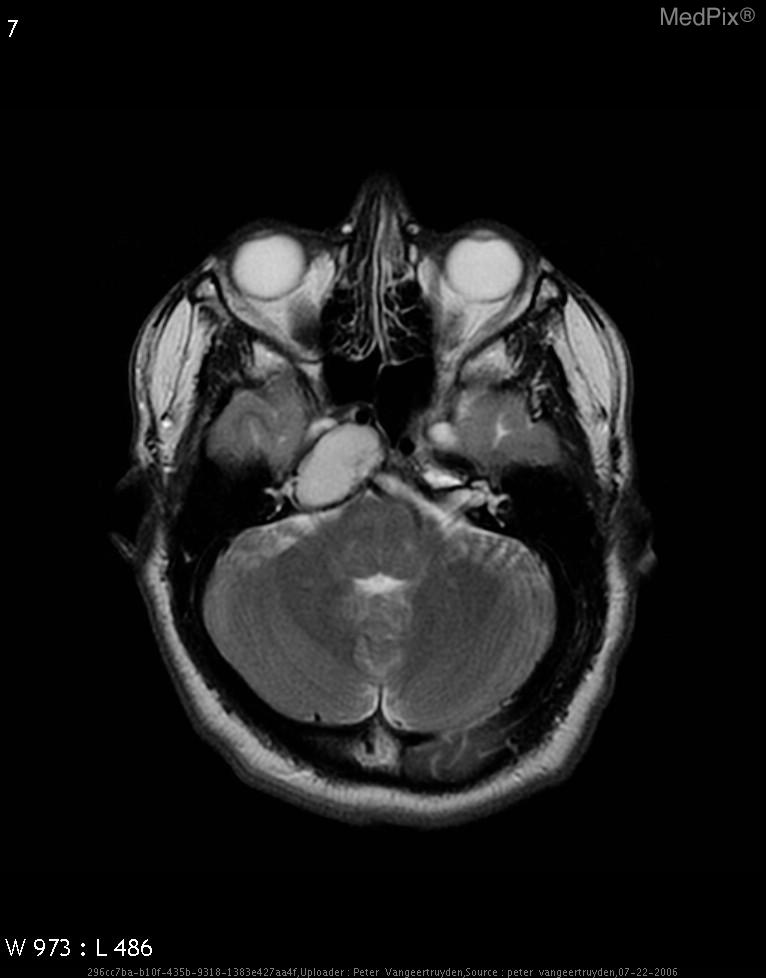}\end{tabular}&
\begin{tabular}{c}\includegraphics[width=0.30\linewidth]{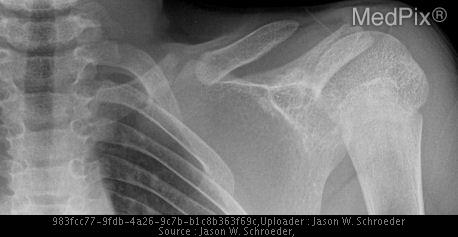}\end{tabular}\\
\begin{tabular}{c}\includegraphics[width=0.25\linewidth]{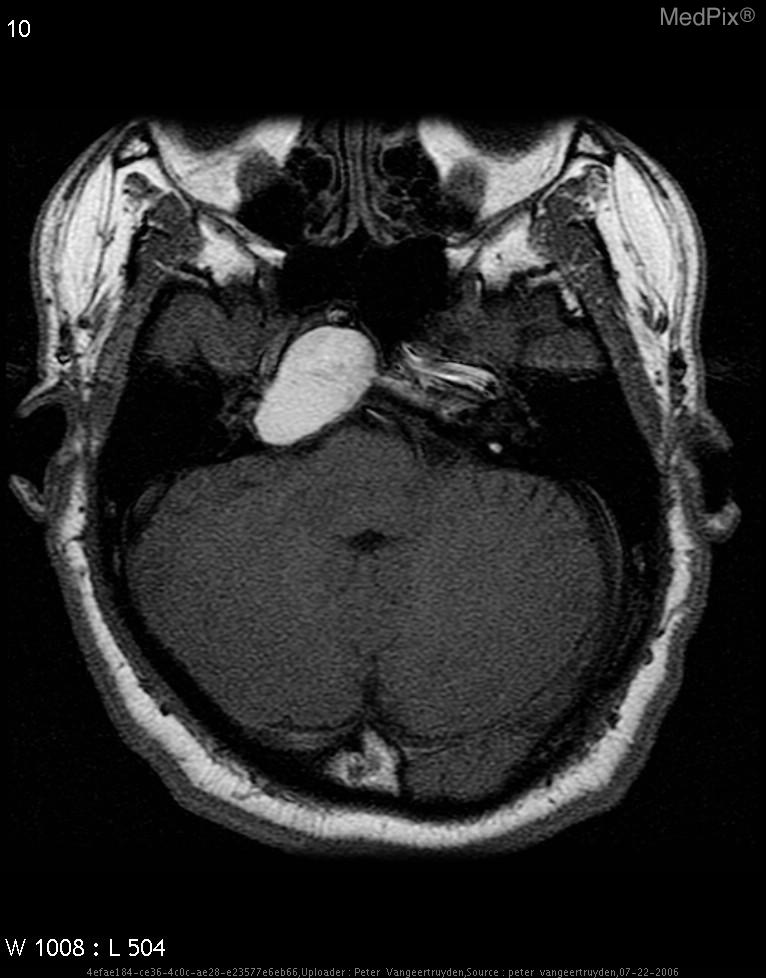}\end{tabular}&
\begin{tabular}{c}\includegraphics[width=0.30\linewidth]{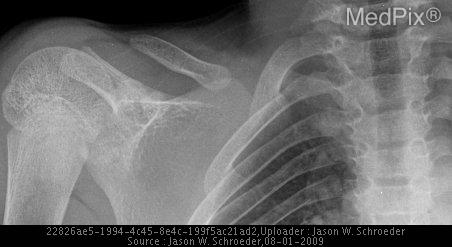}\end{tabular}\\
\begin{tabular}{c}(a)\end{tabular}&\begin{tabular}{c}(b)\end{tabular}\\
\end{tabular}
\caption{Example images from the MedPix medical imaging dataset \cite{MedPixDataset} belonging to classes a) `Cholesterol granuloma' and b) `Cleidocranial Dysostosis, Dysplasia'}
\label{fig::MedPixDataset}
\end{figure}

\begin{figure}[t]
\centering
\begin{tabular}{cc}
\begin{tabular}{c}\includegraphics[width=0.30\linewidth]{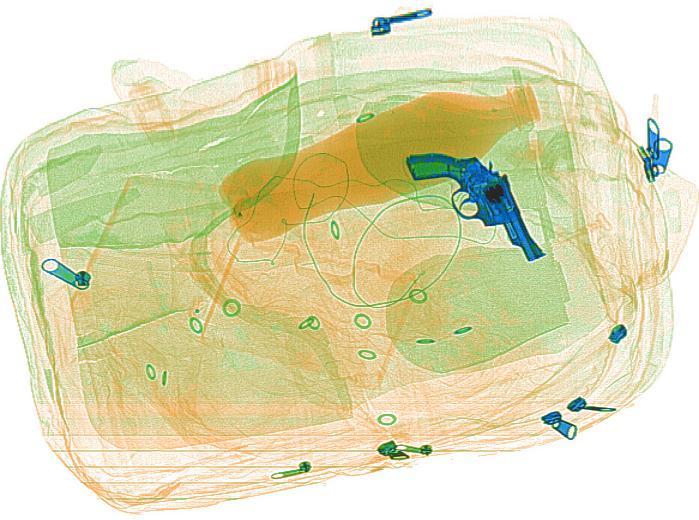}\end{tabular}&
\begin{tabular}{c}\includegraphics[width=0.30\linewidth]{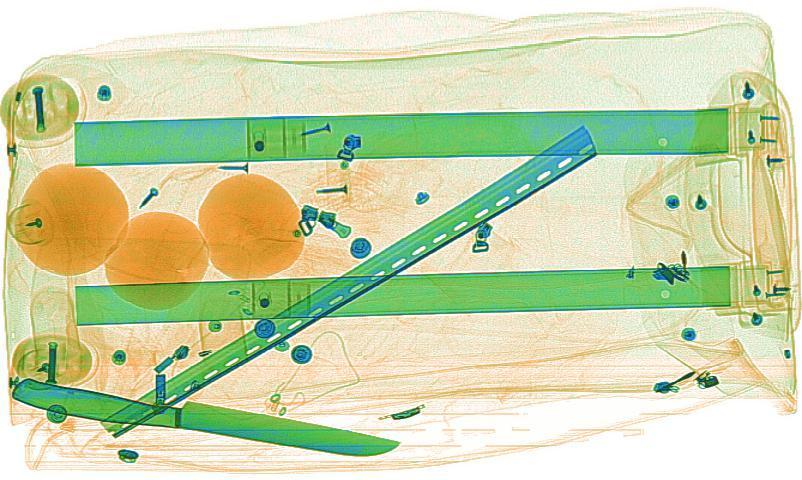}\end{tabular}\\
\begin{tabular}{c}\includegraphics[width=0.30\linewidth]{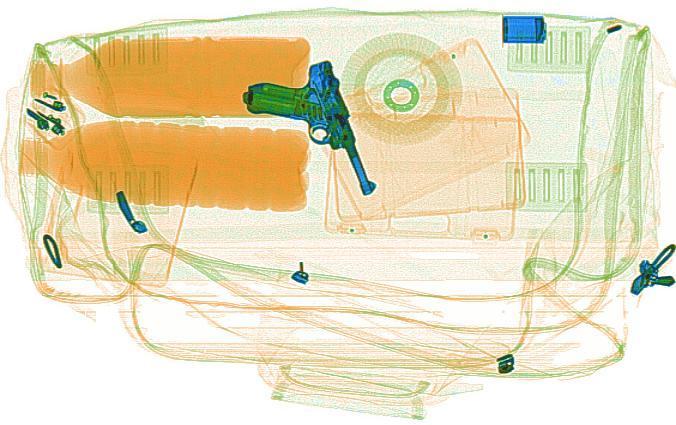}\end{tabular}&
\begin{tabular}{c}\includegraphics[width=0.30\linewidth]{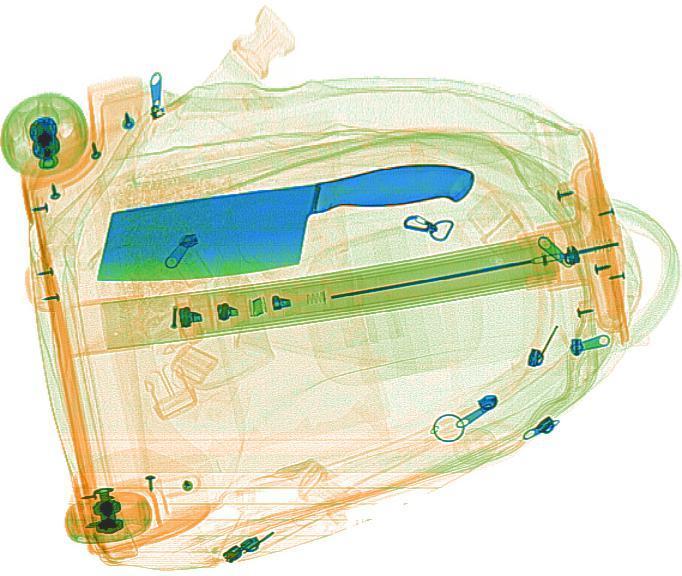}\end{tabular}\\
\begin{tabular}{c}(a)\end{tabular}&\begin{tabular}{c}(b)\end{tabular}\\
\end{tabular}
\caption{Example images from the SIXRay security imaging dataset \cite{Miao2019SIXray} belonging to classes a) `Gun' and b) `Knife'}
\label{fig::SixRay}
\end{figure}

\begin{figure}[t]
\centering
\begin{tabular}{cc}
\begin{tabular}{c}\includegraphics[width=0.25\linewidth]{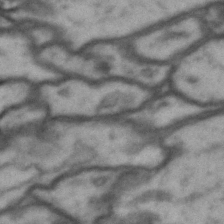}\end{tabular}&
\begin{tabular}{c}\includegraphics[width=0.30\linewidth]{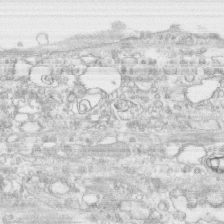}\end{tabular}\\
\begin{tabular}{c}(a)\end{tabular}&\begin{tabular}{c}(b)\end{tabular}\\
\end{tabular}
\caption{Example images from the unlabeled CEM500K microscopy imaging dataset \cite{CEM500K}, depicting a) brain cells of drosophila tissue, and b) brain cells of human tissue}
\label{fig::CEM500K}
\end{figure}

\begin{figure}[t]
\centering
\begin{tabular}{cc}
\begin{tabular}{c}\includegraphics[width=0.25\linewidth]{BIRDSAI_Animals.png}\end{tabular}&
\begin{tabular}{c}\includegraphics[width=0.30\linewidth]{BIRDSAI_Humans.png}\end{tabular}\\
\begin{tabular}{c}(a)\end{tabular}&\begin{tabular}{c}(b)\end{tabular}\\
\end{tabular}
\caption{Example images from the object detection UAV-captured infrared dataset \cite{CEM500K}, depicting objects of classes a) `Animal', and b) `Human'}
\label{fig::BIRDSAI}
\end{figure}

Specialized image domains differing significantly from conventional/natural RBG images constitute a particularly compelling evaluation scenario for SSL, due to the relative scarcity of (sufficiently) large domain-specific datasets and/or ground-truth annotations. In order to evaluate whether under such evaluation settings domain-specific pretraining in the low-data regime outperforms generic low-data pretraining on ImageNet-100 and/or large-scale pretraining on ImageNet-1k, in terms of downstream accuracy, the selected SSL methods are evaluated under the transfer-learning setup in five different (and significantly diverse) visual domains: a) remote sensing (high altitude aerial photographs), b) medical imaging (i.e., chest X-rays), c) security imaging (i.e., airport luggage X-rays), d) microscopy imaging (e.g., blood cells), and e) UAV-captured infrared footage (i.e., aerial thermal images). Evaluation in the first four domains is conducted on a whole-image classification downstream task, while in the UAV-captured infrared domain on an object detection downstream task.

For whole-image classification, the linear probing finetuning protocol classification is employed here as well. For the object detection experiment, the pretrained backbones are combined with the recent real-time object detector YOLOv8 \cite{JocherYOLO82023}. During finetuning, each backbone is kept frozen so that only the neck and detection head modules of YOLOv8 are optimized. In all cases, in-domain SSL pretraining is compared against supervised pretraining on ImageNet-100 and on ImageNet-1k, as well as against an off-the-shelf non-low-data-pretrained DINOv2 baseline.

For the remote sensing domain, MLRSNet \cite{qi2020mlrsnet} and AID \cite{Xia2017AID} have been used as the pretraining and downstream datasets, respectively, while indicative exemplary images are shown in Figure \ref{fig::MLRSNetDataset}. Regarding the medical imaging domain, the MedPix \cite{MedPixDataset} and ChestX-Det \cite{lian2021structure} datasets have been used for pretraining and downstream finetuning, respectively, while example images are shown in Figure \ref{fig::MedPixDataset}. Moreover, for the security imaging domain, two different subsets of the SIXRay \cite{Miao2019SIXray} dataset have been used as the pretraining and downstream subset, namely SIXRay-100 and SIXRay-10, respectively, while indicative examples are shown in Figure \ref{fig::SixRay}. For the microscopy imaging domain, CEM500K \cite{CEM500K} and RaabinWBC \cite{RaabinWBC} have been employed as pretraining and downstream datasets, respectively, while example images are depicted in Figure \ref{fig::CEM500K}. Finally, for the UAV-captured infrared domain, BIRDSAI \cite{BIRDSAI} and HIT-UAV \cite{HITUAV} have been employed as pretraining and downstream datasets, respectively, with indicative images shown in Figure \ref{fig::BIRDSAI}. All domain-specific datasets, utilized either for pretraining or downstream finetuning/evaluation, are low-data in the sense defined in this study (after appropriate subsampling, in the case of CEM500K).

\subsection{Employed datasets}

\begin{longtable}{|c|p{9cm}|}
 \caption{\hl{Image datasets employed in the evaluation study}}
 \label{tab::Datasets} \\
 \hline
    \begin{tabular}{c}Dataset\end{tabular} &
    \begin{tabular}{c}Description\end{tabular} \\
    \hline
 \endfirsthead
 \caption{\hl{Image datasets employed in the evaluation study (continued)}} \\
 \hline
    \begin{tabular}{c}Dataset\end{tabular} &
    \begin{tabular}{c}Description\end{tabular} \\
    \hline
 \endhead
 \hline \multicolumn{2}{r}{\textit{Continued on next page}} \\
 \endfoot
 \hline
 \endlastfoot
    \begin{tabular}{l}ImageNet-1k \cite{imagenet_cvpr09}\end{tabular}&
    \begin{tabular}{p{8.75cm}}It contains 1.3 million natural RGB images, uniformly distributed to 1,000 classes. Each image is assigned only one class label, which is typically object-centric. It is split into training/validation/test sets of 1,281,167/50,000/100,000 images, respectively.\end{tabular}\\
    \hline
    \begin{tabular}{l}ImageNet-100 \cite{zhao2020maintaining}\end{tabular}&
    \begin{tabular}{p{8.75cm}}It is a low-data variant of ImageNet-1k, retaining 100 classes with 1,300 training and 300 test images per class. It is split to training/validation/test sets of 100,000 (1,000 images per class), 30,000 and 30,000 (300 images per class) images, respectively.\end{tabular}\\
    \hline
    \begin{tabular}{l}Places-365 \cite{zhou2017places}\end{tabular}&
    \begin{tabular}{p{8.75cm}}It contains 10 million images distributed across 434 scene-centric classes, having 5,000 to 30,000 training images per class. It is split to training/validation/test sets of 8 million/36,000/328,000 images, respectively.\end{tabular}\\
    \hline
    \begin{tabular}{l}Places-100 \cite{he2021quantifying}\end{tabular}&
    \begin{tabular}{p{8.75cm}}It is a low-data variant of Places-365, derived by randomly selecting 100 classes and 400 images per class from Places-365. It is split to training/test sets of 30,000/10,000 images, respectively.\end{tabular}\\
    \hline
    \begin{tabular}{l}STL-10 \cite{coates2011stl10}\end{tabular}&
    \begin{tabular}{p{8.75cm}}It consists of 500 labeled training images, 800 labeled test images and 100,000 unlabeled images, covering in total 10 classes (namely `Airplane', `Bird', `Cat', `Deer', `Dog', `Horse', `Monkey', `Ship', `Truck' and `Car').\end{tabular}\\ 
    \hline
    \begin{tabular}{l}COCO \cite{lin2014microsoft}\end{tabular}&
    \begin{tabular}{p{8.75cm}}It contains 164,000 images with 80 classes of objects. Its version utilized in this paper is split to training/validation/test sets of 82,000/41,000/41,000 images, respectively.\end{tabular}\\
    \hline
    \begin{tabular}{l}ADE20K \cite{zhou2017scene}\end{tabular}&
    \begin{tabular}{p{8.75cm}}It includes 365 scene classes and contains 25,574/2,000 images for training/test, respectively.\end{tabular}\\
    \hline
    \begin{tabular}{l}SUN Database \cite{xiao2010sun}\end{tabular}&
    \begin{tabular}{p{8.75cm}}The utilized version includes 108,754 scene-centric images falling under 397 classes, with at least 100 images per class. It is split to training/validation/test sets of 76,128/10,875/21,750 images, respectively.\end{tabular}\\
    \hline
    \begin{tabular}{l}ImageNet-A \cite{hendrycks2021natural}\end{tabular}&
    \begin{tabular}{p{8.75cm}}It contains 30,000 images falling under 200 classes of ImageNet-1k, which have been selected based on their ability to fool ResNet-50 classifiers trained on ImageNet-1k. The dataset is typically employed only for inference.\end{tabular}\\
    \hline
    \begin{tabular}{l}ImageNet-P \cite{hendrycks2019benchmarking}\end{tabular}&
    \begin{tabular}{p{8.75cm}}Modification of the ImageNet-1k validation set, so that each image has been slightly perturbed by the application of various transformations (owing to noise, blur, weather conditions and digital distortions). It contains a total of 200,000 images and is typically employed only for inference.\end{tabular}\\
    \hline
    \begin{tabular}{l}ImageNet-C \cite{hendrycks2019benchmarking}\end{tabular}&
    \begin{tabular}{p{8.75cm}}It is a dataset similar to the ImageNet-P one, but with more intense corruptions. It contains a total of 200,000 images and is also typically employed only for inference.\end{tabular}\\
    \hline
    \begin{tabular}{l}ImageNet-\\Long-Tail \cite{openlongtailrecognition}\end{tabular}&
    \begin{tabular}{p{8.75cm}}It has been obtained from ImageNet-1k by sampling using a long tail distribution, so that the resulting classes are imbalanced. It contains 115,846 images, split to training/validation/test sets of 81,092/17,377/17,377 images, respectively.\end{tabular}\\
    \hline
    \begin{tabular}{l}MLRSNet \cite{qi2020mlrsnet}\end{tabular}&
    \begin{tabular}{p{8.75cm}}It is designed for multi-label remote sensing image classification. It contains 100,000 images, each annotated with multiple labels from a set of 60 classes. It is split to training/validation/test sets of 80,000/10,000/10,000 images, respectively.\end{tabular}\\
    \hline
    \begin{tabular}{l}AID \cite{Xia2017AID}\end{tabular}&
    \begin{tabular}{p{8.75cm}}It is a collection of high-resolution aerial images for various land-use and land-cover classification tasks. It is split to training/validation/test sets of 8,000/1,000/1,000 images, respectively.\end{tabular}\\
    \hline
    \begin{tabular}{l}MedPix \cite{MedPixDataset}\end{tabular}&
    \begin{tabular}{p{8.75cm}}It contains 44,000 medical images with annotations for 10 different types of organs and tissues. It includes 32,000 training images, 4,000 validation images, and 8,000 test images.\end{tabular}\\
    \hline
    \begin{tabular}{l}ChestX-Det \cite{lian2021structure}\end{tabular}&
    \begin{tabular}{p{8.75cm}}ChestX-Det is an expanded version of the ChestX-Det10 dataset \cite{liu2020chestx}, comprising 3578 images sourced from NIH ChestX-14. The dataset covers 13 common categories of chest-related diseases and abnormalities. It is divided to 2,077 training, 756 validation and 756 test images.\end{tabular}\\
    \hline
    \begin{tabular}{l}SIXRay \cite{Miao2019SIXray}\end{tabular}&
    \begin{tabular}{p{8.75cm}}It contains 1,059,23 X-ray scan images of airport luggage, with a subset of them depicting prohibited items. Different overlapping subsets have been defined, each with a varying ratio of negative-to-positive images. In this study, two such subsets are utilized: SIXRay-100, containing 742,104/131,769 training/test images, respectively, and SIXRay-10, containing 67,464/11,979 training/test images, respectively.\end{tabular}\\
    \hline
    \begin{tabular}{l}CEM500K \cite{CEM500K}\end{tabular}&
    \begin{tabular}{p{8.75cm}}It comprises 496,544 cellular electron microscopy images. The main dataset is not annotated, since it was constructed for enabling SSL pretraining.\end{tabular}\\
    \hline
    \begin{tabular}{l}RaabinWBC \cite{RaabinWBC}\end{tabular}&
    \begin{tabular}{p{8.75cm}}It contains 16,633 microscopy images of 5 classes of white blood cells, annotated for whole-image classification. The training/validation/test subset contains 10,185/4,339/2,119 images, respectively.\end{tabular}\\
    \hline
    \begin{tabular}{l}BIRDSAI \cite{BIRDSAI}\end{tabular}&
    \begin{tabular}{p{8.75cm}}It contains 61,994 real and 101,542 synthetic infrared images captured from a Unmanned Aerial Vehicle (UAV) view, during night-time surveillance of protected areas in Africa. The real images have been divided into two distinct sets: a training set comprising 40,661 images and a test set comprising 21,333 images. All synthetic images have been incorporated into the training set. It is annotated for object detection using 2 classes (`Animals', `Humans').\end{tabular}\\
    \hline
    \begin{tabular}{l}HIT-UAV \cite{HITUAV}\end{tabular}&
    \begin{tabular}{p{8.75cm}}It contains 2,898 thermal infrared images extracted from UAV video footage in urban settings. It is divided into a training/validation/test subset, of 2,029/290/579 images, respectively, and is annotated for object detection using 4 classes (`Person', `Car', `Bicycle', `Other Vehicle').\end{tabular}\\
    \hline
\end{longtable}

A summary of the datasets employed in the conducted experimental study can be found in Table \ref{tab::Datasets}, excluding the 19 VTAB ones used for the robustness experiments (Section \ref{sec::RobustnessExperiments}); details about them are available in \cite{zhai2019visual}. In certain cases, adjustments needed to be made for the data to be incorporated in the experimental evaluation, mainly targeting to fit to the defined `low-data regime'. In particular, Places-100 was constructed from Places-365 by randomly sampling 500 images per class, instead of 400 as in \cite{he2021quantifying}, for a total of 50,000 images (instead of 40,000) that were then segmented into a training/validation/test set following a 70-15-15 split ratio\footnote{In the sequel, this modified variant is implied wherever Places-100 is mentioned}. For the case of ADE20K, whose official on-line repository contains only a training set and a test set, the former one was randomly split into a training and a validation subset using a 80:20 ratio. Similarly, the labeled images of STL-10 were grouped into 10 predefined folds, each comprising 50 images from the total of 500 available annotated ones; then, one fold was utilized as the validation set and the remaining 9 formed the training one. Moreover, the SIXRay-100 dataset was resized to better fit the low-data regime, by constructing a subset of 300,000 images in the following way: all 8,929 positive images were retained and the negative ones were randomly sampled from the original negative images of the dataset; the formulated dataset was segmented following a 70-15-15 split ratio, resulting in a training/validation/test set of 210,000/45,000/45,000 images, respectively\footnote{In the sequel, this modified low-data variant is implied wherever SIXRay-100 is mentioned}. Overall, SIXRay-10 and the subsampled SIXRay-100 overlap in content, but the test images of SIXRay-10 are deliberately not included in the training or the validation set of SIXRay-100. Finally, the variant of the CEM500K dataset utilized in this study has been derived after random subsampling to 300,000 images, so that it falls under the defined low-data range, \hl{and segmentation into a training and a validation subset using a 80:20 ratio.}

Unless otherwise specified, the following considerations hold in the conducted experiments: a) the training set of the pretraining dataset is employed for both SSL pretraining and supervised pretraining variants, b) in single-dataset experiments, the training set of the pretraining dataset is subsequently also utilized for supervised downstream finetuning, taking into account ground-truth annotation labels, c) each training session exploits an early stopping criterion based on validation loss, to automatically adjust the number of epochs, d) in all experiments except those falling under a single-dataset setup, the training set of each downstream dataset is utilized for supervised downstream finetuning, taking into account ground-truth annotation labels, and e) the annotated test set of each downstream dataset is used for evaluation purposes, by inferring predictions using the trained DNN.




\section{Experimental Results and Insights}
\label{sec::Experiments}
\begin{table*}[!htbp]
\caption{Top-5 accuracy in downstream classification for single-dataset SSL setup}
\centering
\label{tab::Main-Single}
\begin{tabular}{|l|c|c|}
\hline
Method & $M^O_S$(IN-100$\downarrow$IN-100) & $M^S_S$(P-100$\downarrow$P-100)\\
\hline
SimCLR (ResNet) & \textbf{73.5}\% & 69.8\% \\
DINO (ViT) & 72.1\% & 68.5\% \\
MAE (ViT) & 72.7\% & 69.1\% \\
DeepClusterV2 (ResNet) & 70.3\% & 66.8\% \\
\hdashline
None/Random (ResNet) & 70.6\% & 70.3\% \\
None/Random (ViT) & 71.2\% & \textbf{70.9}\% \\
\hline
\end{tabular}
\end{table*}

\begin{table*}[!ht]
\caption{\hl{Top-5 accuracy in downstream classification for transfer-learning SSL setup (top-1 accuracy reported for STL-10 dataset). The `Sup.-Large' and DINOv2 baselines have been pretrained on large-scale datasets.}}
\centering
\small
\setlength{\tabcolsep}{4pt}
\resizebox{\textwidth}{!}{
\begin{tabular}{|l||c|c|c|c|}
\hline
Method&$M^O_T$(IN-100$\downarrow$COCO)&$M^O_T$(IN-100$\downarrow$STL-10)&$M^S_T$(P-100$\downarrow$ADE)&$M^S_T$(P-100$\downarrow$SUN DB)\\
\hline
SimCLR (ResNet) & 70.8\% & 64.2\% & 49.4\% & 51.7\% \\
DINO (ViT) & 75.7\% & 67.1\% & 53.1\% & 55.2\% \\
MAE (ViT) & 74.4\% & 65.7\% & 51.1\% & 53.3\% \\
DeepClusterV2 (ResNet) & 67.0\% & 59.3\% & 46.2\% & 48.1\% \\
\hdashline
Sup. (ResNet) & 73.2\% & 67.8\% & 54.3\% & 53.5\% \\
Sup. (ViT) & 74.9\% & 69.7\% & 56.2\% & 55.8\% \\
Sup.-Large (ResNet) & 79.1\% & 73.6\% & 57.4\% & 59.4\% \\
Sup.-Large (ViT) & \textbf{80.4}\% & 75.4\% & 58.9\% &  \textbf{61.2}\%\\
DINOv2 (ViT) & 79.7\% & \textbf{76.1}\% & \textbf{59.3}\% & 60.5\% \\
\hline
\end{tabular}
\label{tab::Main-Transfer}
}
\end{table*}

\begin{table*}
\caption{\hl{Top-5 accuracy in downstream classification for semi-supervised single-dataset setup}}
\centering
\small
\setlength{\tabcolsep}{4pt}
\resizebox{\textwidth}{!}{
\begin{tabular}{|l||c|c|c|c|}
\hline
&$M^O_S$(IN-100$\downarrow$IN-100)
&$M^O_S$(IN-100$\downarrow$IN-100)&$M^S_S$(P-100$\downarrow$P-100)&$M^S_S$(P-100$\downarrow$P-100)\\
\cline{2-5}
Method& 5\% & 10\% & 5\% & 10\% \\
\hline
\hline
SimCLR (ResNet) & 59.7\% & 66.4\% & 55.2\% & 58.3\% \\
DINO (ViT) & 63.2\% & \textbf{70.6}\% & 59.3\% & 62.6\% \\
MAE (ViT) & 61.3\% & 67.8\% & 57.0\% & 60.4\% \\
DeepClusterV2 (ResNet) & 58.1\% & 64.5\% & 51.5\% & 54.6\% \\
\hdashline
None/Random (ResNet) & 64.0\% & 69.3\% & 60.2\% & 63.5\% \\
None/Random (ViT) & \textbf{64.7}\% &  69.8\% & \textbf{61.3}\%  & \textbf{64.4}\% \\
\hline
\end{tabular}
\label{tab::SemiSupervised}
}
\end{table*}

This section presents the evaluation results for each of the three experimental sets described in Section \ref{sec::Methodology}, along with visual/qualitative and statistical analyses of the image representations learnt by the various SSL methods. Top-5 classification accuracy is the employed evaluation metric, excluding specific downstream tasks with a particularly low number of classes; top-1 accuracy is reported in such cases.

\subsection{Main experimental results}
\label{sec::MainExperimentalResults}
This section presents in detail the comparative evaluation results obtained from the application of the experimental setting described in Section \ref{sec::MainExperiments}. In particular, Table \ref{tab::Main-Single} presents the obtained results for the single-dataset setup for both object- and scene-centric cases, i.e. experiments $M^O_S$(IN-100$\downarrow$IN-100) and $M^S_S$(P-100$\downarrow$P-100), while also including as baselines conventional DNNs trained in a typical supervised setting with random initialization. Similarly, Table \ref{tab::Main-Transfer} illustrates the results for the transfer-learning setup, i.e., experiments $M^O_T$(IN-100$\downarrow$COCO), $M^O_T$(IN-100$\downarrow$STL-10), $M^S_T$(P-100$\downarrow$ADE) and
$M^S_T$(P-100$\downarrow$SUN). Regarding the baselines, it must be noted that the `Sup.-Large' and DINOv2 models have been pretrained in large-scale datasets (ImageNet-1k/Places-205 for object-centric/scene-centric `Sup.-Large' baselines, respectively, and LVD-142M for DINOv2). Moreover, Table \ref{tab::SemiSupervised} provides the results of the single-dataset experiments, using the semi-supervised downstream finetuning protocol.

\underline{Single-dataset setup}: From the results presented in Table \ref{tab::Main-Single}, \textbf{SimCLR is shown to be the best performing SSL method}, despite being evaluated using a ResNet-50 architecture. In particular, in the object-centric setting ($M^O_S$(IN-100$\downarrow$IN-100)) it achieves the overall best accuracy, while in the scene-centric case ($M^S_S$(P-100$\downarrow$P-100)) is slightly outperformed by the baselines (trained from scratch with random parameter initialization). This result is potentially a side-effect of the fact that the pretraining dataset is more than double in size for the object-centric case (ImageNet-100), compared to that of the scene-centric one (Places-100). This finding suggests that there may exist a pretraining dataset size limit (within the range of 35k-100k images), below which SSL is indeed ineffective for representation learning. To verify this insight, an additional experiment has been performed in the $M^O_S$(IN-100$\downarrow$IN-100) setting: ResNet-50 was separately trained via the best method (SimCLR pretraining + downstream finetuning) and via the random initialization baseline (direct training) on 3 randomly sampled subsets of ImageNet-100, containing 80\%, 65\% and 50\% of its training images, respectively (all classes are retained in all cases). The resulting SimCLR downstream accuracy falls from 73.5\% (for the full ImageNet-100) to 70.2\%, 67.7\% and 63.9\%, respectively. Correspondingly, the baseline's accuracy falls from 70.6\% to 69.4\%, 67.4\% and 64.8\%, respectively. This is graphically depicted in Fig. \ref{fig::IN100SizeScaling}. It indeed validates the proposed hypothesis and localizes the hypothesized size limit between 50k and 65k training images.

\begin{figure}[h!]
    \centering
    \begin{tikzpicture}
        \begin{axis}[
            width=12cm, 
            height=8cm, 
            xlabel={IN-100 Sample Size (\%)},
            ylabel={Downstream Accuracy (\%)},
            xmin=45, xmax=105, 
            ymin=60, ymax=75, 
            xtick={50, 65, 80, 100},
            ytick={60, 65, 70, 75},
            legend pos=south west,
            grid=both,
            grid style={dotted, gray},
            legend style={font=\small},
            ymajorgrids=true,
            xmajorgrids=true,
            line width=1pt,
            mark options={scale=1.2},
        ]
        
        \addplot[
            color=blue,
            mark=*,
            ]
            coordinates {
                (100, 73.5)
                (80, 70.2)
                (65, 67.7)
                (50, 63.9)
            };
        \addlegendentry{SimCLR (ResNet)}
        
        \addplot[
            color=red,
            mark=square*,
            ]
            coordinates {
                (100, 70.6)
                (80, 69.4)
                (65, 67.4)
                (50, 64.8)
            };
        \addlegendentry{None/Random (ResNet)}

        \end{axis}
    \end{tikzpicture}
    \caption{\hl{Top-5 accuracy in single-dataset downstream classification as a function of IN-100 sample size}}
    \label{fig::IN100SizeScaling}
\end{figure}

An additional crucial finding is the observed surprising \textbf{dominance of SimCLR/ResNet-50 over DINO/ViT-L/16}. Given that in the examined single-dataset setup the absolute priority is not for computing representations that generalize well to different datasets, this may potentially be a side-effect of the more data-hungry nature of Vision Transformers compared to CNNs \cite{xu2021vitae}, which potentially limits their applicability in the low-data regime. Since DINO's powerful representation learning abilities are mostly evident in combination with ViT architectures, the aforementioned observation indicates that its value is limited in the low-data single-dataset setup.

\underline{Transfer-learning setup}: Examining the results illustrated in Table \ref{tab::Main-Transfer}, it can be seen that in the transfer-learning setup \textbf{DINO proves to be the best-performing SSL method}. Another critical observation though is that \textbf{baseline approaches (supervised pretraining and large-scale-pretrained DINOv2) outpass all SSL methods}. In fact, with the exception of the COCO dataset, even low-data supervised pretraining achieves higher accuracy than DINO. Nevertheless, the obtained results indicate that DINO models more general image representations compared to competing SSL approaches in the low-data regime.

\underline{Semi-supervised single-dataset setup}: From the results presented in Table \ref{tab::SemiSupervised}, it can be observed that \textbf{baseline approaches (trained from scratch with random parameter initialization) in general outperform SSL methods in the low-data semi-supervised single-dataset setting}, with the exception being DINO that performs the best in the object-centric scenario and when using 10\% of the ground-truth class labels. This finding mostly confirms the results of the transfer-learning setting. In the semi-supervised evaluation setup, the ground-truth labels utilized for downstream finetuning contain a significant amount of noise and, thus, convey a more limited amount of useful knowledge about the task. In such a scenario, where there is no transfer but the inherently discriminant capability of learnt representations is important, \textbf{DINO outperforms all SSL competitors}. This reinforces the conclusion that, up to a degree, DINO retains its ability to learn more generally applicable embeddings, even under low-data pretraining conditions.

\begin{table}
\caption{Top-5 accuracy in downstream classification for Noisy-ImageNet-100 robustness protocol evaluation}
\label{tab::Robustness-Noisy}
\centering
\begin{tabular}{|c||c|c|c|}
\hline
&\multicolumn{3}{c|}{Datasets}\\
\cline{2-4}
Method & IN-A-100 & IN-P-100 & IN-C-100 \\
\hline
\hline
SimCLR (ResNet) & 70.3\% & 75.1\% & 68.3\% \\
DINO (ViT) & 73.0\% & 78.2\% & 72.6\% \\
MAE (ViT) & 74.6\% & 76.5\% & 70.4\% \\
DeepClusterV2 (ResNet) & 66.4\% & 68.7\% & 64.2\% \\
\hdashline
Sup. (ResNet) & 77.2\% & 78.0\% & 73.5\% \\
Sup.-Large (ResNet) & 81.8\% & 83.0\% & 76.5\% \\
Sup. (ViT) & 78.3\%  &78.5\% & 74.0\% \\
Sup.-Large (ViT) & \textbf{82.5}\% & 84.2\% & 77.6\% \\
DINOv2 (ViT) & 81.8 \% & \textbf{84.7} \% & \textbf{78.3}\% \\
\hline
\end{tabular}
\end{table}

\begin{table*}
\caption{\hl{Accuracy in downstream classification for Imbalanced-ImageNet-100 (top-5) and VTAB (top-1) robustness protocols evaluation}}
\centering
\small
\setlength{\tabcolsep}{4pt}
\resizebox{\textwidth}{!}{
\begin{tabular}{|c||c|c|c|c|}
\hline
&\multicolumn{4}{c|}{Datasets}\\
\cline{2-5}
Method & Imbalanced-IN-100 & VTAB-Natural & VTAB-Specialized & VTAB-Structured \\
\hline
SimCLR (ResNet) & 63.2\% & 62.5\% & 58.3\% & 60.1\% \\
DINO (ViT) & 79.1\% & 65.2\% & 61.8\% & 63.0\% \\
MAE (ViT) & 69.3\% & 66.7\% & 62.9\% & 64.5\% \\
DeepClusterV2 (ResNet) & 59.1\% & 60.3\% & 56.7\% & 58.2\% \\
\hdashline
Sup. (ResNet) & 78.2\% & 63.2\% & 62.8\% & 62.5\%\\
Sup.-Large (ResNet) & 85.3\% & 66.9\% & 63.4\% & 65.8\% \\
Sup. (ViT) & 80.0\% & 67.1\% & \textbf{64.7}\% & 66.0\% \\
Sup.-Large (ViT) & 86.2\% & 68.2\% & 64.2\% & 66.5\% \\
DINOv2 (ViT) & \textbf{86.8}\% & \textbf{68.7}\% & 64.5\% & \textbf{66.8}\% \\
\hline
\end{tabular}
\label{tab::Robustness-Combined}
}
\end{table*}

\subsection{Robustness experimental results}
This section discusses the outcomes obtained by the execution of the robustness evaluation protocols detailed in Section \ref{sec::RobustnessExperiments}. In particular, Tables \ref{tab::Robustness-Noisy} and \ref{tab::Robustness-Combined} illustrate downstream evaluation results for the Noisy-ImageNet-100, and the Imbalanced-ImageNet-100 and VTAB benchmarks, respectively. It must be highlighted that no downstream finetuning on Noisy-ImageNet-100 is performed, according to the relevant protocols. To this end, all SSL-pretrained models have been finetuned on regular ImageNet-100 and then evaluated on the three Noisy-ImageNet-100 datasets (in Table \ref{tab::Robustness-Noisy}). On the other hand, the non-SSL baselines were trained from scratch on ImageNet-100 (`Sup.') or ImageNet-1k (`Sup.-Large') and then evaluated on Noisy-ImageNet-100. The DINOv2 baseline was pretrained on LVD-142M and then evaluated on Noisy-ImageNet-100, after appending to it a linear classifier finetuned on ImageNet-100. For the Imbalanced-ImageNet-100 benchmark (Table \ref{tab::Robustness-Combined}), the SSL methods and the basic supervised baselines were pretrained on ImageNet-100 and then finetuned on Imbalanced-ImageNet-100, before evaluation on the latter one's test set. The `Sup.-Large' baselines were pretrained on ImageNet-1k, while the off-the-shelf DINOv2 baseline was pretrained on LVD-142M. Moreover, the robustness evaluation protocol for VTAB (Table \ref{tab::Robustness-Combined}) is similar to the one for Imbalanced-ImageNet-100.

From the results presented in Tables \ref{tab::Robustness-Noisy} and \ref{tab::Robustness-Combined}, it can be observed that \textbf{MAE and DINO (both on ViT architectures) are proven to be the most robust low-data SSL approaches}. However, \textbf{no low-data SSL method is able to surpass in performance DINOv2 or the supervised pretraining/training from scratch baselines}; not even the low-data baselines. The latter observation suggests that SSL pretraining in the low-data regime seems to be at a disadvantage with respect to robustness, but the reason is likely different among the three sets of experiments. In particular, the low-data supervised baselines are significantly less robust than their large-scale counterparts in the case of Noisy-ImageNet-100 and Imbalanced-ImageNet-100, but this is not true for VTAB. The particular characteristic of VTAB is that it uses a transfer learning setup with very few downstream finetuning examples, for evaluating how generally applicable different learnt representations are. This differs from both the transfer-learning setup, where there is a reasonably-sized downstream finetuning dataset available, and from the semi-supervised one, where the finetuning dataset simply has noisy ground-truth annotation and no transfer learning takes place. In this challenging scenario, MAE exhibits an advantage over DINO, under low-data pretraining conditions. Given that both methods are evaluated on ViT architectures, this finding suggests that MAE simply needs less pretraining data than DINO, in order to learn generally applicable features that are inherently semantically discriminant. It is likely that this behaviour, which is compatible with the findings of \cite{Kong2023}, gets concealed in the transfer/semi-supervised-learning setup due to the larger finetuning dataset and the lack of transfer, respectively. Still, the low-data regime prevents even MAE from surpassing the supervised pretraining baselines.

\subsection{Domain-specific experimental results}
\label{sec::DomainSpecificExperimentalResults}

\begin{table}
\caption{Top-1 accuracy in downstream classification for domain-specific experiments}
\label{tab::Domain-Specific}
\centering
\begin{tabular}{|c||c|c|c|c|}
\hline
&\multicolumn{4}{c|}{Datasets}\\
\cline{2-5}
Method & AID & ChestX-Det & SIXRay-10 & RaabinWBC\\
\hline\hline
SimCLR (ResNet) & 58.1\% & 68.2\% & 73.3\% & 62.4\%\\
DINO (ViT) & 65.7\% & \textbf{76.4}\% & \textbf{74.1}\% & 80.5\%\\
MAE (ViT) & 63.5\% & 72.1\% & 69.7\% & 78.2\%\\
DeepClusterV2 (ResNet) & 57.4\% & 70.6\% & 66.3\% & 59.3\%\\
\hdashline
Sup. (IN-100, ResNet) & 60.2\% & 70.3\% &  68.4\% & 52.6\%\\
Sup.-Large (IN-1k, ResNet) & 66.2\% & 72.5\% & 70.2\% & 42.4\%\\
Sup. (IN-100, ViT) & 65.1\% & 73.6\% & 72.5\% & 39.4\%\\
Sup.-Large (IN-1k, ViT) & 68.7\% & 74.5\% & 73.9\% & 34.5\%\\
DINOv2 (LVD-142M, ViT) & \textbf{69.4} \% & 74.8\% & 73.5\%  & \textbf{85.0}\%\\
\hline
\end{tabular}
\end{table}

\begin{table}
\caption{Mean Average Precision in downstream object detection for the UAV-captured infrared domain-specific experiments, using the HIT-UAV downstream dataset}
\label{tab::Domain-Specific-ObjectDetection}
\centering
\begin{tabular}{|c||c|c|}
\hline
&\multicolumn{2}{c|}{Metrics}\\
\cline{2-3}
Method & mAP50 & mAP50:95 \\
\hline\hline
SimCLR (ResNet) & 0.804 & 0.497 \\
DINO (ViT) & \textbf{0.854} & \textbf{0.529} \\
MAE (ViT) & 0.837 & 0.518 \\
DeepClusterV2 (ResNet) & 0.812 & 0.500 \\
\hdashline
Sup. (IN-100, ResNet) & 0.785 & 0.457 \\
Sup.-Large (IN-1k, ResNet) & 0.760 & 0.456 \\
Sup. (IN-100, ViT) & 0.802 & 0.497 \\
Sup.-Large (IN-1k, ViT) & 0.810 & 0.502 \\
DINOv2 (LVD-142M, ViT) & 0.818 & 0.507 \\
\hline
\end{tabular}
\end{table}

This section presents in detail comparative evaluation results in domain-specific application cases, as detailed in Section \ref{sec::DomainSpecificExperiments}. In particular, Table \ref{tab::Domain-Specific} illustrates downstream classification results for the AID (remote sensing), ChestX-Det (medical imaging), SIXRay-10 (security imaging) and RaabinWBC (microscopy imaging) datasets. AID is composed of aerial-view RGB images, ChestX-Det and SIXRay-10 comprise X-ray scans, while RaabinWBC contains microscope images of white blood cells. It needs to be highlighted that the security imaging setting exhibits particular key characteristics: a) the pretraining and downstream datasets are overlapping in content, and b) there is extreme imbalance in the size of the supported classes (e.g., the `Negative' class dominates the dataset, while the `Hammer' one contains only 60 images). To this end, the security imaging experiment practically lies in-between the single-dataset and transfer-learning scenarios, while the downstream task is highly challenging, due to the increased class imbalance. Regarding experimental details, the MLRSNet, MedPix, SIXRay-100 and CEM500K datasets were used for SSL pretraining in the corresponding domains. Finally, Table \ref{tab::Domain-Specific-ObjectDetection} depicts downstream object detection results for the HIT-UAV dataset (UAV-captured infrared domain), after pretraining on the BIRDSAI dataset. In all domain-specific experiments, both for whole-image classification and for object detection, SSL-pretrained models are compared against baselines pretrained in a supervised manner on ImageNet-100/ImageNet-1k, with the exception of DINOv2 that was pretrained using the LVD-142M dataset.

From the experimental results, it can be observed that \textbf{in the remote sensing domain (AID dataset) the large-scale pretraining baselines in general perform better than the low-data SSL approaches}: pretrained DINOv2 and supervised ImageNet-1k pretraining lead to the highest performance. This suggests that, even though the remote sensing domain deviates from the one of natural images, the fact that it still remains in the RGB space leads the supervised pretraining methods to surpass the low-data SSL methods (of similar backbone architecture). However, \textbf{in-domain low-data DINO pretraining does outperform low-data supervised pretraining on ImageNet-100}. On the contrary, \textbf{in the medical, security and UAV-captured infrared imaging domains in-domain low-data DINO pretraining outperforms all other approaches}, including the large-scale pretraining ones. This is considered to stem from the significant irrelevance of (conventional) RGB supervised pretraining under a shift to the target domain. Moreover, the aforementioned difference in performance supports the intuition that \textbf{in-domain low-data SSL pretraining is advantageous in highly domain-specific scenarios, compared to large-scale supervised pretraining schemes in conventional RGB settings}. It is notable to mention that, given the extreme class imbalance of SIXRay, SSL methods are able to learn generic features from all images, including those of the dominant `Negative' class, even under a low-data pretraining setting. In contrast, large-scale supervised pretraining on ImageNet-1k does not lead to modeling better representations (most likely due to feature space discrepancy) and the potential of the subsequent supervised finetuning is limited by the class imbalance issue. Finally, \textbf{in the microscopy imaging domain, in-domain low-data DINO pretraining outperforms all other approaches, including the large-scale pretraining ones, except zero-shot DINOv2}. A potential reason for this is that the gigantic LVD-142M pretraining dataset may indeed contain microscopy images, but since it has not been publicly released this remains a conjecture.

\subsection{Explanation of learnt representations}
In order to provide detailed insights and to shed more light on the actual knowledge patterns modeled by each SSL method, qualitative and statistical explanations are estimated for the SSL-learnt representations in the low-data regime. In particular, the silhouette coefficient \cite{Rousseeuw1987silhouettes}, one of the most simple, robust and well-performing cluster validity indices \cite{Arbelaitz2013}, and the RELAX approach \cite{wickstrom2023relax} (`REpresentation LeArning eXplainability'), a recently proposed method that generates attribution-based explanations of representations and models their uncertainty, are used.

With respect to the silhouette coefficient, when clustering $N$ data points into $K$ clusters, coefficient $S_i \in \mathbb{R}$, $S_i \in [-1, 1]$, of the $i$-th data point indicates the compactness of its assigned cluster and its separation from other clusters. The optimal clustering is considered the one where coefficient $S_i$ approaches $1$ for each data point $i$, while a negative silhouette coefficient implies problematic assignment of the $i$-th data point. The average over all data points ($S = \frac{1}{N}\sum_{i=1}^N S_i$) characterizes the quality of the overall dataset clustering. In order to estimate clustering results more robust to the presence of outliers, one can calculate the median $S_m$ value instead. In order to compute $S_m$ when evaluating inferred image representations on downstream image classification tasks, $K$ can be set to the number of supported classes and the representation vectors of all test images belonging to the same ground-truth class can be considered as virtually having an identical cluster assignment. A high $S_m$ value implies globular class representations that are both well separated and significantly compact in the latent space.

\begin{table*}
\caption{\hl{Median silhouette coefficients for object-centric downstream scenarios}}
\label{tab:silhouette_scores_imagenet_100}
\centering
\small
\setlength{\tabcolsep}{4pt}
\resizebox{\textwidth}{!}{
\begin{tabular}{|c||c|c|c|}
\hline
Method & $M^O_S$(IN-100$\downarrow$IN-100) & $M^O_T$(IN-100$\downarrow$COCO) & $M^O_T$(IN-100$\downarrow$STL-10) \\
\hline\hline
SimCLR (ResNet) & 0.53 &0.40  & 0.42  \\
DINO (ViT) & 0.52 & 0.45 & 0.49  \\
MAE (ViT) &  0.49& 0.38 & 0.36 \\
DeepClusterV2 (ResNet) & 0.40 &0.34  &  0.31 \\
\hdashline
Sup. (ResNet)  & 0.51 & 0.48 &0.46  \\
Sup. (ViT) & 0.54 &0.50  &0.50  \\
Sup.-Large (ResNet) & 0.51  &  0.50 & 0.48\\
Sup.-Large (ViT) & 0.54 &0.52 & 0.53 \\
DINOv2 (ViT) & \textbf{0.57} & \textbf{0.54} & \textbf{0.55}\\
\hline
\end{tabular}
}
\end{table*}

\begin{table*}[!ht]
\caption{Median silhouette coefficients for scene-centric downstream scenarios}
\label{tab:silhouette_scores_places_100}
\centering
\begin{tabular}{|c||c|c|c|}
\hline
Method & $M^S_S$(P-100$\downarrow$P-100) & $M^S_T$(P-100$\downarrow$ADE) & $M^S_T$(P-100$\downarrow$SUN)\\
\hline\hline
SimCLR (ResNet) & 0.47 & 0.39 & 0.41 \\
DINO (ViT) & 0.46 & 0.45 & 0.42 \\
MAE (ViT) & 0.43 & 0.38 &  0.36 \\
DeepClusterV2 (ResNet) & 0.40 &0.34  &  0.31  \\
\hdashline
Sup. (ResNet) & 0.48 & 0.42  & 0.41 \\
Sup. (ViT) & \textbf{0.56} &0.49 &0.47 \\
Sup.-Large (ResNet) & 0.48 & 0.43 &0.44 \\
Sup.-Large (ViT) & \textbf{0.56}&  \textbf{0.59} & 0.48 \\
DINOv2 (ViT) &  0.54 &  0.56 & \textbf{0.49} \\
\hline
\end{tabular}
\end{table*}

\begin{table}
\caption{Median silhouette coefficients for domain-specific classification downstream scenarios}
\label{tab::Domain-Specific-Silhouette}
\centering
\begin{tabular}{|c||c|c|c|c|}
\hline
&\multicolumn{4}{c|}{Datasets}\\
\cline{2-5}
Method & AID & ChestX-Det & SIXRay-10 & RaabinWBC\\
\hline\hline
SimCLR (ResNet) & 0.44  &0.47  &0.53 & 0.31 \\
DINO (ViT) & 0.58 & \textbf{0.61} &  0.54 & 0.37 \\
MAE (ViT) & 0.51 & 0.48 &  0.60 & 0.35\\
DeepClusterV2 (ResNet) & 0.45 & 0.50 & 0.48 & 0.28\\
\hdashline
Sup. (ResNet) &0.50  &  0.54 & 0.55 & 0.30 \\
Sup. (ViT)& 0.53 & 0.58  &0.59 & 0.24 \\
Sup.-Large (ResNet) & 0.52 & 0.57 & 0.56 & 0.26\\
Sup.-Large (ViT) & 0.57  & 0.60  &0.61 & 0.24\\
DINOv2 (ViT) & \textbf{0.59}& 0.56 & \textbf{0.63} & \textbf{0.41}\\
\hline
\end{tabular}
\end{table}

Regarding the RELAX-based explanation estimation, it measures similarities in the representation space between an input image and multiple/different masked out versions of it, in order to identify the image regions that are the most important for the feature extractor. The method generates visual/qualitative explanations of the learnt representations in the pixel space in the form of attribution maps, by considering only the output of the backbone DNN that extracts the image embeddings for the test images of any downstream dataset, i.e., without utilizing the classification head and its predictions. By definition, the importance score of each pixel can vary for different masked/semi-occluded versions of the same image; the variance in these scores for a given pixel is a proxy for the uncertainty of the obtained explanation at that pixel. Thus, if image regions that RELAX deems to be highly important for trained feature extractor $F$ also tend to have low uncertainty, then $F$ likely represents robustly their semantic content.

Tables \ref{tab:silhouette_scores_imagenet_100}, \ref{tab:silhouette_scores_places_100} and \ref{tab::Domain-Specific-Silhouette} illustrate the estimated median silhouette coefficient $S_m$ for the considered object-centric, scene-centric and domain-specific classification downstream scenarios, respectively, from Sections \ref{sec::MainExperimentalResults} and \ref{sec::DomainSpecificExperimentalResults}. It must be noted that for the single-dataset setup in Tables \ref{tab:silhouette_scores_imagenet_100} and \ref{tab:silhouette_scores_places_100}, `Sup.' and `Sup.-Large' both imply a single setting, i.e., no pretraining/random initialization. In all cases, the median $S_m$ for each model was calculated by obtaining its inferred representations for the test images of each downstream dataset, after finetuning. As can be seen, the obtained quantitative explanations are in general closely correlated with the respective downstream classification accuracy scores (reported in Sections \ref{sec::MainExperimentalResults} and \ref{sec::DomainSpecificExperimentalResults}), indicating that indeed better performing SSL methods owe their discrimination capability to their ability to form more distinct, compact and well-separated class representations. This observation also holds for the supervised pretraining baselines and for the DINOv2 baseline, which is pretrained on the giant LVD-142M dataset and whose high test accuracy in the majority of transfer-learning setup cases is linked to its ability to generate very accurate/robust class representations with low intra-class variance and high inter-class separation in a zero-shot way; this is also in line with the foundation model status of off-the-shelf DINOv2 models. Additionally, it needs to be highlighted that in-domain low-data SSL pretraining in general leads to almost equally good or even better downstream image representations in domain-specific experimental setting, as also observed in the respective downstream classification tasks.

\begin{figure*}[!htb]
    \centering
    \begin{subfigure}{0.45\textwidth}
        \centering
        \includegraphics[width=1\textwidth]{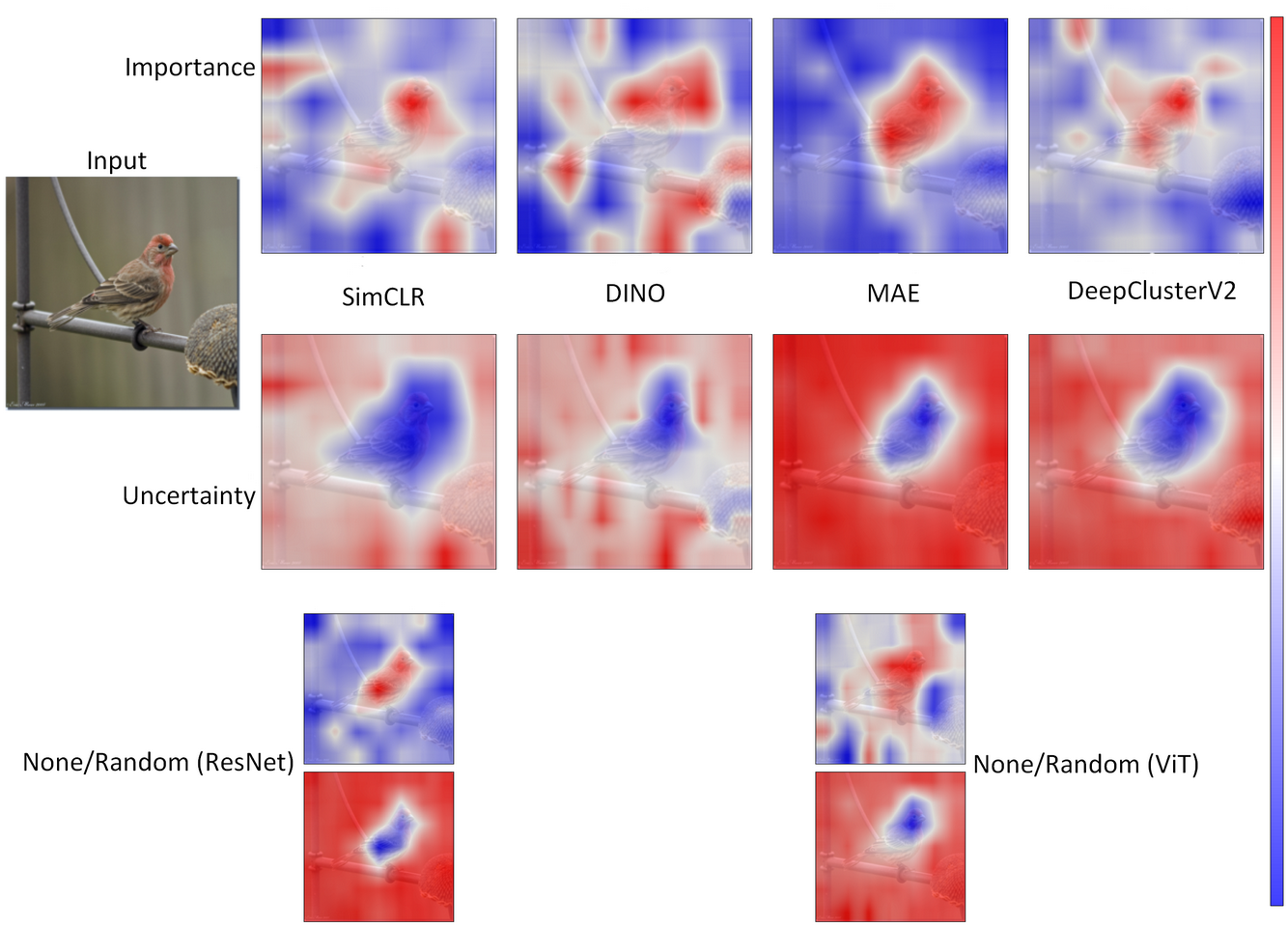}
        \caption{}
        \label{fig:sub1}
    \end{subfigure}
    \hfill
    \begin{subfigure}{0.45\textwidth}
        \centering
        \includegraphics[width=1\textwidth]{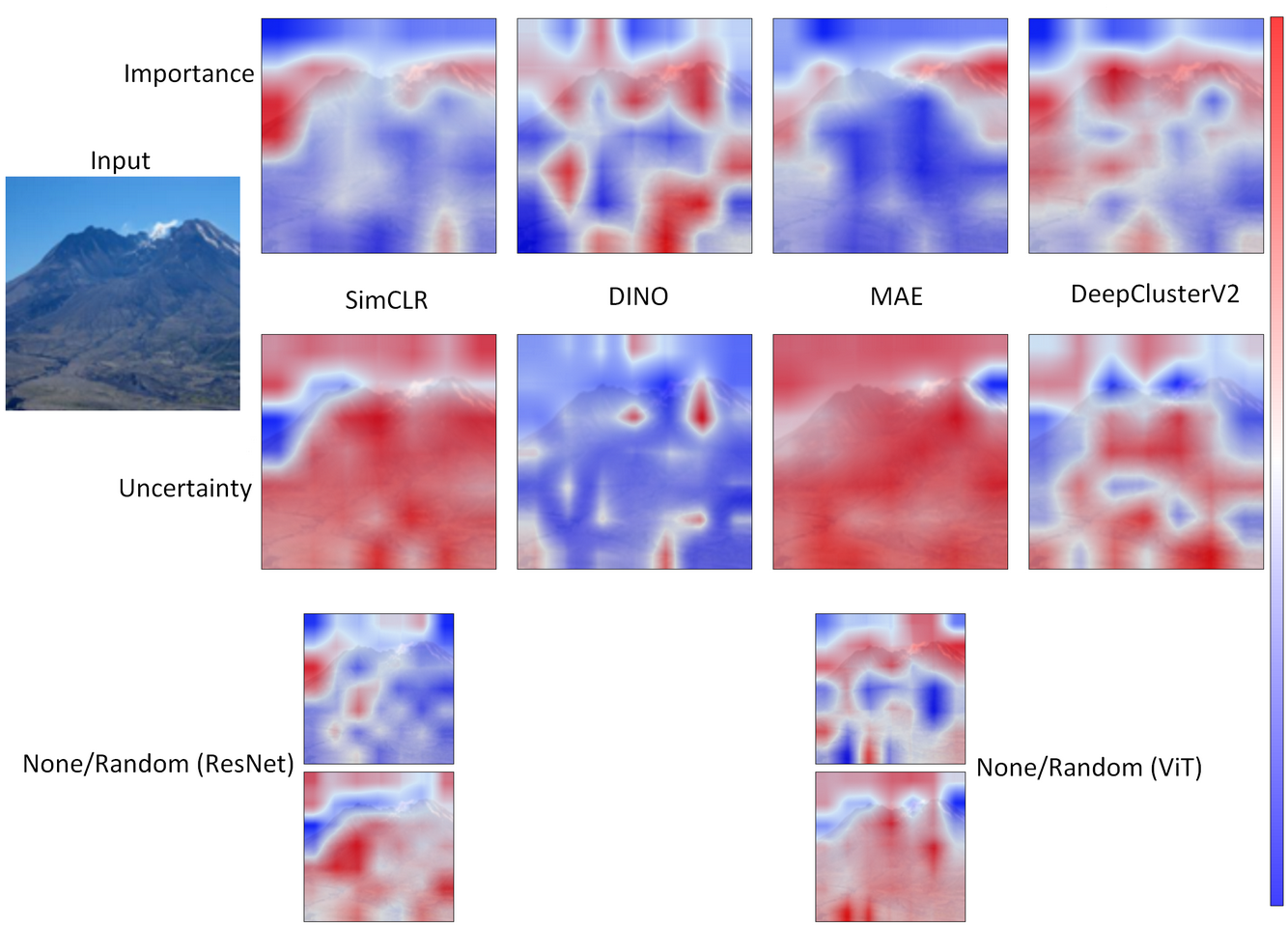}
        \caption{}
        \label{fig:sub2}
    \end{subfigure}
    
    \begin{subfigure}{\textwidth}
        \centering
        \includegraphics[width=0.5\textwidth]{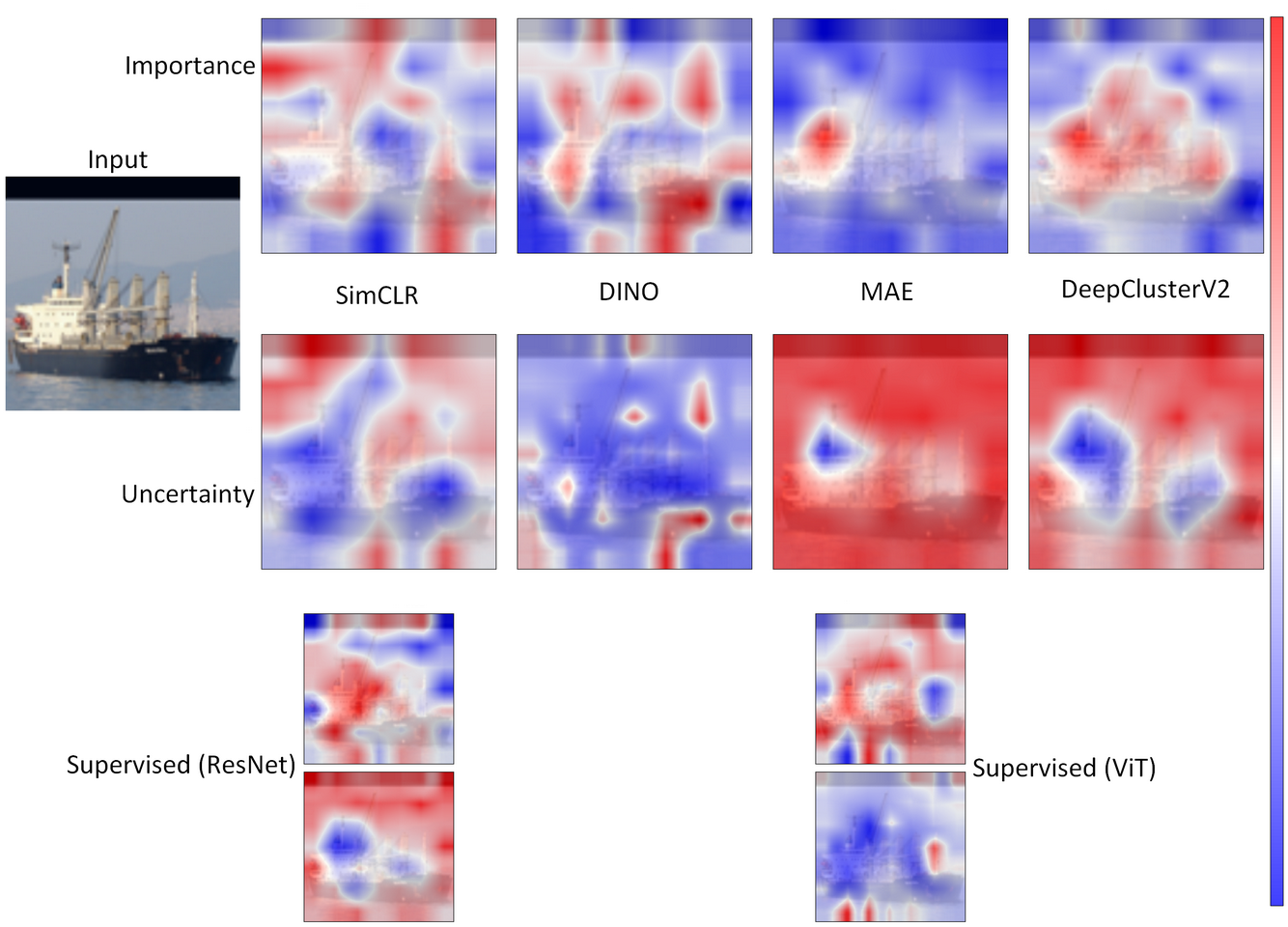}
        \caption{}
        \label{fig:sub3}
    \end{subfigure}
    \caption{RELAX-based explanations for indicative test images from: (a) ImageNet-100 (single-dataset setup), (b) Places-100 (single-dataset setup), and (c) STL-10 (transfer-learning setup). Red/blue color indicates high/low value, respectively.}
    \label{fig:RELAXExplanations}
\end{figure*}

Figure \ref{fig:RELAXExplanations} illustrates indicative RELAX-based visual explanations regarding the representations learnt by the various SSL pretraining methods, using exemplary test images from the ImageNet-100, Places-100 and STL-10 downstream datasets. Additionally, the outcome of low-data SSL pretraining is compared against a low-data supervised pretraining baseline (for the transfer-learning setup) or with downstream training from scratch with random initialization (for the single-dataset setup), both making use of ViT and ResNet architectures. In all cases, an importance and an uncertainty heatmap are depicted, where red/blue color indicates high/low value, respectively. The provided indicative examples were selected after extensive visual inspection of hundreds of generated explanations. From the obtained results, it can be seen that SimCLR and DINO tend to diffuse their attention to multiple image regions, while MAE and DeepClusterV2 are keen to emphasize only on a few contiguous key areas. In the case of MAE, these areas are in general highly consistent with the visible foreground object, a fact that might underlie the method's remarkable VTAB performance under low-data pretraining. On the other hand, DINO tends in general to generate features with lower uncertainty across the image in the majority of cases, which may explain its superiority in most benchmarks.

\subsection{Summary of Key Experimental Insights}

\begin{figure}[h!]
    \centering
    \begin{tikzpicture}
        \begin{axis}[
            width=14cm,
            height=12cm,
            xlabel={Pretraining Set Size},
            ylabel={DINO/ViT Accuracy Gain Over Sup./ViT (\%)},
            xmin=0, xmax=260000,
            ymin=-5, ymax=50,
            xtick={20000, 40000, 60000, 80000, 100000, 120000, 140000, 160000, 180000, 200000, 220000, 240000, 260000},
            ytick={-5, 0, 5, 10, 15, 20, 25, 30, 35, 40, 45, 50},
            xticklabels={20k, 40k, 60k, 80k, 100k, 120k, 140k, 160k, 180k, 200k, 220k, 240k},
            legend pos=north west,
            grid=both,
            grid style={dotted, gray},
            line width=1pt,
            mark options={scale=1.2},
            scaled x ticks=false,
        ]
        \addplot[
            color=blue,
            mark=*,
            nodes near coords,
            point meta=explicit symbolic,
            every node near coord/.append style={anchor=west, font=\small}
            ]
            coordinates {(32000, 2.8) [ChestX-Det]};
        \addlegendentry{MedPix (32k)}

        \addplot[
            color=red,
            mark=*,
            nodes near coords,
            point meta=explicit symbolic,
            every node near coord/.append style={anchor=west, font=\small}
            ]
            coordinates {(35000, -0.6) [SUN DB]};
        \addlegendentry{P-100 (35k)}

        \addplot[
            color=green,
            mark=*,
            nodes near coords,
            point meta=explicit symbolic,
            every node near coord/.append style={anchor=west, font=\small}
            ]
            coordinates {(80000, 0.6) [AID]};
        \addlegendentry{MLRSNet (80k)}

        \addplot[
            color=orange,
            mark=*,
            nodes near coords,
            point meta=explicit symbolic,
            every node near coord/.append style={anchor=west, font=\small}
            ]
            coordinates {(100000, 0.8) [COCO]};
        \addlegendentry{IN-100 (100k)}

        \addplot[
            color=purple,
            mark=*,
            nodes near coords,
            point meta=explicit symbolic,
            every node near coord/.append style={anchor=west, font=\small}
            ]
            coordinates {(210000, 1.6) [SIXRay-10]};
        \addlegendentry{SIXRay-100 (210k)}

        \addplot[
            color=cyan,
            mark=*,
            nodes near coords,
            point meta=explicit symbolic,
            every node near coord/.append style={anchor=south, font=\small}
            ]
            coordinates {(240000, 41.1) [RaabinWBC]};
        \addlegendentry{CEM500K (240k)}

        \end{axis}
    \end{tikzpicture}
    \caption{\hl{DINO/ViT gains compared to Sup./ViT, in the transfer-learning experiments (best viewed in color)}}    \label{fig::DatasetSizeSSLGainsOverSup}
\end{figure}

\begin{figure}[h!]
    \centering
    \begin{tikzpicture}
        \begin{axis}[
            width=14cm,
            height=12cm,
            xlabel={Pretraining Set Size},
            ylabel={DINO/ViT Accuracy Gain Over Sup.-Large/ViT (\%)},
            xmin=0, xmax=260000,
            ymin=-10, ymax=50,
            xtick={20000, 40000, 60000, 80000, 100000, 120000, 140000, 160000, 180000, 200000, 220000, 240000, 260000},
            ytick={-10, -5, 0, 5, 10, 15, 20, 25, 30, 35, 40, 45, 50},
            xticklabels={20k, 40k, 60k, 80k, 100k, 120k, 140k, 160k, 180k, 200k, 220k, 240k},
            legend pos=north west,
            grid=both,
            grid style={dotted, gray},
            line width=1pt,
            mark options={scale=1.2},
            scaled x ticks=false,
        ]
        \addplot[
            color=blue,
            mark=*,
            nodes near coords,
            point meta=explicit symbolic,
            every node near coord/.append style={anchor=west, font=\small}
            ]
            coordinates {(32000, 1.9) [ChestX-Det]};
        \addlegendentry{MedPix (32k)}

        \addplot[
            color=red,
            mark=*,
            nodes near coords,
            point meta=explicit symbolic,
            every node near coord/.append style={anchor=west, font=\small}
            ]
            coordinates {(35000, -6.0) [SUN DB]};
        \addlegendentry{P-100 (35k)}

        \addplot[
            color=green,
            mark=*,
            nodes near coords,
            point meta=explicit symbolic,
            every node near coord/.append style={anchor=west, font=\small}
            ]
            coordinates {(80000, -3.0) [AID]};
        \addlegendentry{MLRSNet (80k)}

        \addplot[
            color=orange,
            mark=*,
            nodes near coords,
            point meta=explicit symbolic,
            every node near coord/.append style={anchor=west, font=\small}
            ]
            coordinates {(100000, -4.7) [COCO]};
        \addlegendentry{IN-100 (100k)}

        \addplot[
            color=purple,
            mark=*,
            nodes near coords,
            point meta=explicit symbolic,
            every node near coord/.append style={anchor=west, font=\small}
            ]
            coordinates {(210000, 0.2) [SIXRay-10]};
        \addlegendentry{SIXRay-100 (210k)}

        \addplot[
            color=cyan,
            mark=*,
            nodes near coords,
            point meta=explicit symbolic,
            every node near coord/.append style={anchor=south, font=\small}
            ]
            coordinates {(240000, 46.0) [RaabinWBC]};
        \addlegendentry{CEM500K (240k)}

        \end{axis}
    \end{tikzpicture}
    \caption{\hl{DINO/ViT gains compared to Sup.-Large/ViT in the transfer-learning experiments (best viewed in color)}}
    \label{fig::DatasetSizeSSLGainsOverSupLarge}
\end{figure}
\hl{This study showed that in the single-dataset setup, where cross-dataset generalization is not prioritized, CNN-based SSL methods (e.g., SimCLR on ResNet) outperform Vision Transformers and ViT-oriented SSL methods (e.g., DINO) in the low-data regime, potentially due to the data hunger of Vision Transformers. A notable insight is the identification of a pretraining dataset size threshold (estimated between 50k-65k images, assuming identical pretraining and downstream image domains), below which SSL becomes ineffective for learning useful representations. In the main transfer-learning setup, DINO on ViT architectures emerges as the leading SSL method within the low-data regime, though it does not surpass the performance of conventional supervised pretraining approaches.}

\hl{In the semi-supervised single-dataset setup, DINO performs strongly, attributed to its capacity to learn more universally applicable embeddings that maintain effectiveness even in low-data pretraining. This ability proves especially advantageous in highly domain-specific transfer-learning settings, where in-domain low-data DINO pretraining consistently outperforms supervised large-scale pretraining on conventional RGB datasets in downstream accuracy. This advantage is most pronounced in fields significantly different from conventional natural image datasets, such as medical or security imaging}

\hl{Representation analysis reveals that DINO’s superior discrimination ability compared to other SSL methods is due to its formation of distinct, compact, and well-separated class representations, with a broader focus across the input image, leading to lower uncertainty in feature generation. In contrast, MAE shows a focused attention on the visible foreground object, supporting higher robustness under low-data pretraining conditions and needing less pretraining data to learn semantically discriminative features. However, its robustness is still lower than that of both large-scale and low-data supervised pretraining approaches, particularly in noisy or imbalanced settings.}

\hl{In order to more clearly showcase that any identified gains of low-data SSL pretraining over supervised pretraining in the transfer-learning setup are due to in-domain pretraining and not due to any differences in the size of the pretraining datasets, across the various sets of experiments, Figures \ref{fig::DatasetSizeSSLGainsOverSup} and \ref{fig::DatasetSizeSSLGainsOverSupLarge} visualize part of the information contained in Tables \ref{tab::Main-Transfer} and \ref{tab::Domain-Specific}. The horizontal axis is marked with the sizes of the SSL pretraining datasets, while the vertical axis represents downstream classification accuracy gains. Figure \ref{fig::DatasetSizeSSLGainsOverSup}/\ref{fig::DatasetSizeSSLGainsOverSupLarge} depicts the reported DINO/ViT gains compared to Sup. (ViT)/Sup.-Large (ViT), respectively. As it can be seen, COCO and SUN DB accuracy scores were the ones selected to be visualized for SSL pretraining in IN-100 and P-100, due to the significantly larger test set sizes of these downstream datasets over STL-10 and ADE20K. In general, it is evident that any downstream classification accuracy gains of DINO do not fully correlate with pretraining dataset size, indicating that pretraining domain relevance is also important.}

\section{Conclusions}
\label{sec::Conclusions}
In this paper, the issue of Self-Supervised Learning (SSL) in real-world application settings, i.e., cases where it is not always feasible to collect and/or to utilize very large (in the order of millions of instances) pretraining datasets, was examined. In this context, based on a taxonomy of modern SSL methods for image analysis, a comprehensive comparative experimental study was performed in the so-called `low-data regime', i.e., assuming a pretraining dataset with ~50k - ~300k images, aiming at shedding light on how the various SSL methods behave in such training scenarios. The performed comparative evaluation showcased that a pretraining dataset size limit exists, in the range ~50k - ~65k images \hl{(assuming identical pretraining and downstream image domains)}, below which SSL is ineffective for representation learning. In the majority of cases, supervised pretraining and large-scale DINOv2 pretraining surpassed low-data SSL pretraining in terms of both downstream accuracy and robustness, with DINO retaining its ability to learn more generally applicable embeddings in the low-data regime. Very interestingly, in domain-specific experiments, in-domain low-data SSL pretraining outperformed all competitors and even large-scale pretraining on general datasets. This finding suggests that when extreme amounts of data are not available and/or the target downstream dataset is significantly different in nature compared to the large-scale datasets commonly used for pretraining (e.g., medical and security imaging against conventional natural RGB images), in-domain low-data SSL pretraining offers recognition advantages. At a finer level of detail, when comparing the performance of SSL methods only, MAE seems to need less pretraining data than DINO, in order to learn generally applicable features that are inherently semantically discriminant and consistent with the visible objects. On the other hand, given a reasonably-sized downstream dataset for finetuning, DINO typically surpasses MAE in the low-data setting, since it achieves higher downstream classification accuracy and generates the most well-separated and compact class representations. Moreover, the superiority of DINO persisted in domain-specific scenarios, when moving beyond the RGB space, where it outperformed all SSL competitors and even large-scale supervised pretraining. This indicates that low-data SSL pretraining not only works, but is also highly relevant in cases of domain-specific downstream tasks where generic (e.g., ImageNet-1k pretraining-derived) representations cannot generalize well.

Furthermore, suggestions for future research can be extrapolated from the above-mentioned insights. For example, given the finding that low-data SSL is indeed useful in domain-specific settings, general SSL methodologies should be adapted for exploiting domain-specific image properties (e.g., particularities of X-ray scans, such as pixel-level visual overlaps among classes), to obtain SSL pretraining schemes with higher data efficiency (i.e., overcoming the identified training dataset size limit) and better domain-specific features. Additionally, despite the superiority of self-distillation in current SSL literature (explained in this study by the lower RELAX uncertainty of DINO), the generative MAE method dominates in low-data with respect to robustness. This is also elucidated in this study via RELAX visualizations, by observing that MAE features tend to emphasize the visible foreground object. New methods combining the best of both worlds in an adjustable manner (e.g., by setting a robustness-performance trade-off in pretraining) should become a priority for interested practitioners working with small datasets.

\section*{Acknowledgment}
The research leading to these results has received funding from the European Union's Horizon Europe research and innovation programme under grant agreement No 101073876 (Ceasefire). This publication reflects only the authors views. The European Union is not liable for any use that may be made of the information contained therein.
\bibliographystyle{elsarticle-num}
\bibliography{Bibliography}

\begin{thebibliography}{100}
\expandafter\ifx\csname url\endcsname\relax
  \def\url#1{\texttt{#1}}\fi
\expandafter\ifx\csname urlprefix\endcsname\relax\def\urlprefix{URL }\fi
\expandafter\ifx\csname href\endcsname\relax
  \def\href#1#2{#2} \def\path#1{#1}\fi

\bibitem{lecun2015deep}
Y.~LeCun, Y.~Bengio, G.~Hinton, Deep learning, Nature 521~(7553) (2015) 436--444.

\bibitem{girshick2014rich}
R.~Girshick, J.~Donahue, T.~Darrell, J.~Malik, Rich feature hierarchies for accurate object detection and semantic segmentation, in: Proceedings of the IEEE Conference on Computer Vision and Pattern Recognition (CVPR), 2014, pp. 580--587.

\bibitem{ren2015faster}
S.~Ren, K.~He, R.~Girshick, J.~Sun, {Faster R-CNN}: Towards real-time object detection with region proposal networks, in: Proceedings of the Advances in Neural Information Processing Systems (NIPS), Vol.~28, 2015.

\bibitem{long2015fully}
J.~Long, E.~Shelhamer, T.~Darrell, Fully convolutional networks for semantic segmentation, in: Proceedings of the IEEE Conference on Computer Vision and Pattern Recognition (CVPR), 2015, pp. 3431--3440.

\bibitem{chen2017deeplab}
L.-C. Chen, G.~Papandreou, I.~Kokkinos, K.~Murphy, A.~L. Yuille, {DeepLab}: Semantic image segmentation with deep convolutional nets, atrous convolution, and fully connected {CRFs}, IEEE Transactions on Pattern Analysis and Machine Intelligence 40~(4) (2017) 834--848.

\bibitem{vinyals2015show}
O.~Vinyals, A.~Toshev, S.~Bengio, D.~Erhan, Show and tell: A neural image caption generator, in: Proceedings of the IEEE Conference on Computer Vision and Pattern Recognition (CVPR), 2015, pp. 3156--3164.

\bibitem{imagenet_cvpr09}
J.~Deng, W.~Dong, R.~Socher, L.~Li, K.~Li, L.~Fei-Fei, {ImageNet}: A large-scale hierarchical image database, in: Proceedings of the IEEE/CVF Conference on Computer Vision and Pattern Recognition (CVPR), 2009.

\bibitem{Wang2023}
R.~Wang, Y.~Hao, L.~Hu, J.~Chen, M.~Chen, D.~Wu, Self-supervised learning with data-efficient supervised fine-tuning for crowd counting, IEEE Transactions on Multimedia 25 (2023) 1538--1546.

\bibitem{he2016deep}
K.~He, X.~Zhang, S.~Ren, J.~Sun, Deep residual learning for image recognition, in: Proceedings of the IEEE Conference on Computer Vision and Pattern Recognition (CVPR), 2016, pp. 770--778.

\bibitem{risojevic2021we}
V.~Risojevi{\'c}, V.~Stojni{\'c}, Do we still need {I}magenet pre-training in remote sensing scene classification?, arXiv preprint arXiv:2111.03690 (2021).

\bibitem{chen2021ssl++}
S.~Chen, J.-H. Xue, J.~Chang, J.~Zhang, J.~Yang, Q.~Tian, {SSL++}: Improving self-supervised learning by mitigating the proxy task-specificity problem, IEEE Transactions on Image Processing 31 (2021) 1134--1148.

\bibitem{azizi2023robust}
S.~Azizi, L.~Culp, J.~Freyberg, B.~Mustafa, S.~Baur, S.~Kornblith, T.~Chen, N.~Tomasev, J.~Mitrovi{\'c}, P.~Strachan, et~al., Robust and data-efficient generalization of self-supervised machine learning for diagnostic imaging, Nature Biomedical Engineering (2023) 1--24.

\bibitem{Zhang2023}
H.~Zhang, Y.~Luo, L.~Zhang, Y.~Wu, M.~Wang, Z.~Shen, Considering three elements of aesthetics: Multi-task self-supervised feature learning for image style classification, Neurocomputing 520 (2023) 262--273.

\bibitem{konkle2022self}
T.~Konkle, G.~A. Alvarez, A self-supervised domain-general learning framework for human ventral stream representation, Nature Communications 13~(1) (2022) 491.

\bibitem{goyal2021self}
P.~Goyal, M.~Caron, B.~Lefaudeux, M.~Xu, P.~Wang, V.~Pai, M.~Singh, V.~Liptchinsky, I.~Misra, A.~Joulin, et~al., Self-supervised pretraining of visual features in the wild, arXiv preprint arXiv:2103.01988 (2021).

\bibitem{Ying2024}
Z.~Ying, D.~Cheng, C.~Chen, X.~Li, P.~Zhu, Y.~Luo, Y.~Liang, Predicting stock market trends with self-supervised learning, Neurocomputing 568 (2024) 127033.

\bibitem{kenton2019bert}
J.~D. M.-W.~C. Kenton, L.~K. Toutanova, {BERT}: Pre-training of deep bidirectional transformers for language understanding, in: Proceedings of the Annual Conference of the North American Chapter of the Association for Computational Linguistics (NAACL), 2019.

\bibitem{schiappa2023self}
M.~C. Schiappa, Y.~S. Rawat, M.~Shah, Self-supervised learning for videos: A survey, ACM Computing Surveys 55~(13s) (2023) 1--37.

\bibitem{Chen23}
Z.~Chen, H.~Wang, C.~W. Chen, Self-supervised video representation learning by serial restoration with elastic complexity, IEEE Transactions on Multimedia (2023) 1--14.

\bibitem{liu2022audio}
S.~Liu, A.~Mallol-Ragolta, E.~Parada-Cabaleiro, K.~Qian, X.~Jing, A.~Kathan, B.~Hu, B.~W. Schuller, Audio self-supervised learning: A survey, Patterns 3~(12) (2022).

\bibitem{Wu23}
Y.~Wu, J.~Liu, M.~Gong, P.~Gong, X.~Fan, A.~K. Qin, Q.~Miao, W.~Ma, Self-supervised intra-modal and cross-modal contrastive learning for point cloud understanding, IEEE Transactions on Multimedia (2023) 1--13.

\bibitem{Tao23}
Z.~Tao, X.~Liu, Y.~Xia, X.~Wang, L.~Yang, X.~Huang, T.-S. Chua, Self-supervised learning for multimedia recommendation, IEEE Transactions on Multimedia 25 (2023) 5107--5116.

\bibitem{Luo22}
F.~Luo, S.~Chen, J.~Chen, Z.~Wu, Y.-G. Jiang, Self-supervised learning for semi-supervised temporal language grounding, IEEE Transactions on Multimedia (2022) 1--11.

\bibitem{Huang2023}
L.~Huang, X.~Fan, T.~Xia, Y.~Li, Y.~Ding, {SC2-N}et: Self-supervised learning for multi-view complementarity representation and consistency fusion network, Neurocomputing 556 (2023) 126695.

\bibitem{Dai2023}
Y.~Dai, Y.~Li, B.~Sun, Object and attribute recognition for product image with self-supervised learning, Neurocomputing 558 (2023) 126763.

\bibitem{gidaris2018unsupervised}
S.~Gidaris, N.~Komodakis, Unsupervised representation learning by predicting image rotations, arXiv preprint arXiv:1803.07728 (2018).

\bibitem{doersch2015unsupervised}
C.~Doersch, A.~Gupta, A.~A. Efros, Unsupervised visual representation learning by context prediction, in: Proceedings of the IEEE International Conference on Computer Vision (ICCV), 2015.

\bibitem{oord2018representation}
A.~v.~d. Oord, Y.~Li, O.~Vinyals, Representation learning with contrastive predictive coding, in: Proceedings of the Advances in Neural Information Processing Systems (NIPS), 2018, pp. 4724--4734.

\bibitem{vincent2008extracting}
P.~Vincent, H.~Larochelle, I.~Lajoie, Y.~Bengio, P.-A. Manzagol, Extracting and composing robust features with denoising autoencoders, in: Proceedings of the International Conference on Machine Learning (ICML), ACM, 2008, pp. 1096--1103.

\bibitem{kingma2013auto}
D.~P. Kingma, M.~Welling, Auto-encoding variational bayes, arXiv preprint arXiv:1312.6114 (2013).

\bibitem{goodfellow2014generative}
I.~Goodfellow, J.~Pouget-Abadie, M.~Mirza, B.~Xu, D.~Warde-Farley, S.~Ozair, A.~Courville, Y.~Bengio, Generative adversarial nets, in: Proceedings of the Advances in Neural Information Processing Systems (NIPS), 2014, pp. 2672--2680.

\bibitem{Lloyd1982}
S.~Lloyd, Least squares quantization in {PCM}, IEEE Transactions on Information Theory 28~(2) (1982) 129--137.

\bibitem{hinton2015distilling}
G.~Hinton, O.~Vinyals, J.~Dean, Distilling the knowledge in a neural network, arXiv preprint arXiv:1503.02531 (2015).

\bibitem{Gui2023survey}
J.~Gui, T.~Chen, J.~Zhang, Q.~Cao, Z.~Sun, H.~Luo, D.~Tao, A survey on self-supervised learning: Algorithms, applications, and future trends, arXiv preprint arXiv:2301.05712 (2023).

\bibitem{gutmann2012noise}
M.~U. Gutmann, A.~Hyv{\"a}rinen, Noise-contrastive estimation of unnormalized statistical models, with applications to natural image statistics, Journal of Machine Learning Research 13~(Feb) (2012) 307--361.

\bibitem{Zhao2020SSL}
Q.~Zhao, J.~Dong, Self-supervised representation learning by predicting visual permutations, Knowledge-Based Systems 210 (2020) 106534.

\bibitem{Zhai2024}
R.~Zhai, B.~Liu, A.~Risteski, Z.~Kolter, P.~Ravikumar, Understanding augmentation-based self-supervised representation learning via {RKHS} approximation and regression, in: Proceedings of the International Conference on Learning Representations (ICLR), 2024.

\bibitem{Purushwalkam2020}
S.~Purushwalkam, A.~Gupta, Demystifying contrastive self-supervised learning: Invariances, augmentations and dataset biases, in: Proceedings of the Advances in Neural Information Processing Systems (NIPS), Vol.~33, 2020, pp. 3407--3418.

\bibitem{Moutakanni2024}
T.~Moutakanni, M.~Oquab, M.~Szafraniec, M.~Vakalopoulou, P.~Bojanowski, You don't need data-augmentation in {Self-Supervised Learning}, arXiv preprint arXiv:2406.09294 (2024).

\bibitem{He2024preventing}
J.~He, J.~Du, W.~Ma, Preventing dimensional collapse in {Self-Supervised Learning} via {Orthogonality Regularization}, arXiv preprint arXiv:2411.00392 (2024).

\bibitem{Jang2023Self}
J.~Jang, S.~Kim, K.~Yoo, C.~Kong, J.~Kim, N.~Kwak, Self-distilled self-supervised representation learning, in: Proceedings of the IEEE/CVF Winter Conference on Applications of Computer Vision, 2023, pp. 2829--2839.

\bibitem{zbontar2021barlow}
J.~Zbontar, L.~Jing, I.~Misra, Y.~LeCun, et~al., {Barlow Twins}: Self-supervised learning via redundancy reduction, arXiv preprint arXiv:2103.03230 (2021).

\bibitem{Kalantidis2021}
Y.~Kalantidis, C.~Lassance, J.~Almazan, D.~Larlus, {TLDR}: Twin learning for dimensionality reduction, Transactions on Machine Learning Research (2022).

\bibitem{caron2021emerging}
M.~Caron, H.~Touvron, I.~Misra, H.~J{\'e}gou, J.~Mairal, P.~Bojanowski, A.~Joulin, Emerging properties in self-supervised {Vision Transformers}, arXiv preprint arXiv:2104.14294 (2021).

\bibitem{Oquab2023DinoV2}
M.~Oquab, T.~Darcet, T.~Moutakanni, H.~Vo, M.~Szafraniec, V.~Khalidov, P.~Fernandez, D.~Haziza, F.~Massa, A.~El-Nouby, et~al., {DINOv2}: Learning robust visual features without supervision, arXiv preprint arXiv:2304.07193 (2023).

\bibitem{chhipa2023can}
P.~C. Chhipa, J.~R. Holmgren, K.~De, R.~Saini, M.~Liwicki, Can self-supervised representation learning methodswithstand distribution shifts and corruptions?, in: Proceedings of the IEEE/CVF International Conference on Computer Vision (ICCV), 2023, pp. 4467--4476.

\bibitem{caron2020unsupervised}
M.~Caron, I.~Misra, J.~Mairal, P.~Goyal, P.~Bojanowski, A.~Joulin, Unsupervised learning of visual features by contrasting cluster assignments, in: Proceedings of the Advances in Neural Information Processing Systems (NIPS), 2020.

\bibitem{he2022masked}
K.~He, X.~Chen, S.~Xie, Y.~Li, P.~Doll{\'a}r, R.~Girshick, Masked autoencoders are scalable vision learners, in: Proceedings of the IEEE/CVF Conference on Computer Vision and Pattern Recognition (CVPR), 2022, pp. 16000--16009.

\bibitem{Kong2023}
X.~Kong, X.~Zhang, Understanding masked image modeling via learning occlusion-invariant feature, in: Proceedings of the IEEE/CVF Conference on Computer Vision and Pattern Recognition (CVPR), 2023, pp. 6241--6251.

\bibitem{Xie2023}
Z.~Xie, Z.~Zhang, Y.~Cao, Y.~Lin, Y.~Wei, Q.~Dai, H.~Hu, On data scaling in {Masked Image Modeling}, in: Proceedings of the IEEE/CVF Conference on Computer Vision and Pattern Recognition (CVPR), 2023, pp. 10365--10374.

\bibitem{El2021}
A.~El-Nouby, G.~Izacard, H.~Touvron, I.~Laptev, H.~Jegou, E.~Grave, Are large-scale datasets necessary for self-supervised pre-training?, arXiv preprint arXiv:2112.10740 (2021).

\bibitem{chen2020simple}
T.~Chen, S.~Kornblith, M.~Norouzi, G.~Hinton, A simple framework for contrastive learning of visual representations, in: Proceedings of the International Conference on Machine Learning (ICML), 2020, pp. 1597--1607.

\bibitem{dosovitskiy2020image}
A.~Dosovitskiy, L.~Beyer, A.~Kolesnikov, D.~Weissenborn, X.~Zhai, T.~Unterthiner, M.~Dehghani, M.~Minderer, G.~Heigold, S.~Gelly, et~al., An image is worth 16x16 words: {Transformers} for image recognition at scale, arXiv preprint arXiv:2010.11929 (2020).

\bibitem{zhao2020maintaining}
B.~Zhao, X.~Xiao, G.~Gan, B.~Zhang, S.-T. Xia, Maintaining discrimination and fairness in class incremental learning, in: Proceedings of the IEEE/CVF Conference on Computer Vision and Pattern Recognition (CVPR), 2020, pp. 13208--13217.

\bibitem{zhou2017places}
B.~Zhou, A.~Lapedriza, A.~Khosla, A.~Oliva, A.~Torralba, Places: A 10 million image database for scene recognition, IEEE Transactions on Pattern Analysis and Machine Intelligence (2017).

\bibitem{he2021quantifying}
X.~He, Y.~Zhang, Quantifying and mitigating privacy risks of contrastive learning, in: Proceedings of the ACM SIGSAC Conference on Computer and Communications Security, 2021, pp. 845--863.

\bibitem{lin2014microsoft}
T.-Y. Lin, M.~Maire, S.~Belongie, J.~Hays, P.~Perona, D.~Ramanan, P.~Doll{\'a}r, C.~L. Zitnick, Microsoft {COCO}: Common objects in context, in: Proceedings of the European Conference on Computer Vision (ECCV), Springer, 2014, pp. 740--755.

\bibitem{coates2011stl10}
A.~Coates, A.~Ng, H.~Lee, An analysis of single layer networks in unsupervised feature learning, in: Proceedings of the International Conference on Artificial Intelligence and Statistics (AISTATS), PMLR, 2011.

\bibitem{zhou2017scene}
B.~Zhou, H.~Zhao, X.~Puig, S.~Fidler, A.~Barriuso, A.~Torralba, Scene parsing through {ADE20k} dataset, in: Proceedings of the IEEE/CVF Conference on Computer Vision and Pattern Recognition (CVPR), 2017, pp. 633--641.

\bibitem{xiao2010sun}
J.~Xiao, J.~Hays, K.~A. Ehinger, A.~Oliva, A.~Torralba, {SUN Database}: Large-scale scene recognition from abbey to zoo, in: Proceedings of the IEEE/CVF Conference on Computer Vision and Pattern Recognition, 2010, pp. 3485--3492.

\bibitem{lee2013pseudo}
D.-H. Lee, et~al., Pseudo-label: The simple and efficient semi-supervised learning method for {Deep Neural Networks}, in: Proceedings of the International Conference on Machine Learning Workshops (ICMLW), Vol.~3, 2013, p. 896.

\bibitem{zhai2019s4l}
X.~Zhai, A.~Oliver, A.~Kolesnikov, L.~Beyer, {S4L}: Self-supervised semi-supervised learning, in: Proceedings of the IEEE/CVF International Conference on Computer Vision (ICCV), 2019, pp. 1476--1485.

\bibitem{hendrycks2021natural}
D.~Hendrycks, K.~Zhao, S.~Basart, J.~Steinhardt, D.~Song, Natural adversarial examples, in: Proceedings of the IEEE/CVF Conference on Computer Vision and Pattern Recognition (CVPR), 2021.

\bibitem{hendrycks2019benchmarking}
D.~Hendrycks, T.~Dietterich, Benchmarking neural network robustness to common corruptions and perturbations, in: Proceedings of the International Conference on Learning Representations (ICLR), 2019.

\bibitem{openlongtailrecognition}
Z.~Liu, Z.~Miao, X.~Zhan, J.~Wang, B.~Gong, S.~X. Yu, Large-scale long-tailed recognition in an open world, in: Proceedings of the IEEE/CVF Conference on Computer Vision and Pattern Recognition (CVPR), 2019.

\bibitem{zhai2019visual}
X.~Zhai, J.~Puigcerver, A.~Kolesnikov, P.~Ruyssen, C.~Riquelme, M.~Lucic, J.~Djolonga, A.~S. Pinto, M.~Neumann, A.~Dosovitskiy, et~al., {The Visual Task Adaptation Benchmark} (2019).

\bibitem{qi2020mlrsnet}
X.~Qi, P.~Zhu, Y.~Wang, L.~Zhang, J.~Peng, M.~Wu, J.~Chen, X.~Zhao, N.~Zang, P.~T. Mathiopoulos, {MLRSNet}: A multi-label high spatial resolution remote sensing dataset for semantic scene understanding, ISPRS Journal of Photogrammetry and Remote Sensing 169 (2020) 337--350.

\bibitem{MedPixDataset}
N.~L. of~Medicine, \href{https://medpix.nlm.nih.gov/home}{Medpix medical image database}.
\newline\urlprefix\url{https://medpix.nlm.nih.gov/home}

\bibitem{Miao2019SIXray}
C.~Miao, L.~Xie, F.~Wan, c.~Su, H.~Liu, j.~Jiao, Q.~Ye, {SIXray}: A large-scale security inspection {X}-ray benchmark for prohibited item discovery in overlapping images, in: Proceedings of the IEEE/CVF Conference on Computer Vision and Pattern Recognition (CVPR), 2019.

\bibitem{CEM500K}
R.~Conrad, K.~Narayan, {CEM500K}, a large-scale heterogeneous unlabeled cellular electron microscopy image dataset for deep learning, Elife 10 (2021) e65894.

\bibitem{JocherYOLO82023}
G.~Jocher, A.~Chaurasia, J.~Qiu, \href{https://github.com/ultralytics/ultralytics}{{Ultralytics YOLO}} (2023).
\newline\urlprefix\url{https://github.com/ultralytics/ultralytics}

\bibitem{Xia2017AID}
G.-S. Xia, J.~Hu, F.~Hu, B.~Shi, X.~Bai, Y.~Zhong, L.~Zhang, X.~Lu, {AID}: A benchmark data set for performance evaluation of aerial scene classification, IEEE Transactions on Geoscience and Remote Sensing 55~(7) (2017) 3965--3981.

\bibitem{lian2021structure}
J.~Lian, J.~Liu, S.~Zhang, K.~Gao, X.~Liu, D.~Zhang, Y.~Yu, A structure-aware relation network for thoracic diseases detection and segmentation, IEEE Transactions on Medical Imaging 40~(8) (2021) 2042--2052.

\bibitem{RaabinWBC}
Z.~M. Kouzehkanan, S.~Saghari, S.~Tavakoli, P.~Rostami, M.~Abaszadeh, F.~Mirzadeh, E.~S. Satlsar, M.~Gheidishahran, F.~Gorgi, S.~Mohammadi, et~al., A large dataset of white blood cells containing cell locations and types, along with segmented nuclei and cytoplasm, Scientific reports 12~(1) (2022) 1123.

\bibitem{BIRDSAI}
E.~Bondi, R.~Jain, P.~Aggrawal, S.~Anand, R.~Hannaford, A.~Kapoor, J.~Piavis, S.~Shah, L.~Joppa, B.~Dilkina, et~al., {BIRDSAI}: A dataset for detection and tracking in aerial thermal infrared videos, in: Proceedings of the IEEE/CVF Winter Conference on Applications of Computer Vision (WACV), 2020, pp. 1747--1756.

\bibitem{HITUAV}
J.~Suo, T.~Wang, X.~Zhang, H.~Chen, W.~Zhou, W.~Shi, {HIT-UAV}: A high-altitude infrared thermal dataset for {Unmanned Aerial Vehicle}-based object detection, Scientific Data 10~(1) (2023) 227.

\bibitem{liu2020chestx}
J.~Liu, J.~Lian, Y.~Yu, {ChestX-Det10: chest X}-ray dataset on detection of thoracic abnormalities, arXiv preprint arXiv:2006.10550 (2020).

\bibitem{xu2021vitae}
Y.~Xu, Q.~Zhang, J.~Zhang, D.~Tao, {ViTAE}: {Vision Transformer} advanced by exploring intrinsic inductive bias, in: Proceedings of the Advances in Neural Information Processing Systems (NIPS), Vol.~34, 2021, pp. 28522--28535.

\bibitem{Rousseeuw1987silhouettes}
P.~J. Rousseeuw, Silhouettes: a graphical aid to the interpretation and validation of cluster analysis, Journal of Computational and Applied Mathematics 20 (1987) 53--65.

\bibitem{Arbelaitz2013}
O.~Arbelaitz, I.~Gurrutxaga, J.~Muguerza, J.~M. P{\'e}rez, I.~Perona, An extensive comparative study of cluster validity indices, Pattern Recognition 46~(1) (2013) 243--256.

\bibitem{wickstrom2023relax}
K.~K. Wickstr{\o}m, D.~J. Trosten, S.~L{\o}kse, A.~Boubekki, K.~{\o}. Mikalsen, M.~C. Kampffmeyer, R.~Jenssen, {RELAX}: Representation learning explainability, International Journal of Computer Vision 131~(6) (2023) 1584--1610.

\bibitem{deng2021does}
W.~Deng, S.~Gould, L.~Zheng, What does rotation prediction tell us about classifier accuracy under varying testing environments?, in: Proceedings of the International Conference on Machine Learning (ICML), PMLR, 2021.

\bibitem{zhuang2019}
C.~Zhuang, A.~Zhai, D.~Yamins, Local aggregation for unsupervised learning of visual embeddings, Proceedings of the IEEE International Conference on Computer Vision (ICCV) (2019) 6002--6012.

\bibitem{timofte2013anchored}
R.~Timofte, V.~De~Smet, L.~Van~Gool, Anchored neighborhood regression for fast example-based super-resolution, in: Proceedings of the IEEE International Conference on Computer Vision (ICCV), 2013, pp. 1920--1927.

\bibitem{bao2021beit}
H.~Bao, L.~Dong, S.~Piao, F.~Wei, {BEiT: BERT} pre-training of image transformers, arXiv preprint arXiv:2106.08254 (2021).

\bibitem{Devlin2019}
J.~Devlin, M.-W. Chang, K.~Lee, K.~Toutanova, {BERT}: Pre-training of deep bidirectional {Transformers} for language understanding, in: Proceedings of the Conference of the North {A}merican Chapter of the Association for Computational Linguistics: Human Language Technologies, Volume 1 (Long and Short Papers), Association for Computational Linguistics, 2019.

\bibitem{ramesh2021zero}
A.~Ramesh, M.~Pavlov, G.~Goh, S.~Gray, C.~Voss, A.~Radford, M.~Chen, I.~Sutskever, Zero-shot text-to-image generation, in: Proceedings of the International Conference on Machine Learning (ICML), PMLR, 2021, pp. 8821--8831.

\bibitem{van2017neural}
A.~Van Den~Oord, O.~Vinyals, et~al., Neural discrete representation learning, in: Proceedings of the Advances in Neural Information Processing Systems (NIPS), Vol.~30, 2017.

\bibitem{assran2023self}
M.~Assran, Q.~Duval, I.~Misra, P.~Bojanowski, P.~Vincent, M.~Rabbat, Y.~LeCun, N.~Ballas, Self-supervised learning from images with a joint-embedding predictive architecture, in: Proceedings of the IEEE/CVF Conference on Computer Vision and Pattern Recognition (CVPR), 2023, pp. 15619--15629.

\bibitem{he2020momentum}
K.~He, H.~Fan, Y.~Wu, S.~Xie, R.~Girshick, Momentum contrast for unsupervised visual representation learning, in: Proceedings of the IEEE Conference on Computer Vision and Pattern Recognition (CVPR), 2020, pp. 9729--9738.

\bibitem{Men2023}
Q.~Men, E.~S.L.~Ho, H.~P.H.~Shum, H.~Leung, Focalized contrastive view-invariant learning for self-supervised skeleton-based action recognition, Neurocomputing 537 (2023) 198--209.

\bibitem{hadsell2006}
R.~Hadsell, S.~Chopra, Y.~LeCun, Dimensionality reduction by learning an invariant mapping, Proceedings of the IEEE Conference on Computer Vision and Pattern Recognition (CVPR) 2 (2006) 1735--1742.

\bibitem{dosovitskiy2014}
A.~Dosovitskiy, J.~T. Springenberg, M.~Riedmiller, T.~Brox, Discriminative unsupervised feature learning with convolutional neural networks, Proceedings of the Advances in Neural Information Processing Systems (NIPS) 27 (2014).

\bibitem{wu2018}
Z.~Wu, Y.~Xiong, S.~X. Yu, D.~Lin, Unsupervised feature learning via non-parametric instance discrimination, in: Proceedings of the IEEE Conference on Computer Vision and Pattern Recognition (CVPR), 2018, pp. 3733--3742.

\bibitem{tian2019}
Y.~Tian, D.~Krishnan, P.~Isola, Contrastive multiview coding, in: Proceedings of the European Conference on Computer Vision (ECCV), 2019, pp. 552--569.

\bibitem{he2019}
K.~He, H.~Fan, Y.~Wu, S.~Xie, R.~Girshick, Momentum contrast for unsupervised visual representation learning, in: Proceedings of the IEEE/CVF Conference on Computer Vision and Pattern Recognition (CVPR), 2019, pp. 9729--9738.

\bibitem{misra2019}
I.~Misra, L.~van~der Maaten, Self-supervised learning of pretext-invariant representations, Proceedings of the IEEE/CVF Conference on Computer Vision and Pattern Recognition (2019) 6707--6717.

\bibitem{doersch2017}
C.~Doersch, A.~Zisserman, Multi-task self-supervised visual learning, in: Proceedings of the IEEE International Conference on Computer Vision (ICCV), 2017, pp. 2051--2060.

\bibitem{ye2019}
M.~Ye, X.~Zhang, P.~C. Yuen, S.-F. Chang, Unsupervised embedding learning via invariant and spreading instance feature, in: Proceedings of the IEEE/CVF Conference on Computer Vision and Pattern Recognition (CVPR), 2019, pp. 6210--6219.

\bibitem{ji2019}
X.~Ji, J.~F. Henriques, A.~Vedaldi, Invariant information clustering for unsupervised image classification and segmentation, in: Proceedings of the IEEE/CVF International Conference on Computer Vision (ICCV), 2019, pp. 9865--9874.

\bibitem{barlow1961possible}
H.~B. Barlow, et~al., Possible principles underlying the transformation of sensory messages, Sensory Communication 1~(01) (1961) 217--233.

\bibitem{grill2020bootstrap}
J.-B. Grill, F.~Strub, F.~Altch{\'e}, C.~Tallec, P.~Richemond, E.~Buchatskaya, C.~Doersch, B.~Avila~Pires, Z.~Guo, M.~Gheshlaghi~Azar, et~al., {Bootstrap Your Own Latent}: a new approach to self-supervised learning, Proceedings of the Advances in Neural Information Processing Systems (NIPS) 33 (2020) 21271--21284.

\bibitem{Mnih2015}
V.~Mnih, K.~Kavukcuoglu, D.~Silver, A.~A. Rusu, J.~Veness, M.~G. Bellemare, A.~Graves, M.~Riedmiller, A.~K. Fidjeland, G.~Ostrovski, et~al., Human-level control through deep reinforcement learning, Nature 518~(7540) (2015) 529--533.

\bibitem{chen2021exploring}
X.~Chen, K.~He, Exploring simple siamese representation learning, in: Proceedings of the IEEE/CVF Conference on Computer Vision and Pattern Recognition, 2021, pp. 15750--15758.

\bibitem{zhang2022does}
C.~Zhang, K.~Zhang, C.-D. Yoo, I.-S. Kweon, How does {SimSiam} avoid collapse without negative samples? {T}owards a unified understanding of progress in {SSL}, in: Proceedings of the International Conference on Learning Representations (ICLR), 2022.

\bibitem{radford2021learning}
A.~Radford, J.~W. Kim, C.~Hallacy, A.~Ramesh, G.~Goh, S.~Agarwal, G.~Sastry, A.~Askell, P.~Mishkin, J.~Clark, et~al., Learning transferable visual models from natural language supervision, in: Proceedings of the International Conference on Machine Learning (ICML), PMLR, 2021, pp. 8748--8763.

\bibitem{zhou2021ibot}
J.~Zhou, C.~Wei, H.~Wang, W.~Shen, C.~Xie, A.~Yuille, T.~Kong, {iBOT}: Image bert pre-training with online tokenizer, arXiv preprint arXiv:2111.07832 (2021).

\bibitem{cuturi2013sinkhorn}
M.~Cuturi, Sinkhorn distances: Lightspeed computation of optimal transport, in: Proceedings of the Advances in Neural Information Processing Systems (NIPS), Vol.~26, 2013.

\bibitem{ruan2022weighted}
Y.~Ruan, S.~Singh, W.~Morningstar, A.~A. Alemi, S.~Ioffe, I.~Fischer, J.~V. Dillon, Weighted ensemble self-supervised learning, in: Proceedings of the International Conference on Learning Representations (ICLR), 2023.

\bibitem{sablayrolles2018spreading}
A.~Sablayrolles, M.~Douze, C.~Schmid, H.~J{\'e}gou, Spreading vectors for similarity search, in: Proceedings of the International Conference on Learning Representations (ICLR), 2018.

\bibitem{zhou2022deep}
X.~Zhou, N.~L. Zhang, Deep clustering with features from self-supervised pretraining, arXiv preprint arXiv:2207.13364 (2022).

\bibitem{caron2018deep}
M.~Caron, P.~Bojanowski, A.~Joulin, M.~Douze, Deep clustering for unsupervised learning of visual features, in: Proceedings of the European Conference on Computer Vision (ECCV), 2018, pp. 132--149.

\bibitem{Asano2020}
Y.~M. Asano, C.~Rupprecht, A.~Vedaldi, Self-labelling via simultaneous clustering and representation learning, in: Proceedings of the International Conference on Learning Representations (ICLR), 2020.

\bibitem{bahdanau}
D.~Bahdanau, K.~Cho, Y.~Bengio, {Neural Machine Translation} by jointly learning to align and translate, in: Proceedings of the International Conference on Learning Representations (ICLR), 2015.

\end{thebibliography}
\newpage
\appendix
\section{Details of SSL pretext tasks}
This Appendix details a selection of important pretext tasks for each of the 4 main visual SSL categories: generative, contrastive, self-distillation and clustering approaches.
\subsection{Generative pretext}
During the early days of visual SSL for DNNs, (relatively) simple generative methods comprised the dominant approach. Research in this algorithmic category is still active, although alternative approaches have also attracted similar or increased attention. Key representative generative SSL methods are described in the followings.

\subsubsection{Rotation Prediction}
Rotation prediction \cite{gidaris2018unsupervised} is a simple SSL pretext task where the DNN must predict the rotation that has been artificially applied to a given input image. Therefore, the degree of rotation is utilized as a pseudo-ground-truth label. Typically, each training image is rotated to one of the $K$ predefined angles (e.g., \(0^{\circ}\), \(90^{\circ}\), \(180^{\circ}\), \(270^{\circ}\)) and the DNN is trained to classify the image into one of these rotation classes. It does not correct/undo the pre-applied rotation, but simply identifies it. From a representation learning standpoint, pretraining on this task forces the DNN model to learn basic image semantics \cite{deng2021does}, despite its simplicity.

Categorical Cross-Entropy is most commonly employed as the classification loss function for rotation prediction, using the actually pre-applied rotation as pseudo-ground-truth label:

\begin{equation}
\mathcal{L}_{Rot} = \mathcal{L}_{CE}(\mathbf{y}, \mathbf{p}),
\end{equation}
\noindent where \(\mathbf{y} \in {0,1}^K \) is the one-hot-encoded label vector corresponding to rotated input image $\mathbf{X} \in \mathbb{R}^{W \times H \times 3}$ and \(\mathbf{p} \in \mathbb{R}^K\) is the output vector of predicted probabilities given $\mathbf{X}$.

After pretraining, the classification neural head can be discarded and the pretrained backbone DNN may be employed for downstream finetuning.

\subsubsection{Colorization}
Colorization \cite{zhuang2019} involves converting grayscale images back into their original colored versions, which are utilized as pseudo-ground-truth. By pretraining on this task, the DNN learns about the natural color distributions and correlations in images. Thus, in order to colorize them correctly, the DNN learns to encode their semantic context (e.g., skies are blue, foliage is green, etc.).

The loss function employed for colorization typically is the mean squared error (MSE) between the DNN's output \(\mathbf{P} \in \mathbb{R}^{W \times H \times 3}\) and the original colored image \(\mathbf{Y} \in \mathbb{R}^{W \times H \times 3}\):

\begin{equation}
\mathcal{L}_{Col} = \mathcal{L}_{MSE}(\mathbf{P}, \mathbf{Y}).
\end{equation}

In this task, $\mathbf{X} \in \mathbb{R}^{W \times H \times 3}$ is the respective, grayscale input image. After pretraining, the colored image prediction head can be discarded and the pretrained feature extractor finetuned for downstream tasks. The head is typically an entire Decoder subnetwork. 

\subsubsection{Super-resolution}
Super-resolution \cite{timofte2013anchored} is an SSL pretext task where the DNN learns to map low-resolution images into their high-resolution original versions, which are exploited as pseudo-ground-truth. This is not simple upsampling, since the DNN must learn to infer realistic fine details; to achieve this, it must gradually learn to encode image semantics that are necessary for a plausible reconstruction of the original high-resolution image.

The typically employed loss function is the Mean Squared Error (MSE) between the predicted high-resolution image \(\mathbf{P}\) and the original high-resolution image \( \mathbf{Y} \):

\begin{equation}
\mathcal{L}_{SR} = \mathcal{L}_{MSE}(\mathbf{P}, \mathbf{Y}).
\end{equation}

In this task, $\mathbf{X} \in \mathbb{R}^{w \times h \times 3}$, where $w < W$ and $h < H$, is the respective, low-resolution input image. After pretraining, the high-resolution image prediction head can be discarded and the pretrained feature extractor finetuned for downstream tasks. The head is typically an entire Decoder subnetwork.

\subsubsection{BEiT}
While the majority of SSL pretext tasks are DNN architecture-agnostic, `Bidirectional Encoder representation from Image Transformers' (BEiT, \cite{bao2021beit}) is specifically designed for Vision Transformers \cite{dosovitskiy2020image}. Inspired by the well-known BERT architecture for Natural Language Processing (NLP) tasks \cite{Devlin2019}, it introduces the so-called `Masked Image Modeling' (MIM) generative pretext task.

This task considers two independent and alternative views of each image: a) a sequence of $M$ non-overlapping image blocks in raw pixel space, which are called `patches', and b) a sequence of $M$ image region representations in the low-dimensional latent space of a pretrained discrete Variational Autoencoder (VAE) \cite{ramesh2021zero}, which are called `visual tokens'. This space has been pre-discretized into a vocabulary of $V = 8,192$ visual words, leading to the assignment of an integer ID $v \in \mathbb{N}, 1\leq v \leq V$ to each image region, based on its initial latent representation, via a nearest neighbour look-up \cite{van2017neural}. Thus, each visual token ends up being an integer index to the vocabulary. The DNN input for a given image $\mathbf{Y}$ is the latter one's corresponding sequence of patches, after approximately 40\% of them have been masked and replaced by a special mask embedding. The respective pseudo-ground-truth target is the sequence of visual tokens for the masked patches of $\mathbf{Y}$. Thus, during MIM pretraining, the DNN must learn to generate the sequence of visual tokens corresponding to the hidden parts of each $\mathbf{Y}$, given as input both its masked and its visible patches.

Obviously, the first subnetwork of the pretrained VAE, which is called `tokenizer', must be available a priori for BEiT pretraining to proceed, since it is utilized to produce the pseudo-ground-truth targets for all training images. Reliance of the image reconstruction on the discrete visual tokens space, instead of the input raw pixel space, was proposed by BEiT as more efficient because it eliminates high-frequency visual details/noise of low or no semantic content. Otherwise, pretraining for MIM may focus on undesired, very fine-grained visual learning, due to the typically high spatial redundancy of images.

The loss function employed for BEiT pretraining is:
\begin{equation}
\mathcal{L}_{BEiT} = \frac{1}{K}\sum_{i=1}^K\mathcal{L}_{CE}(\mathbf{y}_i, \mathbf{p}_i),
\end{equation}
\noindent where $K=0.4M$ is the number of masked patches, $\mathbf{y}_i \in \mathbb{R}^{V}$ is the pseudo-ground-truth discrete distribution over visual tokens for the $i$-th masked patch of image $\mathbf{Y}$, while $\mathbf{p}_i \in \mathbb{R}^{V}$ is the corresponding predicted softmax distribution. The BEiT pseudocode is shown in \ref{alg:BEiTalg}.

\begin{algorithm}
\caption{BEiT pretraining pseudocode example for a single training image}
\label{alg:BEiT}
\begin{algorithmic}[1]
\REQUIRE $\mathbf{Y}$ - Input image, $P(\cdot)$ - Patch extraction function, $E(\cdot)$ - Embedding function, $T(\cdot)$ - ViT.
\STATE Divide $\mathbf{Y}$ into $M$ image patches $\mathbf{b}_i$ using $P$ and generate corresponding visual tokens $\mathbf{y}_i$.
\STATE Mask approximately 40\% of the patches in a blockwise manner, resulting in a masked image $\hat{\mathbf{Y}}$.
\STATE Replace the masked patches in $\hat{\mathbf{Y}}$ with learnable mask embeddings using $E$, creating a corrupted image $\tilde{\mathbf{Y}}$.
\STATE Pass $\tilde{\mathbf{Y}}$ through $T$ and obtain $M$ encoded patch representations $\mathbf{H}\in\mathbb{R}^{L \times M}$, where $\mathbf{h}_i \in \mathbb{R}^L$ is the $i$-th patch representation.
\STATE For each masked patch representation, obtain $\mathbf{p}_i = softmax(\mathbf{W}_c\mathbf{h}_i+\mathbf{b}_c)$, where $\mathbf{W}_c \in \mathbb{R}^{V \times L}$ and $\mathbf{b}_c \in \mathbb{R}^{V}$ are trainable parameters.
\STATE Compute $\mathcal{L}_{BEiT}$.
\STATE Back-propagate from $\mathcal{L}_{BEiT}$ to update parameters of $T(\cdot)$, $\mathbf{W}_c$ and $\mathbf{b}_c$.
\end{algorithmic}
\label{alg:BEiTalg}
\end{algorithm}

After pretraining, a task-specific neural head may be appended and the overall DNN can be finetuned for downstream tasks.

\subsubsection{MAE}
MAE \cite{he2022masked} (Masked AutoEncoders) is an alternative image reconstruction generative SSL pretext task, also designed mainly for Vision Transformers. In contrast to BEiT, it operates only in the raw pixel space and, thus, does not require any auxiliary neural networks. This is achieved by masking a significantly larger-than-typical proportion of the input image $\mathbf{Y}$ ($75\%$) and employing an entire Decoder subnetwork as a neural head for reconstructing the original image directly in pixel space. Importantly, the Encoder and the Decoder are asymmetric in model complexity.

In MAE pretraining, the Encoder only receives as input the non-masked image blocks and encodes them into a set of respective feature vectors. The lightweight Decoder receives the sequence of these feature vectors along with the appropriately placed mask embeddings. The overall DNN is trained so that the Decoder learns to output the full/uncorrupted original image $\mathbf{Y}$, which is utilized as pseudo-ground-truth target. Overall, the very high masking ratio, the very lightweight Decoder and the postponement of mask embedding processing until the decoding stage function synergistically so that the Encoder gradually learns semantically rich representations, instead of unimportant high-frequency visual details/noise, at comparatively low computational/memory demands. Essentially, masking 75\% of the input pixels makes pixel-level reconstruction challenging, despite the inherently high spatial redundancy of typical images, and, thus, the DNN is compelled to learn rich representations during pretraining.

The loss function employed for MAE is typically the MSE between the DNN's output \(\mathbf{P} \in \mathbb{R}^{W \times H \times 3}\) and the original image \(\mathbf{Y} \in \mathbb{R}^{W \times H \times 3}\): 

\begin{equation}
\mathcal{L}_{Mas} = \mathcal{L}_{MSE}(\mathbf{P}, \mathbf{Y}).
\end{equation}
\noindent However, in contrast to typical applications of $\mathcal{L}_{MSE}$, only the pixels corresponding to masked image regions are evaluated within the loss function; the unmasked/visible input patches do not contribute to loss calculations. The MAE pseudocode is shown in \ref{alg::MAEalg}.

\begin{algorithm}
\caption{MAE pretext task pseudocode example for a single training image}
\begin{algorithmic}[1]
\REQUIRE $\mathbf{Y}$ - Input image, $E(\cdot)$ - Encoder function mapping a corrupted version of $\mathbf{Y}$ to a sequence of latent space vectors $\mathbf{z}_i$, $D(\cdot)$ - Decoder function mapping the sequence of all $\mathbf{z}_i$ plus their positional embeddings back to $\mathbf{Y}$.
\STATE Corrupt $\mathbf{Y}$ by masking $75\%$ of its pixels, to generate $\hat{\mathbf{Y}}$.
\STATE Compute all $\mathbf{z}_i$ and their positional embeddings from the non-masked regions $\hat{\mathbf{Y}}$: $\{{\mathbf{z}_1,\dots,\mathbf{z}_i,\dots}\} \leftarrow E(\hat{\mathbf{Y}})$.
\STATE Compute the reconstructed image: $\mathbf{P} \leftarrow D(\{{\mathbf{z}_1,\dots,\mathbf{z}_i,\dots}\})$.
\STATE Compute loss $\mathcal{L}_{Mas}(\mathbf{P}, \mathbf{Y})$.
\STATE Back-propagate from $\mathcal{L}_{Mas}$ to update the parameters of $\mathbf{E}$ and $\mathbf{D}$.
\end{algorithmic}
\label{alg::MAEalg}
\end{algorithm}

After pretraining, the Decoder can be discarded and the pretrained Encoder DNN may be exploited for downstream finetuning.

\subsubsection{I-JEPA}
I-JEPA (`Image-based Joint-Embedding Predictive Architecture') \cite{assran2023self} is a generative SSL DNN pretraining approach that relocates image reconstruction from the pixel space to the latent/embedding space of abstract image representations. The motivation is to render the learnt features more semantically-oriented and less dependent on the pixel-level image details. The method is designed by default for ViT neural architectures and can be considered a modification of MAE where, for each training image, the single input, the $M$ predicted outputs and the $M$ respective pseudo-ground-truth targets are representation vectors in the embedding space. The targets are representations of $M$ potentially overlapping, randomly sampled image blocks of random aspect ratios and scales (each one occupying 15\%–20\% of the full image). They are acquired by appropriately cropping the complete image embedding $M$ times, according to the corresponding masks. Similarly, the input is the representation vector of a randomly sampled image block, initially corresponding to a significant portion of the image (at least 85\%), but subsequently trimmed so that all of its regions spatially overlapping with the selected targets have been removed. Any original image portions \textit{not} included in the final input image block have been essentially masked out.

The main DNN is composed of two ViTs arranged in an Encoder-Decoder scheme. The so-called `Context Encoder' receives the single raw input image block and generates its representation in the latent space. The Decoder\footnote{Called `Predictor' in \cite{assran2023self}.} receives this embedding, along with tokens encoding the position in pixel space of the $M$ target blocks, and predicts the embeddings corresponding to these target blocks. During pretraining, the $M$ pseudo-ground-truth target representations are derived by a separate copy of the Encoder ViT called `Target Encoder', which receives the complete raw input image, generates the respective embedding and spatially crops it $M$ times according to the selected target image blocks. The Context Encoder and the Decoder are trained in the typical manner, via error back-propagation and a Stochastic Gradient Descent variant, but the Target Encoder is simply updated over the training iterations with an exponential moving average of the Context Encoder's parameters\footnote{This design choice seems to have been borrowed from earlier contrastive SSL methods, such as \cite{he2020momentum}}.

The main advantage of I-JEPA is that, since it operates in the relatively low-dimensional latent space of representations, it is inherently more computationally efficient than MAE that functions in the raw pixel space. Additionally, since I-JEPA combines a variant of the image reconstruction task with abstract image representations, the embeddings learnt by the pretrained DNN strike by design a balance between capturing high-level semantics and encoding low-level image information. This suggests that I-JEPA is an SSL pretraining method particularly suitable for a wide range of downstream tasks, from high-level whole-image classification to low-level dense image prediction (e.g., monocular depth estimation).

Assuming $M$ target blocks per training image\footnote{$M$ is a fixed hyperparameter and \cite{assran2023self} suggests $M=5$.}, the loss function employed for I-JEPA is the average MSE between the $M$ Decoder predictions in the latent space and the respective pseudo-ground-truth targets:

\begin{equation}
\mathcal{L}_{IJEPA} = \frac{1}{M}\sum_{i=1}^M\mathcal{L}_{MSE}(D(\mathbf{z}_i), \mathbf{t}_i),
\end{equation}
\noindent where $D(\mathbf{z}_i)$ is the Decoder's prediction for the $i^{th}$ block and $\mathbf{t}_i$ is the corresponding pseudo-ground-truth target for that block.

After pretraining, the Decoder and the Target Encoder can be discarded, while the pretrained Context Encoder DNN may be exploited for downstream finetuning.

\subsection{Contrastive pretext}
Contrastive pretext tasks force the DNN to learn to differentiate between similar (positive) and dissimilar (negative) inputs. Typically, each pair of similar inputs is constructed by randomly augmenting/transforming a training image; then, the original (`anchor') and the transformed version comprise a positive pair. In contrast, a negative pair is composed of two different training images. The goal of contrastive pretext pretraining is for the DNN to learn an image representation that is invariant to appearance and/or geometric transformations (e.g., rotation, translation, scaling, cropping of multiple differently-sized image regions, etc.); therefore focusing on the semantic content \cite{Men2023}. The semantically inconsequential nature of such transformations, at least with respect to whole-image classification decisions, can be considered a form of prior knowledge.

The pretraining procedure relies on a contrastive loss function, such as InfoNCE \cite{oord2018representation} or NT-Xent \cite{chen2020simple}, that maximizes the similarity of DNN-encoded latent representations between the anchor and the positive input, while simultaneously minimizing the respective representational similarity between the anchor and the negative input. Notably, this contrastive SSL learning strategy may require the discrimination between each anchor and a multitude of visually similar though negative samples, a fact that renders the learning task challenging. After pretraining, the DNN has learnt to encode semantically similar/dissimilar inputs close to/away from each other, respectively, in its internal latent space. Key representative contrastive SSL methods are described in the followings.

\subsubsection{Early methods}
Contrastive visual representation learning has a long history \cite{hadsell2006}, but the concept of essentially treating each image as a unique semantic class appeared in \cite{dosovitskiy2014}. This approach poses a significant shift in granularity; from actual classes to individual instances. With the exponential growth in dataset size, maintaining instance-level representations becomes challenging. This led to the use of a `memory bank' \cite{wu2018} for storing each instance representation vector; the result is a streamlining of the training process and more efficient negative sampling. Subsequent algorithms extended the concept and the use of memory banks \cite{zhuang2019} \cite{tian2019} \cite{he2019} \cite{misra2019}.

However, typical contrastive methods with memory banks depend on eventually inserting the representations of all the training images into the cache, as computed by the DNN at their last sampling during the pretraining process. This gives rise to scaling and consistency issues. Thus, an alternative approach eventually emerged: in-batch negative sampling. Instead of using a cache structure, these methods exploit as negative samples the other images present in the same training mini-batch as the anchor \cite{doersch2017} \cite{ye2019} \cite{ji2019}, at the cost of requiring very large mini-batch sizes for proper pretraining.

\subsubsection{MoCo}
MoCo (`Momentum Contrast') \cite{he2020momentum} is a contrastive SSL pretraining algorithm that employs a large, fixed-size queue, as a slowly refurbished memory bank of negative samples, and two neural structures: the so-called `Momentum Encoder' (ME) and `Query Encoder' (QE). The ME is a copy of the QE that is updated over the training iterations with a slow moving average of the QE's parameters. The QE is the main DNN which is being pretrained and it is regularly updated by common training; during inference, it computes the representation of the input image.

During training with a positive pair, the anchor/transformed image is passed through the QE/ME, respectively, to generate the corresponding representations. Optimizing with the contrastive loss gradually leads to: a) maximizing the similarity between the anchor and the transformed image representations, and b) minimizing the similarity between the embedding of the anchor and those of the negative samples extracted from the memory bank. The latter one is constantly being replenished with newer image representations derived by the ME. The slowly progressing update of the ME parameters allows the representations it generates to be relatively consistent across training iterations, while the fixed queue size allows scaling to very large datasets. The employed loss function is the InfoNCE:

\begin{equation}
\mathcal{L}_{\text{MoCo}} = \mathcal{L}_{\text{NCE}}(\mathbf{a}, \mathbf{p}, \mathbf{R}),
\end{equation}
\noindent where $\mathbf{a} \in \mathbb{R}^L$ is the anchor image representation (generated by QE), $\mathbf{p} \in \mathbb{R}^L$ is the corresponding augmented/transformed image representation (generated by ME) and $\mathbf{R} \in \mathbb{R}^{L \times (K+1)}$ is a set of $K$ negative samples randomly selected from the current memory bank along with $\mathbf{p}$.

After pretraining the ME and the memory bank can be discarded, while the trained QE can be utilized for downstream finetuning.

\subsubsection{PIRL}
PIRL (`Pretext-Invariant Representation Learning') \cite{misra2019} is a contrastive SSL pretraining algorithm that utilizes a regular memory bank for negative sampling and two trainable linear projection heads that can be alternatively appended to the employed DNN. The first/second projection head maps the high-dimensional representation of the anchor image and its augmented/transformed counterpart, respectively, to a low-dimensional space before computing the contrastive loss. Thus, during training, both the anchor image and its transformed variant are passed through the single DNN in its current state, but are then projected by different heads. Additionally, PIRL slowly updates the cached representation of each dataset image as training progresses, as an exponential moving average of the stored representations (from prior epochs) of that image. This is in contrast to MoCo, which applies a slow moving average update on the separate Momentum Encoder's parameters, instead of applying it on the cached image representations, but the rationale in PIRL is similar: to ensure consistency of the representations for the negative samples during training.

The employed loss function is InfoNCE and operates similarly to the MoCo case. After pretraining, the projection heads and the memory bank can be discarded, while the trained DNN can be utilized for downstream finetuning.

\subsubsection{SimCLR}
SimCLR (`Simple Contrastive Learning of Representations') \cite{chen2020simple} is a contrastive SSL pretraining algorithm that generates two differently (but randomly) augmented/transformed images from each anchor image and seeks as a pretext task to maximize representational similarity between them. Therefore, each positive pair consists of two differently transformed variants of a single training image; not of such a transformed version and the original anchor. A single DNN is employed for generating all high-dimensional image representations and a single, lightweight, trainable neural projection head, i.e., a shallow, non-linear MultiLayer Perceptron (MLP), is employed for mapping them to a low-dimensional space before computing the contrastive loss. Negative sampling is in-batch (therefore, there is no memory bank) and all negative pairs are also randomly augmented/transformed before passing through the DNN. A large mini-batch size and the composition of multiple strong augmentation transformations have proven instrumental for SimCLR's performance.

The loss function employed in SimCLR is the Normalized Temperature-scaled Cross Entropy (NT-Xent), which is a variant of InfoNCE:
\begin{equation}
\mathcal{L}_{NT-Xent}(\mathcal{B}) = -\log \frac{\exp(\text{sim}(\mathbf{b}_i, \mathbf{b}_j) / \tau)}{\sum_{k=1}^{2B} \mathbb{1}_{[k \neq i]} \exp(\text{sim}(\mathbf{b}_i, \mathbf{b}_k) / \tau)},
\end{equation}
\noindent where $\text{sim}(\cdot)$ is the cosine similarity, $\tau$ is a scalar temperature hyperparameter and $\mathcal{B}$ is the set of $B$ images in the current training mini-batch. $\mathbf{b}_i$ and $\mathbf{b}_j$ are the latent representations of two differently augmented transformations of the in-batch image currently utilized as an anchor. Each image in the mini-batch is similarly augmented twice, resulting in $2B$ representations for negative sampling. The indicator function $\mathbb{1}_{[k \neq i]}$ is 1 when $k \neq i$ and 0 otherwise. The SimCLR pseudocode is shown in \ref{alg:SimCLR_batch}.

\begin{algorithm}
\caption{SimCLR pseudocode for a single mini-batch}
\begin{algorithmic}[1]
\REQUIRE Mini-batch $\mathcal{B} = \{\mathbf{B}^1, \mathbf{B}^2, ..., \mathbf{B}^B\}$, Encoder $g(\cdot)$, Projection head $f(\cdot)$, Temperature $\tau$.
\FOR{each $\mathbf{B}^i \in \mathcal{B}$}
    \STATE Generate two augmented instances $\mathbf{B}^i_1, \mathbf{B}^i_2$ from $\mathbf{B}^i$.
    \STATE Encode augmented instances: $\mathbf{h}^i_1 \leftarrow g(\mathbf{B}^i_1)$, $\mathbf{h}^i_2 \leftarrow g(\mathbf{B}^i_2))$.
    \STATE Project to lower dimension: $\mathbf{b}^i_1 \leftarrow f(\mathbf{h}^i_1)$, $\mathbf{b}^i_2 \leftarrow f(\mathbf{h}^i_2)$.
\ENDFOR
\FOR{each $i = 1$ to $B$}
    \STATE Compute $\mathcal{L}_{NT-Xent}(\mathcal{B})$.
    \STATE Back-propagate from $\mathcal{L}_{NT-Xent}(\mathcal{B})$ to update the parameters of $g(\cdot)$ and $f(\cdot)$.
\ENDFOR7
\end{algorithmic}
\label{alg:SimCLR_batch}
\end{algorithm}

After pretraining, the projection head can be discarded and the trained DNN can be utilized for downstream finetuning.

\subsubsection{Barlow Twins}
`Barlow Twins' \cite{zbontar2021barlow} is an SSL pretraining algorithm that relies on a heavily modified version of the common contrastive pretext tasks. It takes its name after the neuroscientist Horace Barlow, who proposed the efficient coding hypothesis \cite{barlow1961possible}. Similarly to SimCLR, Barlow Twins generates two differently (but randomly) augmented/transformed images (`views') from each training image, which are passed through the single DNN being pretrained. For each pair of views derived from one training image, the DNN is pushed to generate respective latent representations that are similar to each other \textit{and} do not contain redundant vector components. To achieve this, the two image view embeddings are treated as random vectors. Thus, their empirical cross-correlation matrix is evaluated during each pretraining iteration, using the generated representations of multiple different images across the entire current mini-batch as different observations. The loss function penalizes the difference of this matrix from the identity one.

The goal is to obtain an image representation that retains as much information about the input as possible, while capturing as little as possible about the particular distortions introduced by augmentation. No negative samples are considered and, therefore, neither a memory bank nor a very large mini-batch size are required. The removal of negative sampling from regular contrastive SSL pretext tasks would obviously create the danger of training converging to a trivial and useless solution, i.e., a DNN that constantly outputs a common representation vector for any input image. In Barlow Twins this `representation collapse' case is inherently avoided by the inclusion of the redundancy avoidance goal in the loss function, since any collapsed representations within a mini-batch would be heavily cross-correlated and the matrix in question would be far from diagonal. Finally, in contrast to regular contrastive methods, the Barlow Twins objective has been shown to actually require a high-dimensional latent space, in order for the pretrained DNN to perform well in downstream tasks; therefore, no compressive projection head is required. This property potentially stems from the fact that fully decorrelated image representations actually may need a large number of independent components in order to capture the necessary information.

Due to the absence of negative sampling, Barlow Twins does not employ an actual contrastive loss function. Its loss function is the following one:
\begin{equation}
\mathcal{L}_{BT}(\mathcal{B}) = \sum_{i=1}^L (1 - C_{ii})^2 + \lambda \sum_{i=1}^L \sum_{j=1, j\neq i}^L C_{ij}^2,
\end{equation}
\noindent where $\mathcal{B}$ contains the latent representations of the $2B$ views derived from the current mini-batch of size $B$. Thus, $\mathcal{B}$ holds the $L$-dimensional neural embeddings of two differently augmented views of each one among $B$ randomly sampled training images. $\mathbf{C} \in \mathbb{R}^{L \times L}$ is the empirical cross-correlation matrix of the two views, as computed across the mini-batch. The first term penalizes the deviation of the diagonal entries of \(\mathbf{C}\) from 1. The second term, weighted by a hyperparameter \(\lambda\), penalizes high absolute correlations between different features of the latent representations of the two views (non-zero off-diagonal elements of \(\mathbf{C}\)).

After pretraining the trained DNN can be directly utilized for downstream finetuning.

\subsection{Self-distillation pretext}

Neural distillation in general involves two DNNs, a `teacher' and a `student', where the latter one is being trained using the predictions of the former for each input as the corresponding pseudo-ground-truth target. In the case of SSL pretraining, the so-called `self-distillation' paradigm constitutes a category of pretext tasks that have recently emerged from the contrastive SSL methods, but do not require any negative sampling.

In typical formulations, the teacher and the student are of (almost) identical architecture and initialization. During SSL pretraining, their parameters are being updated concurrently, but at a different pace and/or with a different update mechanism, while the student learns to reproduce the latent image representation generated by the teacher for different transformed/augmented versions of the same training image. Eventually, the goal is again to learn an image representation that captures semantic information and is invariant to appearance/geometric perturbations. Key representative self-distillation SSL methods are described in the followings.

\subsubsection{BYOL}
BYOL (`Bootstrap Your Own Latent') \cite{grill2020bootstrap} is a self-distillation SSL pretraining method that employs two DNNs: a `target network' and an `on-line network'\footnote{The terms seem to have been inspired by Deep Q-Learning \cite{Mnih2015}}. These are architecturally identical, with the exception of an appended final prediction head in the latter one. The on-line network is actively trained in the typical manner, while the target network's parameters are being slowly updated across the training iterations as an exponential moving average of the corresponding on-line network's parameters. At each iteration, each of the two DNNs receives as input a differently augmented/transformed version of a common training image. The primary pretraining objective is to maximize the representational similarity of these two `image views' in a low-dimensional projection of the latent space of the two DNNs, across the entire training dataset. Batch normalization is modified so that batch statistics between the two views remain unshared, ensuring independent normalization for each view.

The on-line network learns by taking cues from the target network: as its parameters are being updated with each training iteration (via back-propagation and a Stochastic Gradient Descent variant), the target network receives corresponding updates at a slower pace (via the exponential moving average formulation). This dynamic, combined with the training objective, sets the stage for the on-line network (serving as the student) to `chase' the target network (serving as the teacher), effectively learning from a slightly older version of itself which is being dynamically constructed. Similarly to SimCLR, a shallow, non-linear MLP is utilized as a projector that maps latent image representations to a low-dimensional space before computing the loss; however, different sets of trainable parameters are attached to the on-line and to the target network's projector.

BYOL focuses solely on bringing positive pairs close together in the DNNs' latent space, negating the need for negative sampling. This simplifies learning and trims down the computational overhead. This absence of negative samples within the training objective and the very close relationship between the teacher and the student DNN jointly create the possibility of convergence to a trivial solution that perfectly minimizes the training loss in an undesired manner, i.e., a DNN that always outputs identical representation vectors for any and all input images. However, the use of the exponential moving average update mechanism for the target network and of the final prediction head for the on-line network help to avoid this collapse scenario in practice. Finally, BYOL has been shown to be much more robust than SimCLR to the selected mini-batch size and to the choice of augmentation transformations.

Assuming that $\mathbf{b}_1, \mathbf{b}_2 \in \mathbb{R}^{L}$ are the latent representations of two differently augmented views of a single training image, i.e., one encoded by the target network and one by the on-line network, the loss function employed for BYOL pretraining is their Mean Squared Error (MSE):
\begin{equation}
\mathcal{L}_{BYOL} = \mathcal{L}_{MSE}(\mathbf{b}_1, \mathbf{b}_2).
\end{equation}

After pretraining the teacher/target DNN is discarded, along with the projector and the prediction head of the trained student/on-line DNN. The remaining student backbone can be directly utilized for downstream finetuning.

\subsubsection{SimSiam}
SimSiam \cite{chen2021exploring} is a self-distillation SSL pretraining method that follows closely on the footsteps of BYOL, but utilizes a Siamese architecture: a single, common DNN is exploited for processing both augmented views of each training image, similarly to SimCLR. This is done by dispensing entirely with the exponential moving average update mechanism and employing identical parameters for the student/on-line and the teacher/target network, except of course for the appended prediction head at the end of the on-line network only. The collapse to a trivial solution is avoided by simply \textit{not} updating the DNN parameters (via regular back-propagation and a Stochastic Gradient Descent variant) based on the loss's gradient with regard to the target network's output. Only the loss gradient back-propagating through the on-line network contributes to DNN parameter update during pretraining. Despite conjectures that have been proposed regarding why this design choice helps to avoid collapsing \cite{zhang2022does}, the reason is not yet fully clear. Nevertheless, the addition of a BYOL-like exponential moving average update mechanism for the target network, which thus becomes fully distinct from the on-line DNN parameter-wise, helps to further increase the achievable downstream accuracy. Baseline SimSiam holds an accuracy advantage over BYOL only when the number of pretraining iterations is limited.

The loss function employed for SimSiam pretraining is the negative cosine similarity of the latent representations of two differently augmented views of a single training image, i.e., $\mathbf{b}_1, \mathbf{b}_2 \in \mathbb{R}^{L}$:

\begin{equation}
\mathcal{L}_{\text{SimSiam}} = -\mathcal{L}_{sim}(\mathbf{b}_1, \mathbf{b}_2).
\end{equation}

After pretraining, the projector and the prediction head of the single common DNN can be discarded, while the remaining backbone can be directly utilized for downstream finetuning.

\subsubsection{DINO}
DINO (`DIstillation with NO labels') \cite{caron2021emerging} is a self-distillation SSL method that has been designed with Vision Transformer (ViT) DNNs in mind, unlike the majority of architecture-agnostic SSL algorithms, although it is equally applicable to CNNs. It can be considered a modified version of BYOL, where a softmax final layer is employed in both the student and the teacher DNN, resulting in the output image representations being probability distributions. Additionally, the final prediction head has been removed from the student/on-line network and, therefore, the student and the teacher are architecturally identical (albeit distinct parameter-wise). Since simple removal of the predictor from the on-line network would make convergence to a trivial/collapsed solution significantly more probable, this undesired scenario is avoided by centering and sharpening the teacher/target network's output before computing the loss. Along with the exponential moving average update mechanism for the target network, this is enough to prevent collapse.

A potential reason is that the composition of centering and sharpening explicitly mitigates \textit{both} potential forms of collapse: a) one where the DNN always predicts a high-entropy uniform representation for any input image, and b) one where the DNN always predicts a low-entropy impulse representation for any input image. Then: a) sharpening is implemented by proper tuning of the softmax temperature hyperparameter, while b) centering is equivalent to adding a dynamically updated bias term to the teacher's output; this bias is computed by considering all teacher outputs for the current mini-batch.

An important property observed with DINO is that the teacher almost constantly outperforms the student during pretraining, on the grounds of the former one being essentially an ensemble learning model. Potentially this is a main contributor to the impressive performance of DINO-pretrained DNNs, especially with ViT neural architectures. In fact, ViTs pretrained with DINO on ImageNet without any supervision have been shown to have learnt representations readily suited for semantic image segmentation. Moreover, they lead to competitive downstream classification accuracy without \textit{any} finetuning, by simply utilizing their representations for building a rudimentary k-Nearest Neighbours (kNN) classifier. Such zero-shot learning abilities are comparable to what SSL-pretrained Large Language Models (LLMs), in the field of NLP, or weakly supervised vision-language DNNs like CLIP \cite{radford2021learning} are capable of, a fact that has led to the recent emergence of the umbrella term `foundation models' for all similar cases.

Let us assume that $\mathbf{B}^g_1, \mathbf{B}^g_2$ are two differently augmented `global' crops of a single training image, i.e., covering a significant portion of the original image's area, while $k$ additional, differently augmented `local' crops, i.e., covering a substantially smaller portion of the original image's area, are independently and randomly derived. $\mathcal{V}$ is the set of all $k+2$ views obtained from a single training image, which includes $\mathbf{B}^g_1$ and $\mathbf{B}^g_2$ among its elements. All $k+2$ views are passed through the student network $S(\cdot)$, but only the two global views are passed through the teacher network $T(\cdot)$. Then the loss function employed for DINO pretraining is the following one:
\begin{equation}
    \mathcal{L}_{DINO} = \sum_{\substack{\mathbf{B}_i\in\{\mathbf{B}^g_1, \mathbf{B}^g_2\}}} \sum_{\substack{\mathbf{B}_j\in\mathcal{V} \\ \mathbf{B}_j\neq\mathbf{B}_i}} \mathcal{L}_{CE}(T(\mathbf{B}_i),S(\mathbf{B}_j)),
\end{equation}
\noindent where the outputs of $S(\cdot)$ and $T(\cdot)$ are softmax distributions.

After pretraining, the student and the projection head of the teacher DNN can be discarded. The remaining teacher backbone can be utilized for downstream finetuning, or directly for extracting image representations. The DINO pseudocode is shown in Algorithm \ref{alg::DINOpseudo}.

\begin{algorithm}
\caption{DINO pretext task pseudocode for a single training image}
\begin{algorithmic}[1]
\REQUIRE $\mathbf{Y}$ - Input image, $S(\cdot)$ - Student, $T(\cdot)$ - Teacher, $\mathcal{V}$ - Set of $k+2$ augmented views derived from $\mathbf{Y}$, $\alpha$ - Momentum hyperparameter, $\mathbf{c}$ - Centering parameter, $m>0$ centering update rate hyperparameter, $B$ - Mini-batch size
\STATE Generate $\mathcal{V}$ from $\mathbf{Y}$, containing $k$ local views/augmented crops and 2 global views/augmented crops.
\STATE Compute each $S(\mathbf{B}_i)$, where $\mathbf{B}_i$, $1 \leq i \leq (k+2)$ is the $i$-th element of $\mathcal{V}$.
\STATE Compute $T(\mathbf{B}^g_1)$ and $T(\mathbf{B}^g_2)$, where $\mathbf{B}^g_1, \mathbf{B}^g_2$ are the two global views within $\mathcal{V}$.
\STATE Apply centering to $T(\mathbf{B}^g_1)$ and $T(\mathbf{B}^g_2)$: $T(\mathbf{B}_i) \leftarrow T(\mathbf{B}_i) + \mathbf{c}$.
\STATE Compute $\mathcal{L}_{DINO}$.
\STATE Back-propagate from $\mathcal{L}_{DINO}$ to update the parameters of $S(\cdot)$.
\STATE Update the parameters $\mathbf{w}_T$ of $T(\cdot)$ with an exponential moving average of the parameters $\mathbf{w}_S$ of $S(\cdot)$: $\mathbf{w}_T \leftarrow \alpha \mathbf{w}_T + (1 - \alpha) \mathbf{w}_S$.
\STATE Update $\mathbf{c}$: $\mathbf{c} \leftarrow m\mathbf{c} + (1-m)\frac{1}{B}\sum_{j=1}^B T(\mathbf{B}_j)$, where $\mathbf{B}_j$ is the $j$-th image in the current training mini-batch.
\end{algorithmic}
\label{alg::DINOpseudo}
\end{algorithm}

DINO has been enhanced with the later version DINOv2, which incorporates various improvements borrowed from competing or earlier methods \cite{Oquab2023DinoV2}. The authors have coupled DINOv2 with a very large/complex ViT neural architecture and an extremely large-scale pretraining dataset of 142M images, in order to derive a highly competent foundation model for computer vision with very impressive zero-shot learning abilities. The most important algorithmic improvements over DINO are the following ones:
\begin{itemize}
    \item A complementary image reconstruction training objective is added that is borrowed from iBOT (`image BERT pretraining with On-line Tokenizer') \cite{zhou2021ibot}, an SSL method that essentially merges BEiT with self-distillation. In iBOT, the teacher serves as an on-line (instead of pretrained) visual tokenizer. Thus, in DINOv2, the image view passing through the student has randomly masked regions, while this is not the case for the image view fed to the teacher. Then, the additional objective is prescribed by a CE loss between the generated features of both networks on the masked regions.
    \item DINO's centering of teacher outputs is replaced in DINOv2 by an alternative Sinkhorn–Knopp (SK) normalization process \cite{cuturi2013sinkhorn}. This approach for avoiding solution collapse to a common, low-entropy impulse representation has been borrowed from \cite{ruan2022weighted}, i.e., a recently proposed pretraining-time ensemble learning method for DINO that computes data-dependant importance weights for ensemble members.
    \item The KoLeo regularizer \cite{sablayrolles2018spreading} has been added to the training objectives. This is a differential entropy regularizer that is employed for spreading the output representations of different images within a mini-batch.
\end{itemize}

\subsection{Clustering pretext}
SSL via clustering involves, in general, grouping images into clusters, based on their similarity in the input space, and pretraining a DNN to predict the cluster assignments. Since clustering is an unsupervised task, no actual ground-truth labels are utilized. The cluster assignments are used as pseudo-ground-truth targets. In most cases, any clustering algorithm can be used (e.g., K-Means), but certain SSL methods incorporate a specific on-line clustering approach. The quality of the learnt representation depends on the suitability of the considered clustering algorithm and the similarity metric used to group the images \cite{zhou2022deep}. Key representative clustering SSL methods are described in the followings.

\subsubsection{DeepCluster}
DeepCluster \cite{caron2018deep} attempts to structure the high-dimensional latent space of DNNs via an iterative approach of alternating phases/steps. It assumes a single, randomly initialized DNN to which a final classification neural head has been attached for SSL pretraining. Starting with inferring image representations via this DNN for the entire training set, said generated features are then clustered using K-Means. The resultant cluster assignments are utilized as pseudo-ground-truth labels that guide self-supervised DNN training for regular classification in the subsequent epoch. These two phases alternate. As training progresses, the latent space gets refined and this, in turn, provides a more nuanced structure during the succeeding clustering phase. Thus, during each classification step the pseudo-ground-truth labels are kept fixed, but they are updated once for the entire training set during the next clustering step. This cycle continues, allowing the two alternating phases to continually aid each other.

In order to prevent trivial or collapsed solutions to the optimization problems of both phases, DeepCluster employs common tricks for class-imbalanced classification and for avoiding empty clusters. Data augmentation pre-applied to the training set imbibes a degree of representational invariance to non-semantic transformations. Notably, ImageNet-1k training using DeepCluster leads to clusters that often align with human-understandable semantic classes, but the approach is not sufficiently robust to suboptimal clusterings or to the choice of data augmentation transformations.

The loss function employed for the SSL DNN pretraining phases on the classification pretext task is regular Cross-Entropy (CE):

\begin{equation}
\mathcal{L}_{DeepCluster} = \mathcal{L}_{CE}(\mathbf{y}_i,\mathbf{p}_i),
\end{equation}
\noindent where $\mathbf{y}_i$/$\mathbf{p}_i$ is the pseudo-ground-truth/predicted distribution over $K$ pseudo-classes/clusters, respectively, for the $i$-th training image.

After pretraining, all metadata generated by the clustering phases and the classification head can be discarded, while the pretrained backbone DNN may be utilized for downstream finetuning. The DeepCluster pseudocode is shown in \ref{alg::DeepCluster}.

\begin{algorithm}
\caption{DeepCluster pretext task pseudocode}
\begin{algorithmic}[1]
\REQUIRE $\mathcal{X}$ - Training dataset, $f(\cdot)$ - DNN feature extractor function, $C(\cdot)$ - Clustering function
\STATE Initialize parameters of $f(\cdot)$ (e.g., randomly).
\WHILE{not converged}
    \STATE Compute set of representations $\mathcal{H}$ containing all $\mathbf{h}_i \leftarrow \mathbf{f}(\mathbf{X}_i)$, $\mathbf{X}_i \in \mathcal{X}$.
    \STATE Compute cluster assignments $C(\mathcal{H})$ to obtain set of pseudo-ground-truth-labels $\mathcal{Y}$.
    \FOR{each mini-batch $\mathcal{B}$ sampled from $\mathcal{X}$}
        \STATE Compute $\mathcal{L}_{DeepCluster}$ for each image in $\mathcal{B}$.
        \STATE Back-propagate from $\mathcal{L}_{DeepCluster}$ to update parameters of $f(\cdot)$.
    \ENDFOR
\ENDWHILE
\end{algorithmic}
\label{alg::DeepCluster}
\end{algorithm}

DeepCluster has been enhanced in DeepClusterV2 \cite{caron2020unsupervised}, which incorporates a number of various improvements. The most important ones are the following:
\begin{itemize}
\item Pretraining has been accelerated by removing precautions for avoiding trivial solutions, since this scenario is not commonly observed in practice, and by re-using image representations from the previous epoch at the start of each clustering phase, instead of inferring them anew using the current DNN. This can be considered a form of memory bank caching without momentum.
\item As in \cite{Asano2020}, multiple clustering problems are simultaneously solved independently and in a multitask fashion. Thus, the classification training phases proceed by the sum of the respective losses, assuming a common, shared set of DNN parameters.
\item Strong data augmentation and the MLP projection head are borrowed from SimCLR \cite{chen2020simple}.
\item Spherical K-Means is utilized for the clustering phases, instead of vanilla K-Means.
\end{itemize}

\subsubsection{SeLa}
SeLa (`Self Labelling') \cite{Asano2020} is an algorithm for SSL DNN pretraining via clustering. It can be considered as a modification of DeepCluster, since it also depends on consecutive iterations of two alternating steps. The classification pretext pretraining phase is identical to that of DeepCluster. However, in the clustering phase, instead of simply performing K-Means on all inferred image representations, a replacement method has been devised. Finding the cluster assignment for each image under a constraint of equally sized clusters, for avoiding undesired trivial solutions where all images are placed in a single dominant cluster, is formulated as an optimal transport problem and solved efficiently via the Sinkhorn-Knopp algorithm \cite{cuturi2013sinkhorn}. The overall two-step iterative method is formally derived as a way to optimize a single objective with regard to both DNN parameters and cluster assignments, a fact that mathematically guarantees convergence to a local optimum. Finally, SeLa pre-applies typical data augmentation schemes to the training set and adopts the multi-clustering approach that can also be found in DeepClusterV2. Multi-clustering is implemented in SeLa by using parallel classification heads (one per clustering task), with a shared backbone DNN.

As in DeepCluster, the loss function actually employed for the classification pretext task is Cross-Entropy (CE). However, the general optimization objective with regard to both DNN parameters and cluster assignments, from which the overall algorithm is derived, is the following one:

\begin{equation}
\boldmin_{p,q}: - \frac{1}{N} \sum_{i=1}^{N} \sum_{y=1}^{K} q(y|\mathbf{X}_i) \log p(y|\mathbf{X}_i)
\end{equation}

\indent\indent\indent\indent\indent \textbf{s.t.}:

\begin{equation}
q(y|\mathbf{X}_i) \in {0,1}, \forall y \in {1, 2,\dots,K}, \forall i \in {1, 2,\dots,N},
\end{equation}
\begin{equation}
\sum_{y=1}^{K} q(y|\mathbf{X}_i)=1, \forall i \in {1, 2,\dots,N},
\end{equation}
\begin{equation}
\sum_{i=1}^{N} q(y|\mathbf{X}_i) = \frac{N}{K}, \forall y \in {1, 2,\dots,K},
\end{equation}

\noindent where $K$ is the number of pseudo-classes/clusters, $N$ is the number of training images, $p(y|\mathbf{X}_i)$ is the predicted, softmax-derived probability distribution of class labels over the $K$ classes, given the $i$-th training image $\mathbf{X}_i$, while $q(y|\mathbf{X}_i)$ is the corresponding pseudo-ground-truth distribution over the $K$ classes; $q$ can be considered to be a fully deterministic impulse distribution.

After pretraining, the classification heads can be discarded and the pretrained backbone DNN may be utilized for downstream finetuning.

\subsubsection{SwAV}
SwAV (`Swapping Assignments between multiple Views of the same image') \cite{caron2020unsupervised} is an SSL method the combines characteristics from both contrastive and clustering SSL approaches. It does not utilize negative samples and, therefore, does not rely on a memory bank, a momentum encoder and/or a large mini-batch size. More importantly, the vectors that are being pushed close together by its training objective are \textit{not} directly the image representations of differently augmented/transformed views of a single training instance/image, but their cluster assignment vectors from an on-line image clustering process that occurs in parallel. Thus, the goal of DNN pretraining is to arrive at a set of DNN parameters that are optimal in the following sense: they induce such image representations that the cluster assignment vector of one view can be predicted from the latent representation of the other view, and vice versa. Essentially, representations of semantically identical images are gradually forced to become approximately similar to each other so as to generate appropriate, simplified attention weights vectors \cite{bahdanau}, using each $\mathcal{L}_2$-regularized representation as a query and the $K$ $\mathcal{L}_2$-regularized, learnt cluster centroids within the latent space as a set of keys\footnote{$K$ is a fixed hyperparameter, as in K-Means.}. These simplified attention weights vectors are forced by the training process to eventually approximate the learnt cluster assignment vectors that, as training proceeds, become consistent with the semantic content and invariant to appearance/geometric transformations. The on-line clustering aspect, where processing takes place in-batch instead of partitioning the entire training set in dedicated clustering phases, allows the method to scale well for very large pretraining datasets.

Within each training iteration, the training objective is optimized in two steps. First, the cluster assignments are learnt for the current mini-batch, assuming constant/fixed cluster centroids and DNN parameters. This is done by solving an optimal transport problem with an equipartition constraint, for avoiding collapse to a trivial solution of one dominant cluster, using the iterative Sinkhorn-Knopp algorithm. The approach is similar to that of \cite{Asano2020}, but modified here to operate within a mini-batch due to the on-line clustering setting. Second, assuming fixed cluster assignment vectors, the centroids and the neural network parameters are updated by minimizing the loss function in the typical manner (error back-propagation and a Stochastic Gradient Descent variant). This loss function makes use of the Cross-Entropy (CE) between cluster assignment vectors and the simplified attention weights vectors previously defined:

\begin{equation}
\mathcal{L}_{\text{SwAV}} = \mathcal{L}_{CE}(\mathbf{q}^{v1}_i, \mathbf{p}^{v2}_i) - \mathcal{L}_{CE}(\mathbf{q}^{v2}_i, \mathbf{p}^{v1}_i),
\end{equation}
\begin{equation}
\mathbf{p}^{v*}_i = softmax(\mathbf{C}\mathbf{h}^{v*}_i),
\end{equation}

\noindent where $\mathbf{C} \in \mathbb{R}^{K \times L}$ contains the $K$ $\mathcal{L}_2$-regularized, current cluster centroids and $\mathbf{h}^{v1}_i$/$\mathbf{h}^{v2}_i$ is the $\mathcal{L}_2$-regularized, $L$-dimensional latent representation of the first/second augmented view of the $i$-th training image, respectively. $\mathbf{q}^{v1}_i$/$\mathbf{q}^{v2}_i$ is the $K$-dimensional current cluster assignment vector for the first/second augmented view of the $i$-th training image, respectively.

After pretraining the DNN can be directly utilized for downstream finetuning, while all clustering metadata can be discarded.
\end{document}